%% file: public.tex
\documentclass{article} 
\usepackage[preprint]{neurips_2026}

\input{math_commands.tex}

\usepackage{booktabs}
\usepackage{colortbl}
\usepackage{hyperref}
\usepackage{url}
\usepackage{graphicx}

\usepackage{hyperref} 

\title{OSWorld-Science: A Benchmark of Computer Use Agents for Learning and Using Scientific Software}

\newcommand{\name}{OSWorld-Science}

\newcommand{\paperauthors}{%
\parbox{\dimexpr\textwidth-2\tabcolsep\relax}{\centering\normalfont\small
\mbox{Dingyuan Dai$^{2,*}$} \quad
\mbox{Heli Qi$^{3,*}$} \quad
\mbox{Lei Liu$^{6,*}$} \quad
\mbox{Yinxi Li$^{4,*}$} \quad
\mbox{Baiding Chen$^{5,*}$} \quad
\mbox{Zijun Dou$^{1,*}$} \\[0.4em]
\mbox{Qingcheng Zeng$^{7}$} \quad
\mbox{Qi Kang$^{8}$} \quad
\mbox{Oliver Sun$^{9}$} \quad
\mbox{Eric Wang$^{9}$} \quad
\mbox{Bo Zhou$^{10}$} \quad
\mbox{Haixin Wang$^{2}$} \quad
\mbox{Yufan Du$^{2}$} \quad
\mbox{Shi Bo$^{11}$} \quad
\mbox{Ruihan Lin$^{4}$} \quad
\mbox{Mengqi Yuan$^{12}$} \quad
\mbox{Dunjie Lu$^{12}$} \quad
\mbox{Steven Dillmann$^{13}$} \quad
\mbox{Yiming Shi$^{6}$} \quad
\mbox{Tina Su$^{6}$} \quad
\mbox{Amy Xin$^{1}$} \quad
\mbox{Minghao Liu$^{16}$} \quad
\mbox{Xi Wang$^{14}$} \quad
\mbox{Xu Huang$^{9}$} \\[0.4em]
\mbox{Ge Zhang$^{15,\dagger}$} \quad
\mbox{Pengyu Nie$^{4,\dagger}$} \quad
\mbox{Zhen Yang$^{1,\dagger}$} \quad
\mbox{Jie Tang$^{1,\dagger}$} \quad
\mbox{Juanzi Li$^{1,\dagger}$} \quad
\mbox{Weihao Xuan$^{3,15,16,\dagger}$} \quad
\mbox{Tianyu Liu$^{1,6,*,\dagger}$}
\par\vspace{0.6em}
{\footnotesize
\mbox{$^{1}$Tsinghua University} \quad
\mbox{$^{2}$University of California, Los Angeles} \quad
\mbox{$^{3}$RIKEN AIP} \quad
\mbox{$^{4}$University of Waterloo} \quad
\mbox{$^{5}$Carnegie Mellon University} \quad
\mbox{$^{6}$Yale University} \quad
\mbox{$^{7}$Northwestern University} \quad
\mbox{$^{8}$Zhejiang University} \quad
\mbox{$^{9}$University of California, Berkeley} \quad
\mbox{$^{10}$University of Illinois Chicago} \quad
\mbox{$^{11}$Boston University} \quad
\mbox{$^{12}$The University of Hong Kong} \quad
\mbox{$^{13}$Stanford University} \quad
\mbox{$^{14}$New York University} \quad
\mbox{$^{15}$TokenWave.AI} \quad
\mbox{$^{16}$The University of Tokyo}
\par}
\vspace{0.4em}
{\footnotesize $^{*}$Equal contribution (co-first authors).\quad $^{\dagger}$Corresponding authors.\par}
}
}

\author{\paperauthors}

\makeatletter
\let\originaltoptitlebar\@toptitlebar
\renewcommand{\@toptitlebar}{%
  \hbox to \textwidth{\hfil\includegraphics[width=0.96\textwidth]{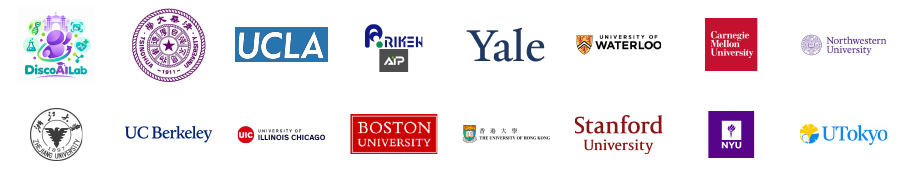}\hfil}%
  \vskip 8pt
  \originaltoptitlebar
}
\makeatother

\begin{document}

\maketitle

\begin{abstract}
Scientific software presents a demanding test for computer-using agents based on visual language models (VLMs): completing a research workflow requires interpreting specialized interfaces, manipulating scientific objects, and producing verifiable results. We thus introduce OSWorld-Science, a benchmark and evaluation environment that combines scientifically meaningful tasks, artifact-based evaluation, and an efficient agent harness for studying computer use in the scientific domain. The benchmark contains \textbf{12} VLMs and \textbf{146} high-quality tasks across several scientific domains and software configurations, covering workflows such as molecular drawing and retrosynthesis, pathology image analysis, statistical computing, and physical simulation. Tasks are developed through expert proposals and iterative human–AI co-design, with selection guided by scientific value and difficulty. Task-specific execution-based evaluators inspect application states and generated artifacts, including molecular structures, segmentation masks, plots, and numerical results, and award partial credit for incomplete outcomes. Our special harness integrates model adapters, interaction-loop control, and trajectory logging to support comparisons of models and interaction strategies. This combination makes OSWorld-Science a testbed for examining how agents coordinate visual interpretation with graphical and command-line actions. Our results shows that current state-of-the-art VLMs with a strong harness still face challenges in addressing key questions in the scientific domains. We also analyze the benchmarking results across multi-linguistics, reasoning efforts, context length and other factors and derive several important conclusions and directions to assist future development. Overall, we provide an integrated framework connecting expert-defined scientific goals to verifiable software outcomes, enabling systematic evaluation of both agent capabilities and harness design in scientific workflows. \textbf{We also welcome public contributions now: \href{https://docs.google.com/forms/d/e/1FAIpQLSeRYkEIXTsllKYqSP8nYVLpA9KYhPFvMFRxseyc4y_HWnadlg/viewform?usp=dialog}{Link}!}
\end{abstract}

\section{Introduction}

\input{introduction}

\section{Related Work}

\input{relatedwork}

\section{Tasks, Environments, and Harness in OSWorld-Science}

\input{methods}

\section{Results}

\input{results}

\section{Conclusion}
To evaluate the ability of CUA in operating software to solve scientific problems, we propose OSWorld-Science, a workflow that includes task design, a CUA harness, and evaluation metrics and results. Our experimental results indicate that CUAs based on state-of-the-art VLMs still have room for improvement in solving scientific problems, particularly in terms of problem understanding and learning how to operate the software. At the same time, in-depth analysis reveals that there are many common issues underlying the agents’ operational errors, which are difficult to resolve through conventional methods (such as increasing reasoning effort or expanding the context window). Therefore, future research should focus on the development of scientific agents.

\section{Dataset availability}
All of the datasets including question sets and docker images can be accessed via the huggingface link \url{https://huggingface.co/datasets/SciAILab/OSWorld-Science-data}. We have added contamination watermark to detect and avoid using our datasets for model training. Codes of this project can be found in this repo \url{https://github.com/DiscoAILab/OSWorld-Science} and website \url{https://huggingface.co/spaces/SciAILab/osworld-science-page}. Since we evaluated some software that requires a license, we do not plan to publish all the trajectories in this version.

\section{Reproducibility statement}
Task definitions, hidden verifiers, oracle references and complete per-run trajectories are
archived, as are the per-run results table, the model matrix and the ablation data used for every
table in this paper. Every number in the tables and figures is generated programmatically from those
files, by scripts released with the benchmark. 

\section{Acknowledgment}
T.L. acknowledges the support of computing resources from Kimi, DeepSeek, Z.ai, Google, Alibaba, and TokenWave.AI. 

\bibliography{iclr2027_conference}
\bibliographystyle{iclr2027_conference}

\newpage

\appendix

\input{ablation_appendix}
\end{document}

%% file: math_commands.tex
\usepackage{amsmath,amsfonts,bm}

\def\eqref#1{equation~\ref{#1}}

\def\1{\bm{1}}

\DeclareMathAlphabet{\mathsfit}{\encodingdefault}{\sfdefault}{m}{sl}
\SetMathAlphabet{\mathsfit}{bold}{\encodingdefault}{\sfdefault}{bx}{n}



%% file: introduction.tex
Recent advances in multimodal foundation models have enabled computer-using agents (CUAs) that perceive graphical interfaces, reason over natural-language instructions, and execute mouse and keyboard actions. Benchmarks such as WebArena and VisualWebArena first studied realistic web interaction, while OSWorld expanded evaluation to open-ended workflows across real desktop applications and operating-system functions \citep{zhou2023webarena,koh2024visualwebarena,xie2024osworld}. Subsequent environments have broadened this paradigm to Windows and mobile devices \citep{bonatti2024windowsagentarena,rawles2024androidworld}. Together, these results suggest that CUAs could make complex software more accessible and improve productivity, but they also reveal large gaps in visual grounding, long-horizon planning, and operational knowledge.

Scientific work is a particularly consequential setting for computer use. Researchers always rely on specialized software for simulation, visualization, data analysis, and experimental control, often facing steep learning curves and heterogeneous data modalities. Scientific agents such as ChemCrow, Coscientist, PaperQA, and The AI Scientist demonstrate that language-model agents can retrieve literature, invoke domain tools, execute experiments, and automate parts of the research cycle \citep{bran2024chemcrow,boiko2023coscientist,lala2023paperqa,lu2026aiscientist,chen2026aiagents,liu2026towards}. However, using professional scientific software remains substantially different from answering scientific questions with LLM-based agents. The success requires agents to understand domain-specific interfaces, maintain state over long workflows, and produce scientifically valid artifacts.

Existing evaluations only partially capture these requirements. General CUA benchmark does not include special consideration of scientific challenges \citep{xie2024osworld, yuan2026osworld2, zhou2023webarena, he2024webvoyager}, while other benchmarks such as ScienceBoard evaluates multimodal agents on curated workflows in professional scientific applications, whereas Terminal-Bench-Science focuses on expert-contributed, terminal-based research workflows with artifact-level verification \citep{sun2025scienceboard,terminalbenchscience2026}. LogicVista considers evaluating the scientific reasoning ability of VLMs \citep{xiao2024logicvista}. These efforts establish the importance of realistic scientific environments, but cannot construct benchmarks around questions that scientists themselves consider important and difficult, how to evaluate heterogeneous outputs such as molecular structures, plots, and statistical results under a unified framework, and how much performance depends on the agent harness rather than the backbone model alone. The latter issue is especially salient as hybrid GUI: tool interfaces can materially change computer-use performance \citep{jia2025osworldmcp}.

Here we introduce \name{} (Figure \ref{fig:overall}), a playground for recording, analyzing, and evaluating how CUAs operate scientific software. Our system has three components. First, in collaboration with domain experts, we construct an importance-driven benchmark centered on scientifically meaningful, challenging research questions in their respective fields. Second, we design evidence-based evaluators for heterogeneous outputs, including chemical formulas, molecular or protein structures, visualizations, and statistical measures, enabling consistent analysis across disciplines. Third, we improved the general CUA harness to adapt to more VLMs and created several plug-in components to make it more efficient, including the loop controller, the VLM adapter, and the user interaction platform. Finally, we use fine-grained trajectory and error analysis to refine the agent harness for scientific software, improving adaptation without additional model training. Our benchmark is deliberately specialized and challenging: across a range of frontier open and proprietary models, the strongest agent only achieves a $\sim$70\% success rate on average and $\sim$40\% in the most difficult set, underscoring the need for both more powerful agents and better scientific-computer-use systems.

%% file: relatedwork.tex
\paragraph{Computer-using agents and benchmarks.}
Early realistic environments concentrated on web navigation. WebArena provides reproducible websites and functional task evaluation, and VisualWebArena adds visually grounded tasks that require joint image-text understanding \citep{zhou2023webarena,koh2024visualwebarena}. OSWorld extends evaluation from the browser to a full operating system, covering open-ended tasks across desktop applications, file operations, and cross-application workflows with execution-based graders \citep{xie2024osworld}. Windows Agent Arena emphasizes scalable evaluation in a real Windows environment, while AndroidWorld introduces dynamically generated tasks and state-based evaluation on mobile devices \citep{bonatti2024windowsagentarena,rawles2024androidworld}. More recently, OSWorld-MCP studies agents that can choose between GUI actions and structured tools, demonstrating that the interaction layer and harness are themselves important experimental variables \citep{jia2025osworldmcp}. In contrast to these general-purpose benchmarks, \name{} focuses on professional scientific software, expert-defined problem significance, heterogeneous scientific artifacts, and harness design for domain-intensive workflows.

\paragraph{Agents for scientific discovery.}
LLM-based scientific agents have been developed for literature synthesis, domain-tool use, experimental planning, and end-to-end research automation. PaperQA retrieves and synthesizes evidence from full-text scientific literature \citep{lala2023paperqa}; ChemCrow integrates expert chemistry tools for synthesis, drug discovery, and materials tasks \citep{bran2024chemcrow}; and Coscientist couples language models with documentation search, code execution, and laboratory automation \citep{boiko2023coscientist}. The AI Scientist further explores open-ended automation of idea generation, experimentation, analysis, and paper writing \citep{lu2026aiscientist}. These systems demonstrate the value of domain knowledge and external tools, but their evaluations are typically tied to a particular agent architecture, domain, or tool collection.

\paragraph{Benchmarking agents for scientific research.}
ScienceBoard and Terminal-Bench-Science are the closest benchmark efforts to our setting. ScienceBoard offers a multimodal desktop environment with professional scientific applications and human-curated workflows \citep{sun2025scienceboard}; Terminal-Bench-Science evaluates command-line agents on expert-contributed research tasks using reproducible tests over concrete artifacts \citep{terminalbenchscience2026}. Moreover, SciAgentArena \citep{liu2026benchmarking} sets a standard for evaluating multi-step challenges across domains. \name{} is complementary: it centers task collection on the importance and difficulty judgments of practicing scientists, supports evaluation across diverse visual and structured outputs, and explicitly treats the agent harness as an object of controlled analysis and optimization.

%% file: methods.tex
\begin{figure}[h]
    \centering
    \includegraphics[width=1\linewidth]{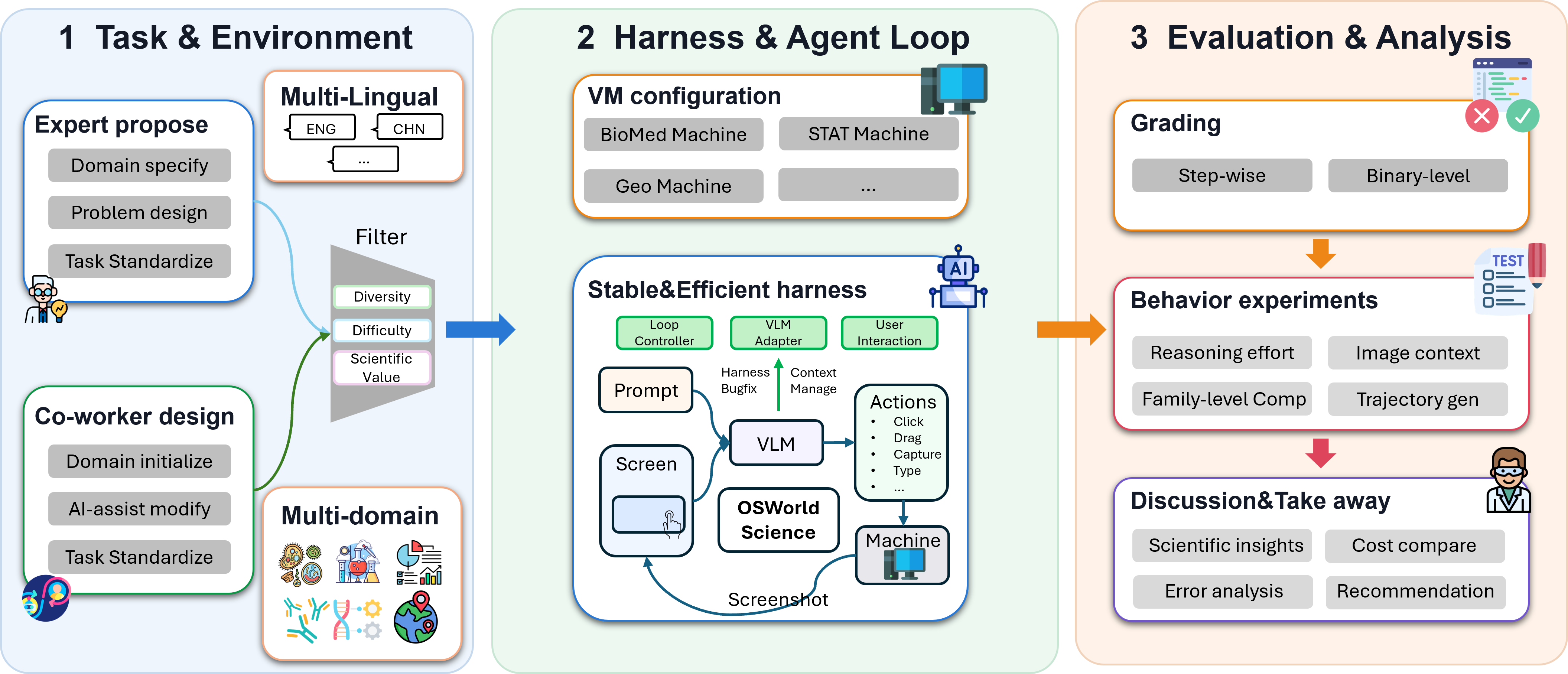}
    \caption{Landscape of task collection, environment setup, harness construction, and evaluation pipelines of OSWorld-Science.}
    \label{fig:overall}
\end{figure}

\textbf{Task definition.} 
Similar to OSWorld, we also formulate an autonomous scientific computer-use task in
\textsc{OSWorld-Science} as a goal-conditioned partially observable Markov
decision process (POMDP) \citep{xie2024osworld},
\[
\mathcal{M}
=
(\mathcal{S}, \mathcal{O}, \mathcal{A}, \mathcal{T},
\Omega, r, \gamma, \rho_0, \mathcal{G}, p_g),
\]
where $\mathcal{S}$ denotes the full environment state, including the
operating-system state, application-internal states, loaded scientific data,
intermediate analysis results, and task artifacts that may not be directly
visible to the agent. $\mathcal{O}$ is the observation space containing the
information accessible to the agent, such as the task instruction,
application screenshots, accessibility-tree information, visible numerical
results, and other interface-level observations. $\mathcal{A}$ denotes the
space of executable computer actions, and
\[
\mathcal{T}: \mathcal{S} \times \mathcal{A} \rightarrow \mathcal{S}
\]
is the environment transition function. The observation function
$\Omega$ maps the underlying state to the partial observation available to
the agent. The reward function
\[
r: \mathcal{S} \times \mathcal{A} \times \mathcal{G}
\rightarrow [0,1]
\]
measures the degree to which the resulting scientific state and artifacts
satisfy the task goal. Here, $\gamma$ is the discount factor, $\rho_0$ is
the initial-state distribution, $\mathcal{G}$ is the space of scientific
task goals, and $p_g$ denotes the distribution over task goals.

At interaction step $t$, the agent receives a goal $g \in \mathcal{G}$ and
a partial observation $o_t \in \mathcal{O}$,
\[
o_t = \Omega(s_t),
\]
which may include the natural-language scientific instruction together with
a screenshot, an accessibility tree, or their combination. Based on
$(g,o_t)$ and its interaction history, the agent produces an executable
action
\[
a_t \sim \pi(a_t \mid g, o_{\leq t}, a_{<t}),
\qquad a_t \in \mathcal{A}.
\]
Actions can include low-level GUI interactions, such as mouse clicks,
keyboard input, drag-and-drop operations, menu navigation, and keyboard
shortcuts. For example, an agent may execute
\texttt{click(300, 540, button='right')}. We also consider CLI actions, as long as the agent can ground and call the CLI settings in scientific softwares. These actions modify the scientific
software environment according to
\[
s_{t+1} = \mathcal{T}(s_t,a_t),
\qquad
o_{t+1} = \Omega(s_{t+1}).
\]

Unlike generic desktop tasks, scientific computer-use tasks frequently
require the agent to manipulate structured scientific objects and produce
quantitatively verifiable artifacts. Examples include loading and
registering medical images, placing anatomical landmarks, performing image
segmentation, configuring simulation parameters, manipulating molecular
structures, generating plots, or exporting analysis results. Consequently,
the task state may contain both visible GUI state and latent scientific
state, such as voxel coordinates, segmentation masks, transformation
matrices, measurement values, model parameters, or generated files. The interaction continues until the agent emits a terminal action
(\texttt{DONE}, or \texttt{FAIL}) or reaches the maximum interaction budget without valid outcomes (\texttt{VOID}).
\textsc{OSWorld-Science} uses execution-based task evaluators that directly
inspect the resulting environment state and generated artifacts. For a task
with goal $g$, the final reward is computed as the following variable
\[
R(\tau,g)
=
E_g(s_T, \mathcal{F}_T)
\in [0,1],
\]
where $\tau=(s_0,a_0,\ldots,s_T)$ is the execution trajectory,
$\mathcal{F}_T$ denotes the set of task-relevant output artifacts, and
$E_g$ is a goal-specific evaluator. A score of $1$ indicates that all
scientific requirements are satisfied, while a value between $0$ and $1$
represents partial completion according to task-specific criteria. A score
of $0$ is assigned when the required scientific outcome is not achieved. we have defined task-specific evaluators and explain them in our appendix in details.

\textbf{Task generation.}
We consider task design from three perspectives: multidisciplinary backgrounds, difficulty levels, and multilingual support. Therefore, our evaluation takes into account the diverse needs of researchers at different levels and is both comprehensive and instructive. In the current version, we consider seven domains: biology, medicine, chemistry, physics, geographic information, linguistics, and neuroscience, across three languages (English, Chinese, and Japanese). Our criteria for selecting tasks are based on their specific scientific value. The tasks included in the evaluation framework are problems of interest to scientists that require the use of software to solve; therefore, the set of tasks includes both those that can be solved using state-of-the-art models and those that cannot, thereby establishing a gradient of difficulty. 

We offer two approaches for designing tasks: 1. Expert propose and 2. Co-worker design. In the first approach, we recruit domain experts (Researchers with PhD-level knowledge in the selected field) and ask them to design their expected tasks with the template provided by our algorithm team. We will later remove some tasks based on the diversity of the tasks and the diversity of the software to ensure that we can evaluate the VLM's broader capabilities. Taking into account the limited experience and time available to some experts in certain fields, as well as the fact that there is some existing data in these fields that can be used as a reference, we have also designed a second approach. We introduce an AI-assisted Co-worker task design framework. By using this framework, we select anchor tasks (e.g. tutorial for a software), interact it with advanced AI softwares such as OpenAI Co-worker, propose new questions to address problems in the similar scenarios. We will interact and iterate with the Co-worker and treat it as a Co-Scientist and work together to refine the tasks. We found that by adopting this approach, we can still design high-quality tasks. 

\textbf{Environment setup.}
Considering the use cases for scientific software, we have chosen Ubuntu as the base image. To avoid the conflicts of different software, for each domain, we create a task-specific image by installing the required software and test its interaction with our base agent framework. In our benchmark, we consider 17 software in total, and ensure that each domain has their own preferred software being tested. Our evaluation also includes both open-source (e.g., QuPath) and closed-source (e.g. SAS) software. For closed-source software, we obtained the necessary licenses in advance and have committed to not using the trajectory data for agent training. The default resolution of our machine is 1920x1680. 

\textbf{VLM and harness setup.}
Here we consider both closed-source VLMs, including the GPT series (GPT-5.6 Luna, Terra, and Sol, and GPT-6 Astra)
\citep{openai2026gpt56luna,openai2026gpt56terra,openai2026gpt56sol,openai2026gpt6astra}, the Claude series (Sonnet 5, Opus 5, and Fable 5.1) \citep{anthropic2026sonnet5,anthropic2026opus5,anthropic2026fable51}, Qwen3.7-Plus \citep{alibaba2026qwen37plus}, and Gemini 3.1 Pro Preview \citep{google2026gemini31pro}; as well as open-weight VLMs, including MiniMax-M3 \citep{lai2026minimaxm3}, Kimi K3 \citep{kimi2026k3}, and Qwen-CUA \citep{lu2026qwencua}. To ensure a fair comparison, every model is driven by the same root agent: the multimodal PromptAgent (mmagent) of the OSWorld base release, ported into our harness with its message construction and prompting unchanged. At each step the agent receives the current 1920×1080 screenshot together with the last five (screenshot, response) pairs. The system prompt states the task, the screen resolution for coordinate grounding, and the machine's control. The agent replies with a short reflection and a single block of \texttt{pyautogui} code, or one of the special tokens \texttt{<WAIT>}, \texttt{<DONE>} and \texttt{<FAIL>}. The harness executes this code inside the guest VM, which gives the VLM direct control of the keyboard and mouse across GUI applications and the terminal. It then waits 1s and captures the next screenshot; \texttt{<WAIT>} pauses for 2s, and an unparsable reply is treated as \texttt{<WAIT>}. Every run uses the same budget: the per-task step limit (100 steps for most tasks), a 16k-token reply limit, and the provider's default sampling settings. Episodes end on these conditions: \texttt{<DONE>}, \texttt{<FAIL>}, the step limit, or an empty reply. For scoring, only the files and application state collected from the VM after the episode are graded. For every step the harness logs the model request and response, token usage, the executed code with its return code and stderr. These logs are used for cost accounting and failure analysis and are not fed back to the model. During development we fixed several reliability problems. Model calls are retried up to four times, with back-off and a doubled token limit when a reply is truncated, and a failed prediction is rolled back so the step's history stays consistent. A reply is capped at 10 executed actions, and for the self-hosted models (Qwen-CUA), we utilized the maximal waiting time to access a reliable response.

%% file: results.tex
\begin{figure}[h]
    \centering
    \includegraphics[width=1.0\textwidth]{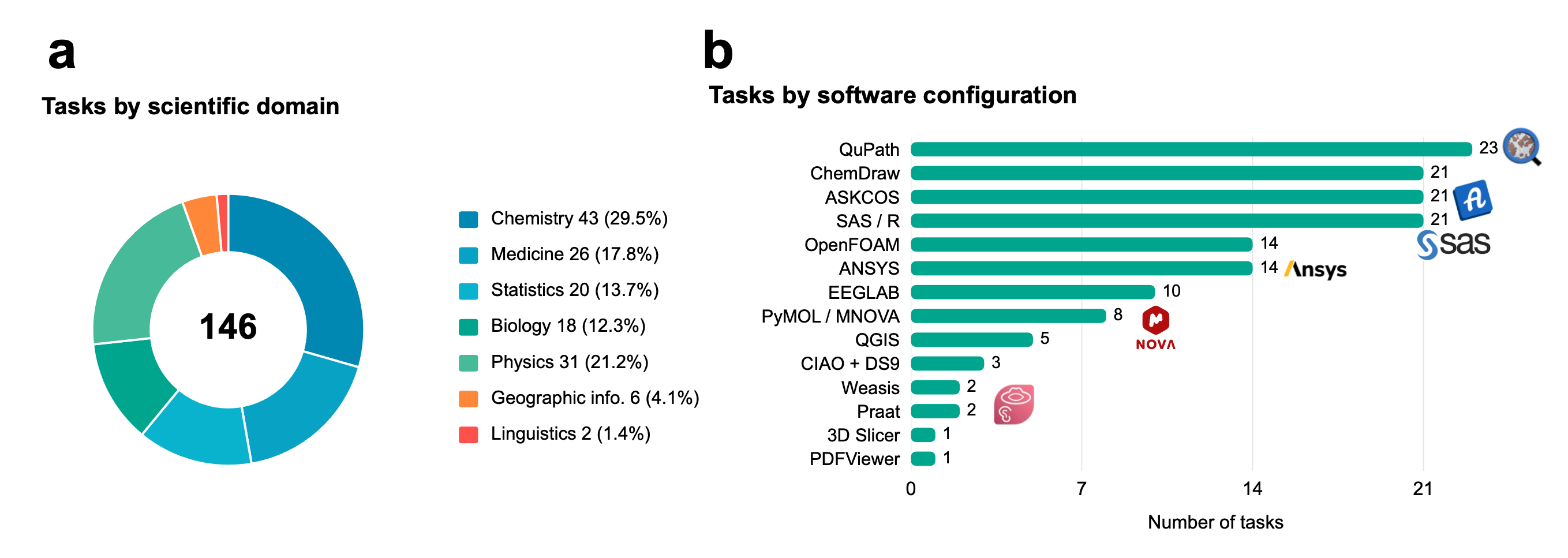}
    \caption{Composition of the collected OSWorld-Science task set. (a) Task shares by scientific domain. (b) Task counts by recorded software configuration. Panels (a) and (d) are normalized by the full task set. Slash and plus labels preserve paired configurations in the source inventory.}
    \label{fig:dataset-statistics}
\end{figure}
\textbf{Dataset overview.} The current OSWorld-Science task set contains 146 tasks spanning seven scientific domains and 14 recorded software configurations (Figures~\ref{fig:dataset-statistics} and \ref{fig:dataset-statistics-cont}). Chemistry contributes the largest share with 43 tasks (29.5\%), followed by physics with 31 (21.2\%), medicine with 26 (17.8\%), statistics with 20 (13.7\%), biology with 18 (12.3\%), geographic information with 6 (4.1\%), and linguistics with 2 (1.4\%). This uneven distribution reflects the current collection effort and motivates reporting both aggregate and domain-macro-averaged performance when comparing agents.

The task set covers a broad range of specialized scientific workflows. QuPath~\citep{bankhead2017qupath} accounts for 23 tasks (15.8\%), ChemDraw~\citep{evans2014chemdraw} and ASKCOS~\citep{tu2025askcos} for 21 each (14.4\% each), and SAS/R~\citep{sas_software,r_software} for 20 (13.7\%); together, these four configurations comprise 85 tasks (58.2\%). The remaining 61 tasks (41.8\%) span ten configurations: OpenFOAM~\citep{openfoam_software}, ANSYS~\citep{ansys_software}, EEGLAB~\citep{delorme2004eeglab}, PyMOL/MNOVA~\citep{pymol_software,mnova_software}, QGIS~\citep{qgis_software}, CIAO$+$DS9~\citep{fruscione2006ciao,joye2003ds9}, Weasis~\citep{roduit_weasis}, Praat~\citep{boersma2001praat}, 3D Slicer~\citep{fedorov2012slicer}, and PDFViewer. Across these environments, the annotations define ten distinct operation templates covering document extraction and table analysis, molecular reasoning and drawing, pathology and radiology inspection, simulation and statistical analysis, GIS digitization, and sound/EEG analysis. Our scoring method is determined by the question and is normalized to a 0–1 range as the pass rate.

The task annotations also characterize the benchmark's interaction demands. Under the benchmark's visual-observation protocol, all 146 tasks require screenshot-based observation of the scientific application state; explicit screenshot capture is separately listed in the operation annotations for 21 tasks (14.4\%). Click operations appear in 124 tasks (84.9\%), text entry in 101 (69.2\%), drawing in 21 (14.4\%). These capability counts are non-exclusive because a task may require several primitives. CLI access is supported for 122 tasks (83.6\%), whereas 24 tasks (16.4\%) are GUI-only under the recorded configurations. This mix supports evaluation of both direct GUI control and hybrid GUI-CLI interaction in workflows that require visual inspection of scientific application state. Details of experiment setup are shown in Appendix \ref{sec:model setting}.

\begin{figure}[ht]
    \centering
    \includegraphics[width=1\linewidth]{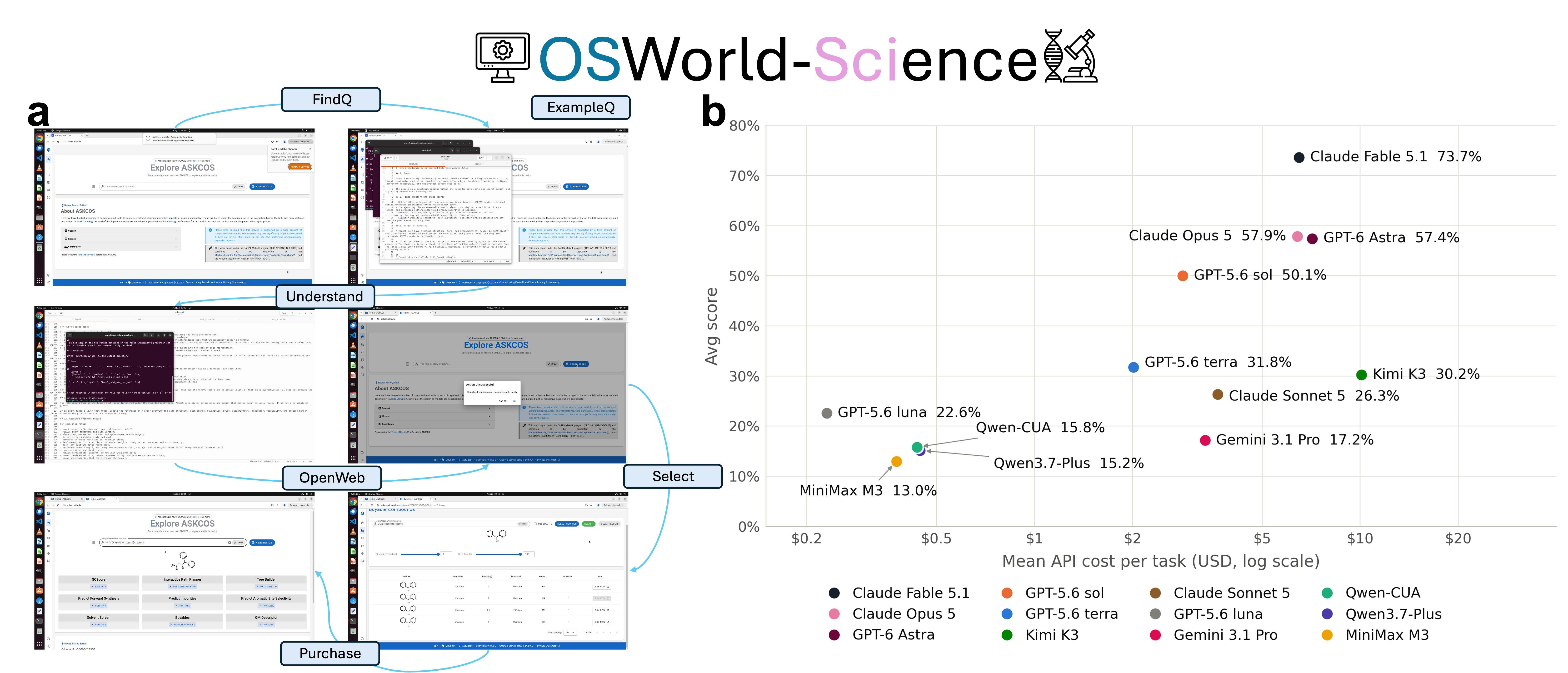}
    \caption{OSWorld-Science example workflow and overall performance-cost comparison. (a) An agent uses ASKCOS to address a scientific task. (b) Mean task-specific score, including partial credit and with \texttt{VOID} runs scored zero, versus mean API cost per task in U.S. dollars (logarithmic scale) across 12 evaluated VLMs.}
    \label{fig:overall_score}
\end{figure}

\textbf{Results interpretation.} Figure~\ref{fig:overall_score}(a) illustrates how the agent uses software (e.g. ASKCOS) to address a scientific problem, while Figure~\ref{fig:overall_score}(b) compares overall performance and API cost across 12 VLMs. Models use their default reasoning effort under the evaluation settings described in Appendix~\ref{sec:model setting}, following the OSWorld evaluation framework~\citep{yuan2026osworld2}. Performance is the mean task-specific score, including partial credit, with \texttt{VOID} runs assigned zero. Claude Fable~5.1 achieves the highest mean score, 73.7\%, at \$6.5 per task, compared with 57.9\% at \$6.7 for Claude Opus~5 and 57.4\% at \$7.7 for GPT-6 Astra. GPT-5.6 sol offers an intermediate trade-off, scoring 50.1\% at \$2.79 per task, followed by GPT-5.6 terra at 31.8\% and \$2.0. At the low-cost end, GPT-5.6 luna reaches 22.6\% at \$0.23 per task, close to Claude Sonnet~5 (26.3\% at \$3.6) and Kimi K3 (30.2\% at \$11.7), whose costs are approximately 16 and 51 times higher, respectively. Appendix~\ref{sec:app-performance-cost} provides the complementary output-token comparison (Figure~\ref{fig:app-output-tokens}) and separate overall cost and score rankings (Figures~\ref{fig:app-overall-cost} and~\ref{fig:app-overall-score}). Overall, strong and closed-source models occupy the top tier of the ecosystem, while they still cannot perfectly address all problems (Additional ablation studies in Appendix \ref{sec:app-ablation}).

The domain breakdown (Figure~\ref{fig:app-domain-cost}) shows substantial variation in difficulty and model rankings. Among the models plotted for statistics, mean scores range from approximately 66\% to 97\%, with GPT-5.6 luna reaching 88\% at about \$0.06 per task. Chemistry is much harder: Fable~5.1 and Opus~5 lead at approximately 44\% and 39\%, GPT-5.6 sol reaches 20\%, and eight models score approximately 6\% or less. EEG performance is highly uneven: Claude Fable 5.1 reaches 89\%, Astra 54\%, and Claude Opus~5 33\%, while all remaining models score 15\% or less. Astra leads medical imaging and geoscience at approximately 69\% in each, whereas Fable~5.1 reaches approximately 96\% in biology. Rankings also change sharply across domains: Astra scores about 6\% in chemistry but 100\% on the two linguistics tasks. Therefore, different models also have their own areas of strength, and natural science is still a hard domain to release the full power of CUA.



\textbf{Trajectory analysis.}
We label each of the $\sim$1500 released runs by its graded artifact: full credit, wrong content, invalid artifact, no deliverable, or ended by the harness or environment. Interaction style is measured with one definition across software: GUI and CLI step shares, narration, thinking share, and repeated replies (Appendix~\ref{sec:app-traj-all}).

\begin{figure}[ht]
    \centering
    \includegraphics[width=\textwidth]{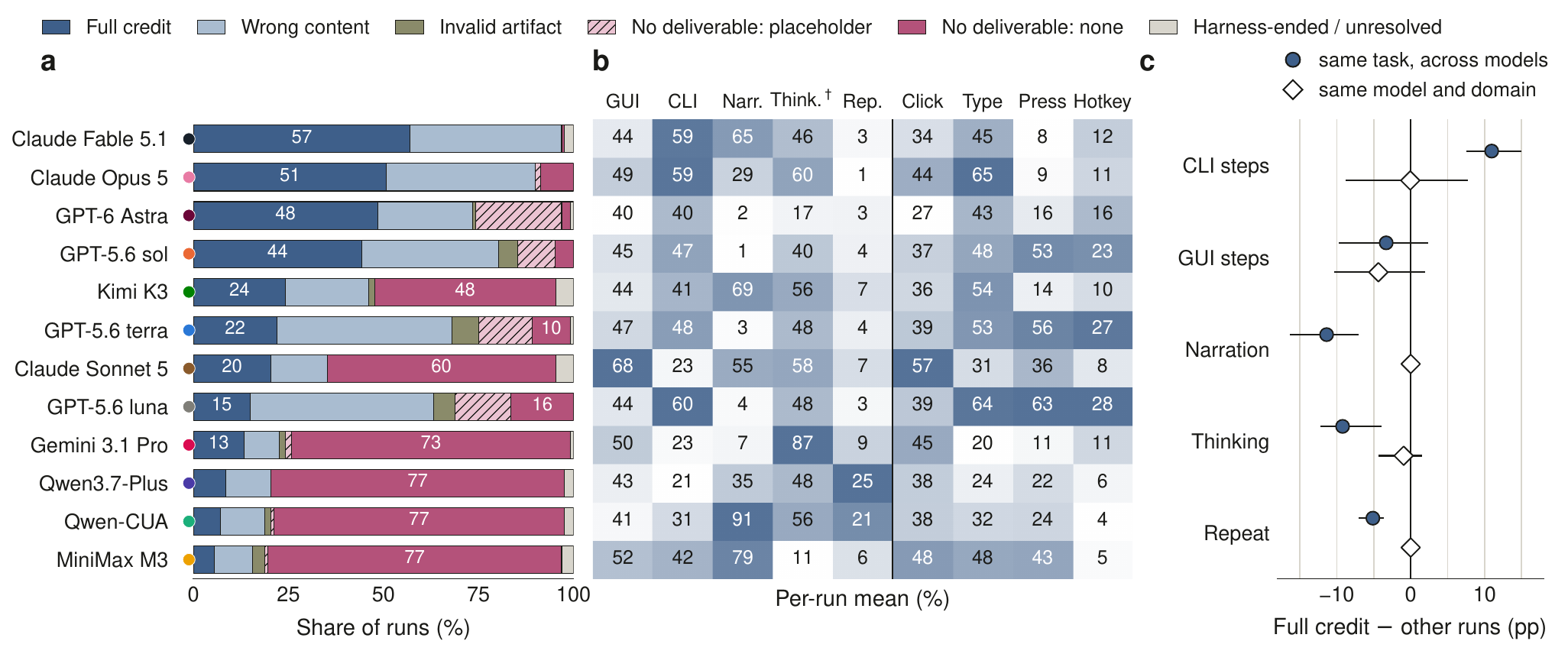}
    \caption{Trajectory analysis of the 1{,}530 released runs. (a) Outcome by model. (b) Per-run means (\%) of the interaction measures and of steps containing four common automation calls ($^\dagger$chemistry and runs without recorded thinking excluded); shading is normalized per column. (c) Full-credit minus other runs: median over tasks (circles; 86 tasks, 69 for thinking) or over model-domain groups (diamonds; 39 groups, 34 for thinking), with bootstrap 95\% intervals.}
    \label{fig:trajectory}
\end{figure}

\emph{Most failures produce no answer artifact.} Only 402 runs (26.3\%) receive full credit. Of the other 1{,}100 runs (excluding 20 harness-ended and 8 unresolved runs), 667 (60.6\%) end without the graded artifact, against 400 with wrong content and 33 with an invalid one, and 613 of the 667 exhaust their budget. Models separate on this axis (Figure~\ref{fig:trajectory}a). Claude Fable~5.1, Claude Opus~5, GPT-6 Astra, and GPT-5.6 sol deliver nothing in 1-25\% of their runs; Claude Sonnet~5, Gemini~3.1 Pro, Qwen3.7-Plus, Qwen-CUA, and MiniMax~M3 deliver nothing in 60-78\%. The GPT models produce 78 of the 84 placeholder-only artifacts, 77 of them in chemistry, where they typically write an empty skeleton and run out of budget before filling it. Budget exhaustion is seldom a literal loop: only 15\% of the budget-exhausted runs without a deliverable repeat at least 30\% of their replies, and 84\% of those come from the two Qwen models.

\emph{Narration and thinking are model signatures, not predictors of success.} GUI share depends mainly on the software (29\% of steps in OpenFOAM, 90\% in Praat). Narration and thinking are stable across software for each model (mean rank correlation 0.74 and 0.69, against 0.14 for GUI share; Figure~\ref{fig:trajectory}b). The GPT models narrate in under 5\% of their replies and Qwen-CUA in 91\%; Gemini~3.1 Pro spends 87\% of its output tokens on thinking and MiniMax~M3 11\%. Within a task, full-credit runs devote more of their steps to CLI (median $+11$ points, in 83\% of 86 tasks) and narrate, think, and repeat less. These differences vanish among runs of the same model in the same domain, where all 95\% intervals include zero (Figure~\ref{fig:trajectory}c). With model and task fixed effects, no measure predicts full credit after Bonferroni correction. The associations thus reflect which models succeed rather than how a given model interacts.

\begin{figure}[h]
    \centering
    \includegraphics[width=\textwidth]{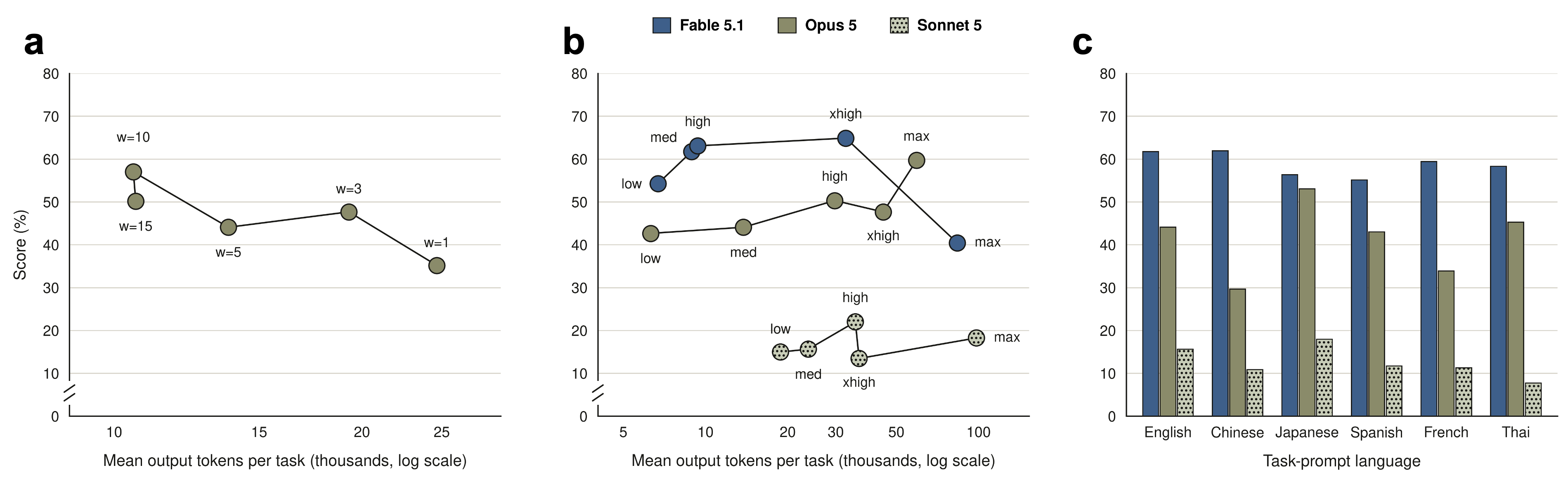}
    \caption{Ablations on medical subset, the 23 QuPath tasks; one run per cell, \texttt{VOID} runs scored 0, output tokens include thinking tokens. Top left: reasoning-effort sweep at $w=5$. Top right: history-window sweep for Opus~5 at medium effort. Bottom: task-prompt language at medium effort and $w=5$; the English bars are the medium cells of the effort sweep. No cell differs significantly from its default under a paired sign-flip test after Bonferroni correction over the 22 non-default cells (smallest uncorrected $p=0.04$, Opus~5 with French prompts); lines follow sweep order and are visual guides only.}
    \label{fig:ablation}
\end{figure}

\textbf{Ablation studies.} We do ablation studies on three main factors of our GUI agent configuration on QuPath-Bench from the medicine domain, which includes 23 QuPath pathology tasks (15.8\% of the task set, the largest single-software group, graded with partial credit): the reasoning-effort setting of the backbone, the history window length $w$ (the number of most recent screenshots), and the language of the task (Figure~\ref{fig:ablation} and Table~\ref{tab:ablation}). Claude Opus~5 and Sonnet~5 serve as backbones of the ablation experiments. Each cell is a single run per task ($n=1$), and all three sweeps share the default cell (medium effort, $w=5$, English prompts) if not mentioned specifically, which is also the QuPath entry of the main results (44.1\% for Opus~5 and 15.7\% for Sonnet~5).

\emph{Reasoning effort.} From low to max, mean output tokens per task grow about tenfold for Opus~5 (6k to 59k) and fivefold for Sonnet~5 (19k to 98k). Opus~5 rises from 44.1\% (medium) to 59.7\% (max), but only high and max add fully solved tasks (binary 34.8\% at low, medium, and xhigh; 43.5\% at high; 52.2\% at max): from medium to max, eight tasks improve, five regress, and ten are unchanged, eight of which score the same at every effort level. Sonnet~5 is non-monotone and never exceeds 22.0\%, which it reaches at high rather than at max effort (18.3\%). Sonnet~5 also spends more output tokens than Opus~5 at every level except xhigh while solving far fewer tasks. Thus, effort is an expensive lever for our Opus~5 and Sonnet~5 backbones: relative to low, max costs roughly five to ten times the output tokens and gains 17.0 points for Opus~5 but only 3.3 points for Sonnet~5.

\emph{History window.} The window acts on the number of steps at a constant price per step: the mean trajectory shortens from 52 steps at $w=1$ to 22 at $w=10$, so output tokens fall from 25k to 11k for the same reason and the score rises from 35.2\% to 57.0\%. Moreover, \texttt{VOID} runs disappear from $w=10$ on, and unsolved runs shrink from 62 steps at $w=1$ to 23 at $w=10$: with enough history the agent stops repeating actions it can no longer see and abandons hopeless runs instead of retrying them. Beyond ten screenshots there is no further benefit: $w=15$ uses the same output tokens as $w=10$ and scores lower.

\emph{Task language.} Prompt language changes Opus~5's interaction style more than its reasoning: Chinese and Japanese prompts leave the scripting strategy and the trajectory length intact, whereas Spanish, French, and Thai prompts shift the model toward GUI manipulation and lengthen trajectories from 27 to 36-44 steps. It also changes the model's working language (Table~\ref{tab:response-language}): Opus~5 writes its prose replies in the prompt language for Chinese (89\%) and Japanese (65\%), partly for Spanish (43\%), and in English for French and Thai, whereas Sonnet~5 replies in English whatever the prompt language, and neither backbone drifts into a third language. Both backbones score highest with Japanese prompts and below English with Chinese ones, and Sonnet~5 has a median trajectory of 100 steps in every language, which leaves prompt language little room to matter.

\begin{figure*}[ht]
    \centering
    \includegraphics[width=0.9\textwidth]{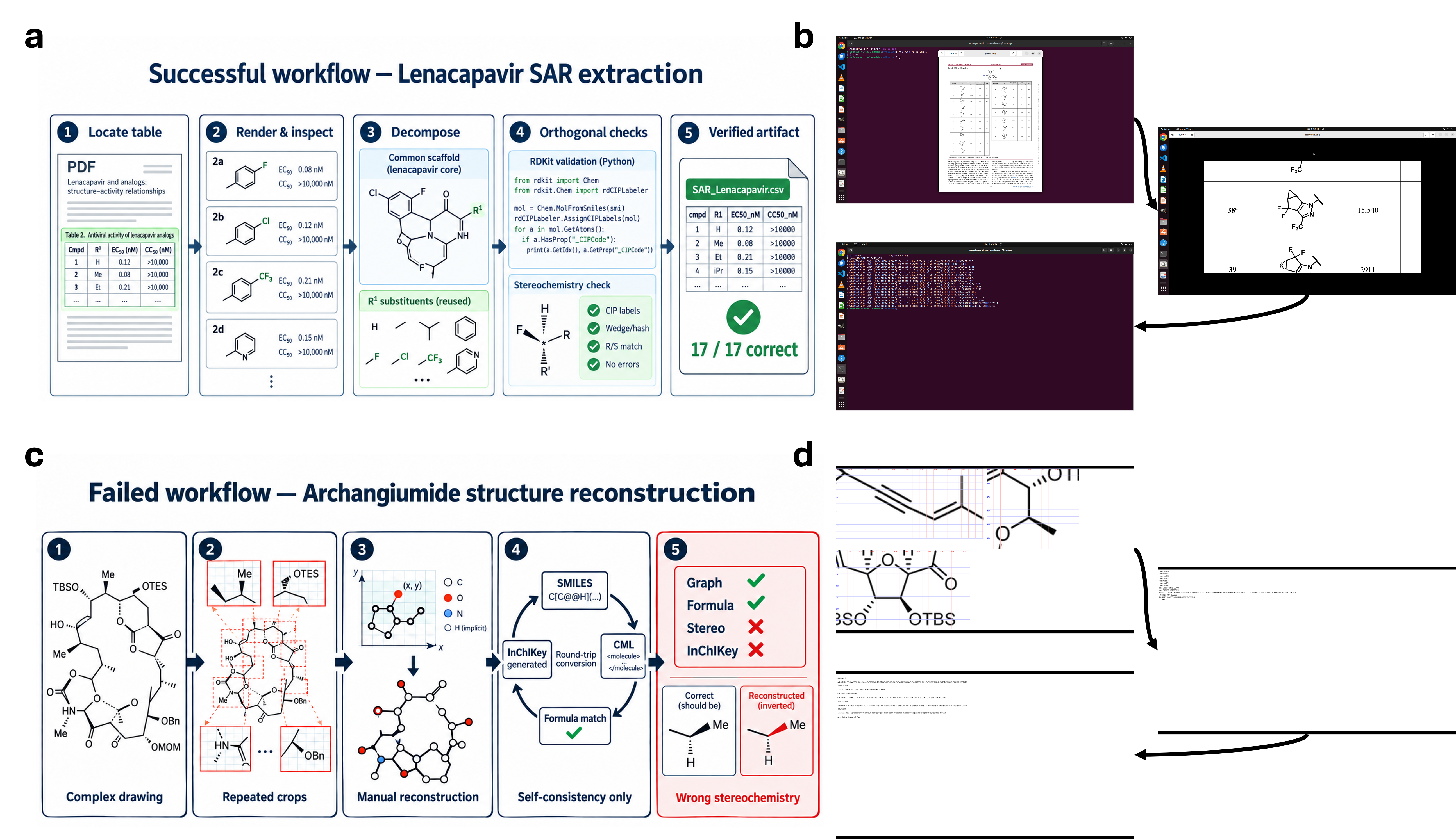}
    \caption{Contrasting Opus~5 chemistry workflows with representative raw trajectory frames shown directly below each workflow diagram. (a) and (b): the successful Lenacapavir run moves from source-table localization to focused stereochemical inspection and final validation of the complete 17-row CSV. (c) and (d): the failed Archangiumide run converts coordinate-guided reconstruction into a chemically valid, internally consistent structure, yet its own round-trip check still reveals a stereochemical mismatch. The decisive difference is independent source-grounded validation rather than trajectory length or the number of checks.}
    \label{fig:opus5-case-studies}
\end{figure*}

\textbf{Selective case study.} We contrast two Opus~5 chemistry trajectories with nearly identical horizons (Figure~\ref{fig:opus5-case-studies}) but opposite outcomes. The comparison shows that success depends less on the number of interaction steps than on whether the agent preserves the task's semantic structure and validates uncertain decisions with evidence independent of its own reconstruction. More case studies across domains can be found in our appendices \ref{sec:app-traj-all}.

\emph{Successful trajectory.} On the Lenacapavir SAR task, Opus~5 localized Table~2 with \texttt{pdftotext}, progressively rendered the relevant page at 200-1200~dpi, and separated a shared molecular scaffold from the 17 variable $R^1$ substituents. This decomposition let it encode the invariant scaffold once rather than redraw each ligand independently. Before writing the CSV, it cross-checked the assay column and table footnotes, parsed every SMILES with RDKit, and explicitly verified the intended CIP assignments for the common benzylic center and the stereodefined fused rings. The evaluator found all 17 rows fully correct (score~1.0 after 70 steps). The productive pattern was iterative but structured: each new crop resolved a named uncertainty, and the final artifact was checked against row count, schema, chemical parsability, and stereochemical constraints.

\emph{Failed trajectory.} On the Archangiumide structure-reconstruction task, Opus~5 spent most of 68 steps generating enlarged crops, coordinate grids, and hand-mapped atom/bond coordinates for a densely substituted macrocycle. It committed to a molecular graph only near the end, then validated that same hypothesis by round-tripping its SMILES through CML/Open Babel and recomputing the formula and InChIKey. These checks established internal consistency, but they were not independent of the original wedge-to-configuration interpretation. The evaluator consequently found the molecular graph and formula correct but the stereochemistry and InChIKey wrong, producing a binary score of~0.0. Once an inverted stereocenter entered the reconstruction, every downstream conversion faithfully reproduced the same error.

\emph{Takeaway.} Long trajectories and numerous validation actions are therefore insufficient when all checks share one latent assumption. For stereo-heavy scientific workflows, agents should maintain explicit uncertainty at each stereocenter and compare wedge/hash geometry and derived CIP labels directly against the source (or an independent reference) before canonicalization. Orthogonal validation, rather than self-consistency alone, is what separates the two cases. 

%% file: ablation_appendix.tex

\setcounter{topnumber}{2}
\setcounter{bottomnumber}{1}
\setcounter{totalnumber}{3}
\renewcommand{\topfraction}{0.85}
\renewcommand{\bottomfraction}{0.5}
\renewcommand{\textfraction}{0.1}
\renewcommand{\floatpagefraction}{0.75}
\makeatletter
\setlength{\@fptop}{0pt}
\makeatother

\section{Experimental details}
\label{sec:model setting}

\textbf{Implementation details.}
The experiments conducted in this paper followed OSWorld’s original design, which allows models to determine actions by analyzing screenshots and trajectories. With the exception of Kimi-K3, all VLMs utilized our modified version of mmagent: an agent designed to operate scientific software, to solve the problems. For Kimi-K3, due to certain specific requirements regarding the model’s adaptation to the tool, we adopted the officially provided design.

\textbf{Grading details.}
Our problems can generally be divided into two categories. The first category consists of problems with a single, definitive solution—such as drawing a molecular structure or counting the number of airplanes on a map. Scoring for these problems is based on exact matches; different problems may award partial credit for steps, so the final pass rate serves as an average metric. The second category consists of problems involving the results of computational simulations, such as regression parameters in statistics and models in astrophysics. For these problems, scoring is determined by the discrepancy between the parameters generated by the model and the actual parameters; if the error in the parameters falls within a certain range, we consider the model to have successfully solved the problem.

\input{performance_appendix}

\section{Additional ablation analyses}
\label{sec:app-ablation}

This appendix supplements the ablation studies of the main text with performance and trajectory statistics for every cell (Table~\ref{tab:ablation-traj}), the composition of run outcomes (Table~\ref{tab:ablation-outcomes}), and per-task scores for every cell (Tables~\ref{tab:pertask-effort}-\ref{tab:pertask-language}). All cells use Claude Opus~5 or Claude Sonnet~5 on the 23 QuPath tasks with one run per task and a limit of 100 interaction steps.

\paragraph{Steps and outcomes.} Every \texttt{VOID} run in the three sweeps used all 100 steps, and the runs that use all 100 steps but still leave a gradable state mostly score 0 or 0.1-0.2, and never more than 0.67. For Opus~5, runs that finish within 20 steps average 57\%, runs of 21-50 steps 44\%, and runs of 100 steps 3\%. Sonnet~5 uses all 100 steps in 60\% of its runs (139 of 230); these runs average 3\%, against 32\% for the runs that finish earlier. The per-task tables show that the same handful of long, GUI-heavy tasks sits at the limit in every sweep and flips between \texttt{VOID} and full credit depending on whether it finishes within 100 steps.

\paragraph{Reasoning effort.} Table~\ref{tab:ablation-traj} decomposes the token growth. For Opus~5 the thinking share rises from 0.40 to 0.75 and tokens per step from 281 to 1,346, while the mean trajectory doubles from 22 to 44 steps (13 to 30 steps on solved tasks). The CLI share stays at 0.78-0.84, so the extra steps are GUI inspection added to a scripting strategy: the GUI share rises from 0.20 at low to 0.45 at high. From xhigh on, Opus~5 also starts to narrate its actions (prose in 60-67\% of replies, against 8-16\% at lower effort). Per task (Table~\ref{tab:pertask-effort}), the max setting newly solves five tasks, two of which are GUI-heavy tasks that return \texttt{VOID} at medium and finish in 73 and 79 steps at max, and regresses on five, including a task that medium effort solves with a 16-step script and that turns into a 100-step GUI exploration (GUI share 0.6-0.8) at high effort and above. Sonnet~5's trajectories sit at or near the limit from medium effort on (median 93-100 steps; 4-8 \texttt{VOID} runs at every level), its GUI share is 0.60-0.90 against a CLI share of 0.21-0.38, and it narrates 58-100\% of its replies. Effort raises its tokens per step from 273 to 1,082 while the mean number of steps changes little (69 at low, 83-91 from medium on), so the token count grows while the outcome composition barely moves (2-4 solved tasks).

\paragraph{History window.} Tokens per step (468-531) and the thinking share (0.54-0.59) are flat across $w$, so the token savings come entirely from shorter trajectories: 52, 36, 27, 22, and 23 mean steps for $w=1$, 3, 5, 10, and 15, with 3, 3, 2, 0, and 0 \texttt{VOID} runs. Unsolved runs shorten from 62 steps at $w=1$ to 23-24 at $w\ge10$ and solved runs from 26 steps at $w=1$ to 14-21 at $w\ge3$, and the GUI share falls from 0.45 to 0.29 at $w=10$. The two long GUI-heavy tasks that return \texttt{VOID} at every $w\le5$ finish in 43-57 steps at $w\ge10$ (Table~\ref{tab:pertask-window}). $w=15$ matches $w=10$ in steps and tokens but loses ground on five tasks while gaining on three.

\paragraph{Task language.} For Opus~5, Chinese and Japanese prompts keep the CLI share at 0.82-0.83 and the trajectory at 29-32 steps (27 in English), whereas Spanish, French, and Thai prompts lower the CLI share to 0.70-0.74, raise the GUI share to 0.53-0.62, and lengthen trajectories to 36-44 steps; tokens per step fall from 510 to 281-441, so total tokens stay within 10-17k. The reply-language shares of Table~\ref{tab:response-language} in the main text rest on 28-65 prose replies per cell for Opus~5, which otherwise replies with code only, and on 1,300-1,660 for Sonnet~5. Per task (Table~\ref{tab:pertask-language}), the two long GUI-heavy tasks flip between \texttt{VOID} and full credit across languages and account for much of the spread between languages. Of the four tasks that the English run solves and that fail under Chinese prompts, three are solved in English with short trajectories (12-27 steps) and fail under Chinese prompts in equally short runs (13-25 steps), that is, with a wrong result rather than by running out of steps; the fourth (QP\_004\_T3) returns \texttt{VOID}. Sonnet~5 has a median of 100 steps in every language and 710 \texttt{VOID} runs; Thai prompts produce its shortest trajectories (75 steps) and its lowest score (7.7\%), so it stops earlier without finishing.

\begin{table}[t]
    \centering
    \footnotesize
    \setlength{\tabcolsep}{5pt}
    \renewcommand{\arraystretch}{1.05}
    \caption{Performance and trajectory statistics for every ablation cell (23 QuPath tasks, one run per cell, step limit 100). Partial, Binary, Steps, and \#Tok.\ as in Table~\ref{tab:ablation}. Tok./step: mean output tokens per task divided by mean steps. Think: share of output tokens that are thinking tokens. GUI and CLI: share of action steps that contain a mouse action or a terminal/script command, respectively (a step can count as both). Prose: share of model replies that contain at least 20 characters of natural-language text outside code blocks and control tokens. Shaded rows are the shared default configuration (medium effort, $w=5$, English prompts), a single run per backbone that appears in every block.}
    \label{tab:ablation-traj}
    \begin{tabular}{@{}lccccccccc@{}}
        \toprule
        Setting & Partial & Binary & Steps & \#Tok. & Tok./step & Think & GUI & CLI & Prose \\
        \midrule
        \multicolumn{10}{@{}l}{\textbf{Opus~5}} \\
        \multicolumn{10}{@{}l}{\quad \textit{Reasoning effort} ($w=5$, English prompts)} \\
        \quad low & 42.7 & 34.8 & 22 & 6.3 & 281 & 0.40 & 0.20 & 0.84 & 0.08 \\
        \rowcolor[gray]{0.92}
        \quad medium & 44.1 & 34.8 & 27 & 13.8 & 510 & 0.57 & 0.37 & 0.84 & 0.15 \\
        \quad high & 50.3 & 43.5 & 43 & 29.7 & 699 & 0.68 & 0.45 & 0.78 & 0.16 \\
        \quad xhigh & 47.7 & 34.8 & 42 & 44.8 & 1061 & 0.70 & 0.46 & 0.83 & 0.67 \\
        \quad max & 59.7 & 52.2 & 44 & 59.3 & 1346 & 0.75 & 0.41 & 0.82 & 0.60 \\
        \multicolumn{10}{@{}l}{\quad \textit{History window} (medium effort, English prompts)} \\
        \quad $w=1$ & 35.2 & 26.1 & 52 & 24.7 & 471 & 0.59 & 0.45 & 0.82 & 0.06 \\
        \quad $w=3$ & 47.7 & 39.1 & 36 & 19.3 & 531 & 0.58 & 0.41 & 0.84 & 0.10 \\
        \rowcolor[gray]{0.92}
        \quad $w=5$ & 44.1 & 34.8 & 27 & 13.8 & 510 & 0.57 & 0.37 & 0.84 & 0.15 \\
        \quad $w=10$ & 57.0 & 43.5 & 22 & 10.6 & 484 & 0.55 & 0.29 & 0.80 & 0.20 \\
        \quad $w=15$ & 50.2 & 43.5 & 23 & 10.6 & 468 & 0.54 & 0.38 & 0.78 & 0.12 \\
        \multicolumn{10}{@{}l}{\quad \textit{Task-prompt language} (medium effort, $w=5$)} \\
        \rowcolor[gray]{0.92}
        \quad English & 44.1 & 34.8 & 27 & 13.8 & 510 & 0.57 & 0.37 & 0.84 & 0.15 \\
        \quad Chinese & 29.7 & 21.7 & 29 & 11.1 & 386 & 0.52 & 0.49 & 0.82 & 0.14 \\
        \quad Japanese & 53.1 & 43.5 & 32 & 14.0 & 441 & 0.55 & 0.39 & 0.83 & 0.09 \\
        \quad Spanish & 43.0 & 39.1 & 44 & 16.0 & 361 & 0.57 & 0.53 & 0.71 & 0.03 \\
        \quad French & 33.9 & 26.1 & 42 & 16.5 & 393 & 0.55 & 0.53 & 0.74 & 0.05 \\
        \quad Thai & 45.3 & 39.1 & 36 & 10.1 & 281 & 0.55 & 0.62 & 0.70 & 0.06 \\
        \midrule
        \multicolumn{10}{@{}l}{\textbf{Sonnet~5}} \\
        \multicolumn{10}{@{}l}{\quad \textit{Reasoning effort} ($w=5$, English prompts)} \\
        \quad low & 15.0 & 8.7 & 69 & 18.8 & 273 & 0.46 & 0.60 & 0.38 & 0.58 \\
        \rowcolor[gray]{0.92}
        \quad medium & 15.7 & 13.0 & 87 & 23.8 & 275 & 0.55 & 0.81 & 0.21 & 0.81 \\
        \quad high & 22.0 & 17.4 & 83 & 35.4 & 429 & 0.62 & 0.90 & 0.25 & 0.94 \\
        \quad xhigh & 13.5 & 8.7 & 85 & 36.5 & 429 & 0.65 & 0.80 & 0.21 & 0.97 \\
        \quad max & 18.3 & 13.0 & 91 & 98.2 & 1082 & 0.77 & 0.90 & 0.28 & 1.00 \\
        \multicolumn{10}{@{}l}{\quad \textit{Task-prompt language} (medium effort, $w=5$)} \\
        \rowcolor[gray]{0.92}
        \quad English & 15.7 & 13.0 & 87 & 23.8 & 275 & 0.55 & 0.81 & 0.21 & 0.81 \\
        \quad Chinese & 10.9 & 8.7 & 89 & 26.1 & 295 & 0.56 & 0.79 & 0.23 & 0.78 \\
        \quad Japanese & 18.0 & 13.0 & 86 & 28.0 & 326 & 0.56 & 0.79 & 0.24 & 0.85 \\
        \quad Spanish & 11.7 & 8.7 & 86 & 21.3 & 248 & 0.57 & 0.80 & 0.21 & 0.78 \\
        \quad French & 11.3 & 8.7 & 88 & 23.5 & 268 & 0.54 & 0.79 & 0.25 & 0.75 \\
        \quad Thai & 7.7 & 4.3 & 75 & 28.4 & 378 & 0.60 & 0.75 & 0.38 & 0.73 \\
        \bottomrule
    \end{tabular}
\end{table}

\begin{table}[t]
    \centering
    \footnotesize
    \setlength{\tabcolsep}{4pt}
    \renewcommand{\arraystretch}{1.05}
    \caption{Outcome composition of every ablation cell. Score: mean partial-credit score (\%), as in Table~\ref{tab:ablation}; the remaining columns count tasks out of 23. Solved: score 1. Partial: score strictly between 0 and 1. Zero: score 0 with a gradable final state. \texttt{VOID}: no gradable outcome. Shaded rows are the shared default configuration.}
    \label{tab:ablation-outcomes}
    \begin{tabular}{@{}lcccccccccc@{}}
        \toprule
        & \multicolumn{5}{c}{Opus~5} & \multicolumn{5}{c}{Sonnet~5} \\
        \cmidrule(lr){2-6} \cmidrule(lr){7-11}
        Setting & Score & Solved & Partial & Zero & \texttt{VOID} & Score & Solved & Partial & Zero & \texttt{VOID} \\
        \midrule
        \multicolumn{11}{@{}l}{\textit{Reasoning effort} ($w=5$, English prompts)} \\
        low & 42.7 & 8 & 7 & 6 & 2 & 15.0 & 2 & 4 & 11 & 6 \\
        \rowcolor[gray]{0.92}
        medium & 44.1 & 8 & 8 & 5 & 2 & 15.7 & 3 & 4 & 8 & 8 \\
        high & 50.3 & 10 & 4 & 5 & 4 & 22.0 & 4 & 7 & 6 & 6 \\
        xhigh & 47.7 & 8 & 6 & 6 & 3 & 13.5 & 2 & 6 & 11 & 4 \\
        max & 59.7 & 12 & 5 & 3 & 3 & 18.3 & 3 & 6 & 6 & 8 \\
        \midrule
        \multicolumn{11}{@{}l}{\textit{History window} (medium effort, English prompts)} \\
        $w=1$ & 35.2 & 6 & 8 & 6 & 3 & \multicolumn{5}{c}{--} \\
        $w=3$ & 47.7 & 9 & 7 & 4 & 3 & \multicolumn{5}{c}{--} \\
        \rowcolor[gray]{0.92}
        $w=5$ & 44.1 & 8 & 8 & 5 & 2 & \multicolumn{5}{c}{--} \\
        $w=10$ & 57.0 & 10 & 8 & 5 & 0 & \multicolumn{5}{c}{--} \\
        $w=15$ & 50.2 & 10 & 6 & 7 & 0 & \multicolumn{5}{c}{--} \\
        \midrule
        \multicolumn{11}{@{}l}{\textit{Task-prompt language} (medium effort, $w=5$)} \\
        \rowcolor[gray]{0.92}
        English & 44.1 & 8 & 8 & 5 & 2 & 15.7 & 3 & 4 & 8 & 8 \\
        Chinese & 29.7 & 5 & 8 & 8 & 2 & 10.9 & 2 & 3 & 9 & 9 \\
        Japanese & 53.1 & 10 & 8 & 4 & 1 & 18.0 & 3 & 6 & 6 & 8 \\
        Spanish & 43.0 & 9 & 4 & 8 & 2 & 11.7 & 2 & 4 & 7 & 10 \\
        French & 33.9 & 6 & 8 & 6 & 3 & 11.3 & 2 & 4 & 9 & 8 \\
        Thai & 45.3 & 9 & 6 & 7 & 1 & 7.7 & 1 & 5 & 10 & 7 \\
        \bottomrule
    \end{tabular}
\end{table}

\begin{table}[t]
    \centering
    \scriptsize
    \setlength{\tabcolsep}{3pt}
    \renewcommand{\arraystretch}{1.02}
    \caption{Per-task scores in the reasoning-effort sweep ($w=5$, English prompts). V: \texttt{VOID} run (no gradable outcome). The bottom rows give the mean score with \texttt{VOID} scored 0, the number of fully solved tasks, and the number of \texttt{VOID} runs.}
    \label{tab:pertask-effort}
    \begin{tabular}{@{}lcccccccccc@{}}
        \toprule
        & \multicolumn{5}{c}{Opus~5} & \multicolumn{5}{c}{Sonnet~5} \\
        \cmidrule(lr){2-6} \cmidrule(lr){7-11}
        Task & low & medium & high & xhigh & max & low & medium & high & xhigh & max \\
        \midrule
        QP\_001\_T1-1 & 1.00 & 0.00 & 0.00 & 0.00 & 1.00 & 0.00 & 0.00 & 0.00 & 0.00 & 0.00 \\
        QP\_001\_T1-2 & 1.00 & 1.00 & 1.00 & 0.00 & 1.00 & 0.00 & 0.00 & 1.00 & 0.00 & 1.00 \\
        QP\_001\_T2-1 & 0.08 & 0.03 & 0.00 & 0.00 & 0.10 & 0.00 & V & V & 0.00 & V \\
        QP\_001\_T2-2 & 0.27 & 0.23 & V & V & V & 0.00 & V & V & V & V \\
        QP\_001\_T3 & 0.00 & 0.00 & 0.58 & 0.30 & 0.63 & 0.45 & V & 0.17 & 0.00 & 0.20 \\
        QP\_002\_T1 & 0.00 & 0.00 & 1.00 & 1.00 & 1.00 & 1.00 & 0.00 & 0.00 & 0.00 & V \\
        QP\_002\_T2 & 1.00 & 1.00 & 1.00 & 1.00 & 1.00 & 0.00 & 1.00 & 1.00 & 0.00 & V \\
        QP\_002\_T3-1 & 0.00 & 0.00 & 0.00 & 0.00 & 0.00 & 0.00 & 0.00 & 0.00 & 0.00 & 0.00 \\
        QP\_002\_T3-2 & 0.00 & 0.00 & 0.00 & 0.00 & 0.00 & 0.00 & 0.00 & 0.00 & 0.00 & 0.00 \\
        QP\_002\_T3-3 & 0.00 & 1.00 & 1.00 & 1.00 & 1.00 & 0.00 & 0.00 & 0.00 & 0.00 & 0.00 \\
        QP\_002\_T3-4 & 0.00 & 0.67 & 0.00 & 0.00 & 0.00 & 0.00 & 0.00 & 0.00 & 0.00 & 0.00 \\
        QP\_002\_T4 & 0.70 & 0.10 & 1.00 & 0.70 & 1.00 & 0.70 & 0.10 & 0.10 & 0.10 & 0.10 \\
        QP\_002\_T5-1 & 0.20 & 0.55 & 0.40 & 0.40 & 0.40 & 0.00 & 0.20 & 0.20 & 0.20 & 0.20 \\
        QP\_002\_T5-2 & 0.20 & 0.20 & 0.40 & 0.70 & 0.40 & 0.20 & 0.20 & 0.20 & 0.40 & 0.40 \\
        QP\_002\_T5-3 & 0.20 & 0.20 & 0.20 & 0.20 & 0.20 & 0.00 & 0.00 & 0.20 & 0.20 & 0.20 \\
        QP\_003\_T1 & 1.00 & 1.00 & 1.00 & 1.00 & 1.00 & V & 1.00 & 1.00 & 1.00 & 1.00 \\
        QP\_003\_T2 & 1.00 & 1.00 & 1.00 & 1.00 & 1.00 & V & V & V & V & V \\
        QP\_003\_T3 & 1.00 & 1.00 & 1.00 & 1.00 & 1.00 & 1.00 & 1.00 & 1.00 & 1.00 & 1.00 \\
        QP\_003\_T4 & 1.00 & 1.00 & 1.00 & 1.00 & 1.00 & V & V & V & V & 0.00 \\
        QP\_003\_T5 & 0.17 & 0.17 & V & 0.67 & V & V & V & V & 0.00 & V \\
        QP\_004\_T1 & V & V & V & 1.00 & 1.00 & V & V & 0.10 & 0.10 & V \\
        QP\_004\_T2 & V & V & 1.00 & V & 1.00 & 0.10 & 0.10 & 0.10 & 0.10 & 0.10 \\
        QP\_004\_T3 & 1.00 & 1.00 & V & V & V & V & V & V & V & V \\
        \midrule
        Mean score & 42.7 & 44.1 & 50.3 & 47.7 & 59.7 & 15.0 & 15.7 & 22.0 & 13.5 & 18.3 \\
        Solved & 8 & 8 & 10 & 8 & 12 & 2 & 3 & 4 & 2 & 3 \\
        \texttt{VOID} & 2 & 2 & 4 & 3 & 3 & 6 & 8 & 6 & 4 & 8 \\
        \bottomrule
    \end{tabular}
\end{table}

\begin{table}[t]
    \centering
    \footnotesize
    \setlength{\tabcolsep}{6pt}
    \renewcommand{\arraystretch}{1.02}
    \caption{Per-task scores in the history-window sweep (Opus~5, medium effort, English prompts). V: \texttt{VOID} run (no gradable outcome). The bottom rows give the mean score with \texttt{VOID} scored 0, the number of fully solved tasks, and the number of \texttt{VOID} runs.}
    \label{tab:pertask-window}
    \begin{tabular}{@{}lccccc@{}}
        \toprule
        Task & $w$=1 & $w$=3 & $w$=5 & $w$=10 & $w$=15 \\
        \midrule
        QP\_001\_T1-1 & 0.00 & 0.00 & 0.00 & 0.00 & 0.00 \\
        QP\_001\_T1-2 & 1.00 & 1.00 & 1.00 & 1.00 & 1.00 \\
        QP\_001\_T2-1 & 0.15 & 0.09 & 0.03 & 0.00 & 0.00 \\
        QP\_001\_T2-2 & 0.17 & V & 0.23 & 0.24 & 0.31 \\
        QP\_001\_T3 & 0.07 & 0.30 & 0.00 & 0.38 & 0.00 \\
        QP\_002\_T1 & 0.00 & 1.00 & 0.00 & 1.00 & 1.00 \\
        QP\_002\_T2 & 1.00 & 1.00 & 1.00 & 1.00 & 1.00 \\
        QP\_002\_T3-1 & 0.00 & 0.00 & 0.00 & 0.00 & 0.00 \\
        QP\_002\_T3-2 & 0.00 & 0.00 & 0.00 & 0.00 & 0.00 \\
        QP\_002\_T3-3 & 0.00 & 1.00 & 1.00 & 0.00 & 1.00 \\
        QP\_002\_T3-4 & 0.00 & 0.00 & 0.67 & 0.33 & 0.33 \\
        QP\_002\_T4 & 0.10 & 0.10 & 0.10 & 0.70 & 0.10 \\
        QP\_002\_T5-1 & 0.40 & 0.40 & 0.55 & 0.40 & 0.20 \\
        QP\_002\_T5-2 & 0.40 & 0.55 & 0.20 & 0.20 & 0.40 \\
        QP\_002\_T5-3 & 0.20 & 0.20 & 0.20 & 0.20 & 0.20 \\
        QP\_003\_T1 & 1.00 & 1.00 & 1.00 & 1.00 & 1.00 \\
        QP\_003\_T2 & 1.00 & 1.00 & 1.00 & 1.00 & 1.00 \\
        QP\_003\_T3 & 1.00 & 1.00 & 1.00 & 1.00 & 1.00 \\
        QP\_003\_T4 & 1.00 & 1.00 & 1.00 & 1.00 & 0.00 \\
        QP\_003\_T5 & V & 0.33 & 0.17 & 0.67 & 0.00 \\
        QP\_004\_T1 & V & V & V & 1.00 & 1.00 \\
        QP\_004\_T2 & V & V & V & 1.00 & 1.00 \\
        QP\_004\_T3 & 0.61 & 1.00 & 1.00 & 1.00 & 1.00 \\
        \midrule
        Mean score & 35.2 & 47.7 & 44.1 & 57.0 & 50.2 \\
        Solved & 6 & 9 & 8 & 10 & 10 \\
        \texttt{VOID} & 3 & 3 & 2 & 0 & 0 \\
        \bottomrule
    \end{tabular}
\end{table}

\begin{table}[t]
    \centering
    \scriptsize
    \setlength{\tabcolsep}{3pt}
    \renewcommand{\arraystretch}{1.02}
    \caption{Per-task scores in the task-language sweep (medium effort, $w=5$). V: \texttt{VOID} run (no gradable outcome). The bottom rows give the mean score with \texttt{VOID} scored 0, the number of fully solved tasks, and the number of \texttt{VOID} runs.}
    \label{tab:pertask-language}
    \begin{tabular}{@{}lcccccccccccc@{}}
        \toprule
        & \multicolumn{6}{c}{Opus~5} & \multicolumn{6}{c}{Sonnet~5} \\
        \cmidrule(lr){2-7} \cmidrule(lr){8-13}
        Task & en & zh & ja & es & fr & th & en & zh & ja & es & fr & th \\
        \midrule
        QP\_001\_T1-1 & 0.00 & 0.00 & 0.00 & 0.00 & 0.00 & 0.00 & 0.00 & 0.00 & 0.00 & 0.00 & 0.00 & 0.00 \\
        QP\_001\_T1-2 & 1.00 & 0.00 & 0.00 & 0.00 & 1.00 & 0.00 & 0.00 & 0.00 & 1.00 & 0.00 & 0.00 & 0.00 \\
        QP\_001\_T2-1 & 0.03 & 0.12 & V & 0.00 & 0.00 & 0.11 & V & V & V & V & V & V \\
        QP\_001\_T2-2 & 0.23 & 0.00 & 0.17 & 0.00 & 0.22 & 0.00 & V & V & V & V & V & V \\
        QP\_001\_T3 & 0.00 & 0.34 & 0.32 & 0.00 & 0.16 & 0.60 & V & V & V & V & V & 0.18 \\
        QP\_002\_T1 & 0.00 & 0.00 & 1.00 & 1.00 & 0.00 & 1.00 & 0.00 & 1.00 & 0.00 & 1.00 & 0.00 & 0.00 \\
        QP\_002\_T2 & 1.00 & 1.00 & 1.00 & 1.00 & 1.00 & 1.00 & 1.00 & 0.00 & 1.00 & 0.00 & V & 0.00 \\
        QP\_002\_T3-1 & 0.00 & 0.00 & 0.00 & 0.00 & 0.00 & 0.00 & 0.00 & 0.00 & 0.00 & 0.00 & 0.00 & 0.00 \\
        QP\_002\_T3-2 & 0.00 & 0.00 & 0.00 & 0.00 & 0.00 & 0.00 & 0.00 & 0.00 & 0.00 & 0.00 & 0.00 & 0.00 \\
        QP\_002\_T3-3 & 1.00 & 0.00 & 1.00 & 1.00 & 0.00 & 1.00 & 0.00 & 0.00 & 0.00 & 0.00 & 0.00 & 0.00 \\
        QP\_002\_T3-4 & 0.67 & 0.33 & 0.33 & 0.00 & 0.33 & 0.00 & 0.00 & 0.00 & 0.00 & 0.00 & 0.00 & 0.00 \\
        QP\_002\_T4 & 0.10 & 0.10 & 0.10 & 0.10 & 0.10 & 0.10 & 0.10 & 0.10 & 0.10 & 0.10 & 0.10 & 0.10 \\
        QP\_002\_T5-1 & 0.55 & 0.20 & 0.20 & 0.40 & 0.20 & 0.20 & 0.20 & 0.00 & 0.20 & 0.20 & 0.20 & 0.20 \\
        QP\_002\_T5-2 & 0.20 & 0.20 & 0.55 & 0.20 & 0.20 & 0.20 & 0.20 & 0.20 & 0.20 & 0.20 & 0.00 & 0.00 \\
        QP\_002\_T5-3 & 0.20 & 0.20 & 0.20 & 0.20 & 0.20 & 0.20 & 0.00 & 0.20 & 0.20 & 0.20 & 0.20 & 0.20 \\
        QP\_003\_T1 & 1.00 & 1.00 & 1.00 & 1.00 & 1.00 & 1.00 & 1.00 & 1.00 & V & 1.00 & 1.00 & V \\
        QP\_003\_T2 & 1.00 & 0.00 & 1.00 & 1.00 & 1.00 & 1.00 & V & 0.00 & V & V & 0.00 & 0.00 \\
        QP\_003\_T3 & 1.00 & 1.00 & 1.00 & 1.00 & 1.00 & 1.00 & 1.00 & V & 1.00 & V & 1.00 & 1.00 \\
        QP\_003\_T4 & 1.00 & 1.00 & 1.00 & 1.00 & 1.00 & 1.00 & V & V & V & V & V & V \\
        QP\_003\_T5 & 0.17 & 0.33 & 0.33 & V & V & 0.00 & V & V & 0.33 & V & V & V \\
        QP\_004\_T1 & V & 1.00 & 1.00 & V & V & V & V & V & V & V & V & V \\
        QP\_004\_T2 & V & V & 1.00 & 1.00 & V & 1.00 & 0.10 & V & 0.10 & V & 0.10 & 0.10 \\
        QP\_004\_T3 & 1.00 & V & 1.00 & 1.00 & 0.37 & 1.00 & V & V & V & V & V & V \\
        \midrule
        Mean score & 44.1 & 29.7 & 53.1 & 43.0 & 33.9 & 45.3 & 15.7 & 10.9 & 18.0 & 11.7 & 11.3 & 7.7 \\
        Solved & 8 & 5 & 10 & 9 & 6 & 9 & 3 & 2 & 3 & 2 & 2 & 1 \\
        \texttt{VOID} & 2 & 2 & 1 & 2 & 3 & 1 & 8 & 9 & 8 & 10 & 8 & 7 \\
        \bottomrule
    \end{tabular}
\end{table}
\clearpage

\section{Additional trajectory analysis}
\subsection{Cross-domain trajectory statistics}
\label{sec:app-traj-all}

\begin{figure}[h]
    \centering
    \includegraphics[width=0.62\textwidth]{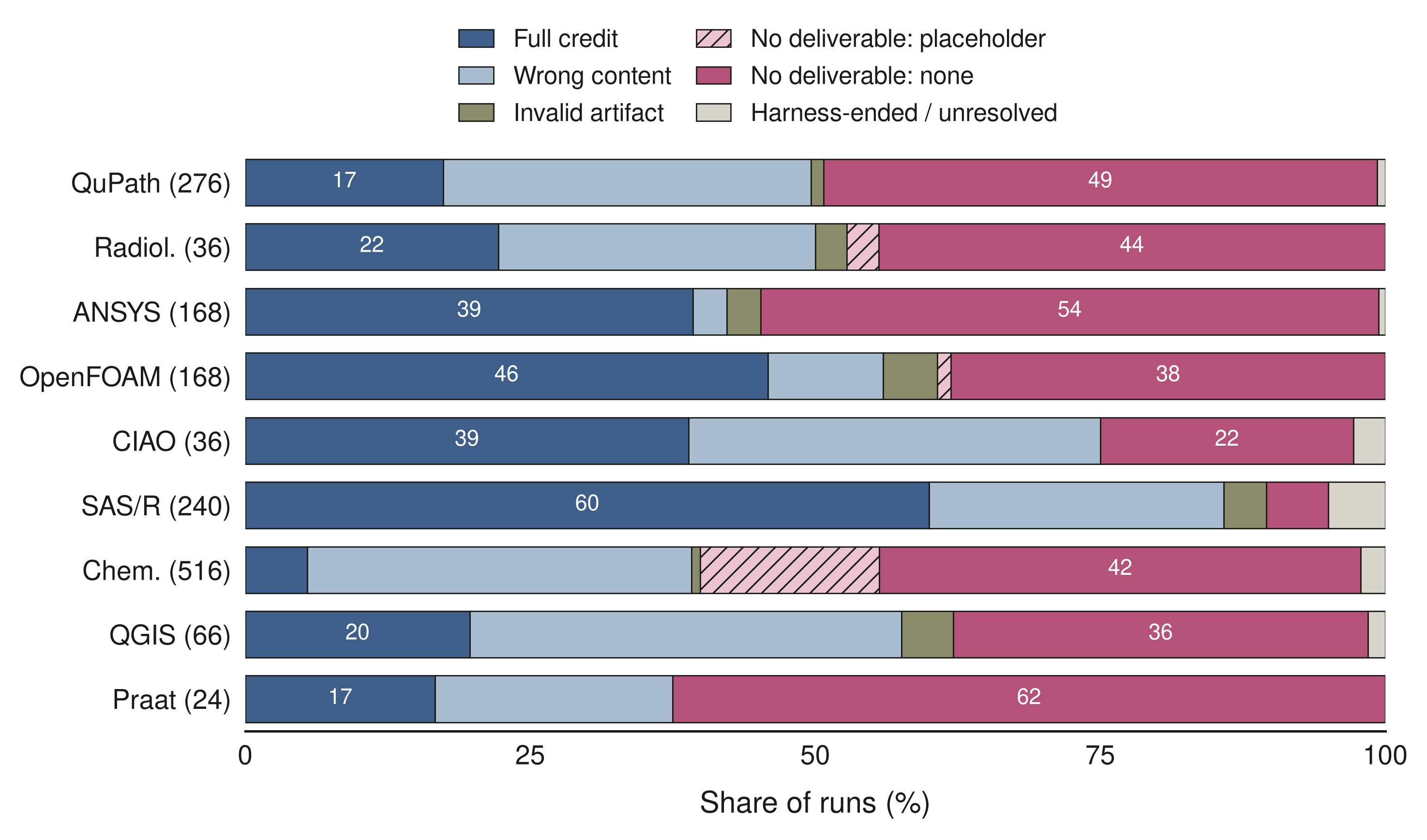}
    \caption{Outcome of each run by software configuration, all models pooled (12 models, 11 on QGIS; number of runs in parentheses); categories as in Figure~\ref{fig:trajectory}a.}
    \label{fig:traj-domains}
\end{figure}

\paragraph{Runs.} The analysis covers every released run: 12 models on the 128 tasks whose trajectories are in the released archive (all domains except biology), one run per model and task, from the runs of the main results. GPT-5.6 sol has no QGIS runs, which gives 1{,}530 runs. Ten chemistry runs have a score but no released trajectory, and five further runs contain no action, so the interaction measures use 1{,}515 runs. Table~\ref{tab:traj-outcome-domain} and Figure~\ref{fig:traj-domains} break the outcomes down by software configuration.

\paragraph{Outcome labels.} For each software configuration, an adapter extracts facts from the released records: the score, whether a gate failed, the state of the graded artifact, the harness status, the harness step cap and the task's declared budget, and any evidence that the harness or the environment ended the run. A single function then assigns the label, checking the conditions in this order:
\begin{enumerate}
    \item \emph{Full credit}: the score equals the maximum and no gate failed.
    \item \emph{Ended by the harness or environment}: there is no usable artifact, and either the records show a cause outside the agent or the run stopped at a harness step cap below the task's declared budget.
    \item \emph{No deliverable}: the graded artifact is missing or empty, contains only a placeholder (a header, rows with only the given identifiers, a route without leaves, or a placeholder entry), or is a task-provided file left unchanged.
    \item \emph{Unresolved}: the released records do not determine the state of the artifact (8 chemistry runs without a released trajectory).
    \item \emph{Invalid artifact}: the file itself is malformed (unparsable, wrong schema, non-numeric values such as a bare \texttt{NaN}), an input-integrity or format gate failed, or a required secondary file is missing.
    \item \emph{Wrong content}: all remaining runs.
\end{enumerate}
The label depends on the graded artifact, not on the reason a run stopped. A run with a well-formed but incorrect artifact therefore counts as wrong content even if it also exhausted its budget. Two conventions keep the labels comparable across software:
\begin{itemize}
    \item A grader check that finds a required object missing or misclassified inside a well-formed file counts as wrong content.
    \item A wall-clock limit that applies equally to all models counts as part of the budget.
\end{itemize}
The extracted facts were audited run by run on 1553 runs per configuration. The audits and the label definitions above changed 41 extracted facts.

\emph{Harness-ended runs.} The 20 harness-ended runs comprise:
\begin{itemize}
    \item 7 runs that stopped at a harness step cap below the task's declared budget;
    \item 6 runs in which requests exceeded the model API's per-request size limit;
    \item 6 runs ended by consecutive empty replies or timeouts, 3 of them of undetermined cause (possibly refusals);
    \item 1 run whose screen stayed blank throughout.
\end{itemize}
Six further runs were affected by such causes but left a usable artifact and are labeled by it (1 full credit, 3 wrong content, 2 invalid artifact).

\emph{Runs without a deliverable.} Of the 667 runs without a deliverable, 613 exhausted their step or time budget, 47 stopped of their own accord, and 7 were stopped after consecutive replies without an executable action. The 22 chemistry structure-recognition tasks name GChemPaint as the drawing tool, which is not installed in the evaluation image. We keep these runs as agent outcomes because the graded file can be written without the editor, but 76 of their 145 runs without a deliverable mention GChemPaint.

\paragraph{Interaction measures.} An action step is one executed action record.
\begin{itemize}
    \item \emph{GUI step}: the step contains a pointer call (click, move, drag, or scroll). Unlike the route analysis in Appendix~\ref{sec:app-stat}, which counts any desktop automation call as GUI interaction, a GUI step here requires a pointer action. A click into a terminal window counts as a GUI step.
    \item \emph{CLI step}: the step opens a terminal, runs a shell command, or calls the application's scripting interface. Examples of scripting interfaces are the QuPath Groovy API, the Fluent text console and Scheme, the OpenFOAM utilities and solvers, R and SAS, the CIAO tools and \texttt{pycrates}, PyQGIS and GDAL, RDKit, Open Babel and the ASKCOS API, and \texttt{pydicom} and the Slicer console. For ANSYS, which runs on Windows, the equivalent \texttt{cmd} and PowerShell commands are included.
    \item A step can count as both GUI and CLI. Typing and key presses alone count as neither.
    \item \emph{CLI lower bound.} The CLI share is a lower bound. Commands outside the shared shell vocabulary and the software's scripting interface are not counted (for example, plain Python typed into the QGIS console). Action records in the released logs are also truncated at 2{,}000 characters (600 in chemistry, where truncated actions are partly recovered from the reply).
    \item \emph{Narration}: a reply narrates when its text outside code blocks and termination tokens is at least 20 characters long, with CJK characters counted twice.
    \item \emph{Thinking share}: thinking tokens divided by output tokens, summed over the per-call usage records. For QuPath, ANSYS, OpenFOAM, and CIAO, these records come from the original run logs rather than the released archive. The share is unavailable for chemistry, whose logs contain only totals. The usage records of Claude Fable~5.1 on QGIS contain no thinking counts. Runs in which every call reports zero thinking tokens are treated as missing: 6 Kimi K3 runs on QGIS and 23 Qwen-CUA runs on ANSYS, OpenFOAM, and CIAO.
    \item Narration and thinking both depend on the prompt format of each software configuration, so they are best compared within a configuration.
    \item \emph{Repeated reply}: all actions of one reply are concatenated and normalized (comments removed, whitespace collapsed, sleep durations zeroed). A reply is repeated when its normalized action is identical to one of the previous three.
    \item \emph{Automation calls}: calls of the automation library, grouped into click, double click (including \texttt{clicks=2}), right click, drag, move, scroll, typing, key press, and hotkey.
\end{itemize}
Tables~\ref{tab:traj-gui} and~\ref{tab:traj-cli} give the GUI and CLI shares by model and software configuration. Table~\ref{tab:traj-ops} gives the remaining measures by model.

\paragraph{Matched comparisons.} These comparisons use the runs with at least one action and exclude harness-ended and unresolved runs.
\begin{itemize}
    \item \emph{Task-level contrast}: for each task with at least one full-credit run and one other run (86 tasks; 69 for thinking), we take the difference between the means of the two groups. We report the median over tasks with a bootstrap 95\% interval (5{,}000 resamples).
    \item \emph{Modeldomain contrast}: the same statistic computed within model-domain groups that have at least two runs of each outcome (39 groups; 34 for thinking). For narration the interval is degenerate at zero, because in many groups neither outcome contains any narration.
    \item \emph{Fixed-effects model}: one linear probability model per measure, regressing full credit on the measure with model and task fixed effects. Standard errors are cluster-robust by task (CR1). The coefficients per unit share are:
    \begin{itemize}
        \item CLI: $+0.036$ (SE $0.037$, $p=0.32$);
        \item GUI: $-0.067$ ($0.033$, $p=0.05$);
        \item narration: $+0.023$ ($0.027$, $p=0.39$);
        \item repetition: $+0.148$ ($0.085$, $p=0.08$);
        \item thinking: $-0.105$ ($0.087$, $p=0.22$).
    \end{itemize}
    The models use $N=1{,}496$ runs, except thinking with 951 runs on 85 tasks. None is significant after a Bonferroni correction over the five measures.
\end{itemize}

\paragraph{Further observations.} These details complement the summary in the main text.
\begin{itemize}
    \item \emph{Placeholders}: in 72 of the 77 chemistry placeholder runs of the GPT models, the agent exhausts the budget before filling the skeleton. Placeholders account for 29 of GPT-6 Astra's 32 runs without a deliverable.
    \item \emph{Repetition}: the two Qwen models repeat 21-25\% of their replies on average; every other model repeats at most 9\%.
    \item \emph{Interaction habits}: GPT-5.6 sol, terra, and luna press a key in 53-63\% of their steps, almost always Enter to confirm typed input. MiniMax~M3 narrates in 79\% of its replies.
    \item \emph{Rank correlations}: the mean cross-software rank correlation of the CLI share is 0.43. All four correlations exclude Praat, and the thinking correlation also excludes chemistry.
    \item \emph{Within-group differences}: among runs of the same model in the same domain, every median difference lies within 5 percentage points of zero. In the fixed-effects models, GUI share leans negative (uncorrected $p=0.05$).
\end{itemize}

\paragraph{Efficiency of successful runs.} Runs with full credit take a median of 19 steps:
\begin{itemize}
    \item 12-14 steps for Claude Fable~5.1, Gemini~3.1 Pro, GPT-5.6 luna, and GPT-6 Astra;
    \item 18-21 steps for Claude Opus~5, GPT-5.6 sol, and GPT-5.6 terra;
    \item 27-60 steps for Qwen3.7-Plus, Qwen-CUA, Claude Sonnet~5, Kimi~K3, and MiniMax~M3 (Table~\ref{tab:traj-ops}).
\end{itemize}
Across the 128 tasks, no model obtains full credit on 40 tasks, and 2 tasks are solved with full credit by every model.

\input{tab_traj_outcome_domain}
\input{tab_traj_gui}
\input{tab_traj_cli}
\input{tab_traj_ops}

\subsection{Geographic questions}
\label{sec:app-geo}
\suppressfloats[t]

\definecolor{geoA}{rgb}{0.98,0.85,0.82}
\definecolor{geoD}{rgb}{0.99,0.92,0.78}
\definecolor{geoE}{rgb}{0.84,0.90,0.98}
\definecolor{geoF}{rgb}{0.89,0.86,0.96}
\definecolor{geoH}{rgb}{0.88,0.88,0.88}
\definecolor{geoP}{rgb}{0.84,0.94,0.84}

This appendix analyzes the trajectories on the six geoscience tasks, which run in QGIS on remote-sensing imagery. The tasks are NDVI zonal statistics (GEO\_001), aircraft point annotation (GEO\_002), SAR-optical image registration (GEO\_003), building footprint digitization (GEO\_004), road centerline digitization (GEO\_005), and building damage grading (GEO\_006). We evaluate eleven backbones with one run per task, a limit of 100 interaction steps, and a screen resolution of $1920\times1080$, which gives 66 runs. Thirteen runs succeed, and the mean partial-credit score is 32.5\%. Of the 53 failed runs, 28 deliver an incorrect result, 21 are \texttt{VOID}, and 4 stop early without issuing further actions. We read each trajectory and assign each failed run one primary cause. The causes are ordered along the execution path, from an action that never takes effect to a finished artifact that misses the target (Table~\ref{tab:geo-taxonomy}). Table~\ref{tab:geo-matrix} gives the outcome of every run.

\begin{figure}[b]
    \centering
    \includegraphics[width=\textwidth]{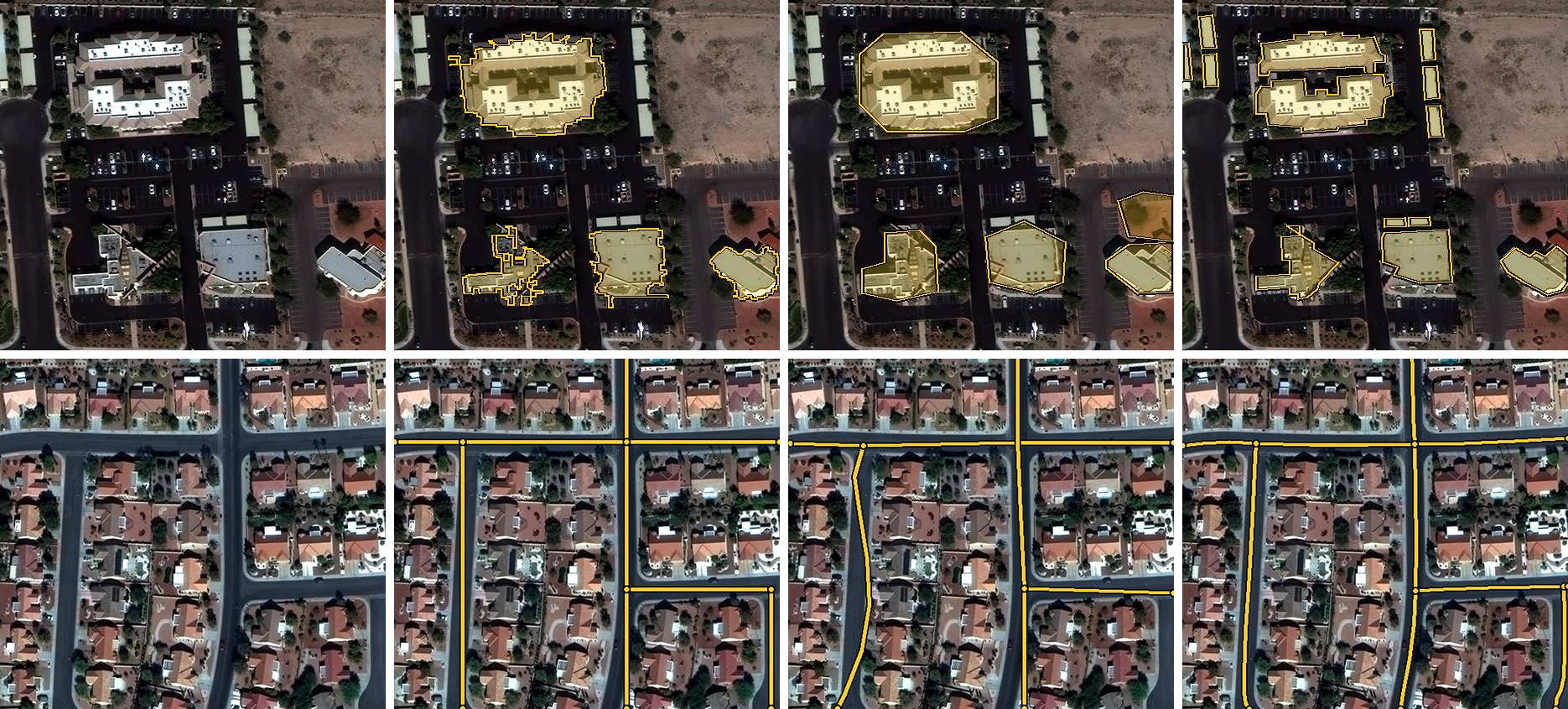}
    \caption{Delivered geometries (yellow) on two digitization tasks. Columns from left to right: the input image, Opus~5, Fable~5.1, and GPT-6-Astra. Top row: building footprints (GEO\_004). The scene contains 4 large buildings and 8 small sheds beside the parking lots, and only GPT-6-Astra digitizes most of the sheds. Bottom row: road centerlines (GEO\_005). Opus~5 and GPT-6-Astra cover the full network, and Fable~5.1 leaves part of it out.}
    \label{fig:geo-deliveries}
\end{figure}

\paragraph{Successful runs.} The 13 successful runs fall on GEO\_001 (7 of 11 runs), GEO\_002 (4), and GEO\_005 (2). Compared with failed runs on the same tasks, they share two practices. First, 11 of the 13 compute the result with a script in the QGIS Python console or a terminal. The two exceptions annotate the 15 aircraft of GEO\_002 by clicking in the interface. Second, all seven successful GEO\_001 runs read back the content of their result table before they stop. The length of a trajectory says little about its method. On GEO\_005 (Figure~\ref{fig:geo-deliveries}, bottom row), GPT-6-Astra writes the road network with one script in 4 steps and scores 0.98. Opus~5 measures the road positions from the image, repairs gaps over 88 steps, and scores 0.90. Fable~5.1 draws five correct segments in 10 steps, reports the task as complete, and scores 0.76 because part of the network is missing.

\paragraph{Interaction-layer failures.} In 23 of the 53 failed runs (43\%), the intended action never takes effect on the desktop. The task instruction states the screen resolution of $1920\times1080$, but four backbones (Gemini~3.1~Pro, Qwen3.7-plus, Qwen-CUA, and MiniMax-M3) do not follow it, so their clicks miss the target (18 runs). These backbones fail this way on almost every task (Table~\ref{tab:geo-matrix}), so the cause follows the backbone and not the task. The other five runs stop early because the replies contain no executable action (4 runs) or issue malformed actions that never execute (1 run). No run of GPT-6-Astra or Fable~5.1 fails at this stage.

\paragraph{Output verification.} Several failed runs stop without checking the content of their output, or check a quantity that cannot reveal the error. GPT-5.6-terra declares GEO\_001 done after 5 steps although its statistics are empty. On GEO\_003, Fable~5.1 inspects the residual of its registration before it stops and reports a mean error of about 2.8 pixels. This value is the fit residual on its own four control points, whereas the error at 104 independent check points is 216 pixels. The case is consistent with the chemistry case study (Figure~\ref{fig:opus5-case-studies}), in which validation steps that share one assumption fail to expose the error. When agents do verify, they mostly confirm that the output exists and has the expected structure. None of the eight runs that deliver footprints on GEO\_004 checks whether all buildings are covered.

\paragraph{Method and completion failures.} Among runs whose actions reach the desktop, the most frequent causes are an inadequate method (E, 12 runs) and stopping before the work is complete (F and G, 13 runs). On the registration task GEO\_003, the six runs that deliver a result place between 4 and 9 control points, which is too few for a deformation that varies across the image. Opus~5 measures such varying displacements and still corrects them with one global shift. The GEO\_004 scene contains 12 buildings, 4 large ones and 8 small sheds (Figure~\ref{fig:geo-deliveries}, top row). Six of the seven runs with cause F on this task digitize mainly the large buildings and report the task as complete. GPT-6-Astra is the exception. It digitizes 10 of the 12 buildings, including 6 sheds, and misses the two smallest. On GEO\_006, the way an agent inspects the imagery matters. Fable~5.1 crops a before and after image pair for each of the 30 buildings and grades 20 of them correctly, the highest result on this task. The other runs that deliver labels compare the two images at a coarser level and grade 8 to 13 buildings correctly. Two runs (H) use all 100 steps and write no output.
\begin{table}[t]
    \centering
    \footnotesize
    \setlength{\tabcolsep}{4pt}
    \renewcommand{\arraystretch}{1.05}
    \caption{Primary causes of the 53 failed runs on the geoscience tasks, ordered along the execution path. Each failed run has exactly one cause. The last column lists the affected backbones with their number of runs.}
    \label{tab:geo-taxonomy}
    \begin{tabular}{@{}l@{\hspace{5pt}}p{0.17\textwidth}p{0.37\textwidth}cp{0.25\textwidth}@{}}
        \toprule
        & Cause & Criterion & Runs & Backbones (runs) \\
        \midrule
        \multicolumn{5}{@{}l}{\textit{Action does not take effect}} \\
        A & Coordinate-frame mismatch & Click coordinates do not follow the $1920\times1080$ pixel frame stated in the instruction, so clicks miss their target & 18 & Qwen3.7-plus (6), Gemini~3.1~Pro (5), Qwen-CUA (4), MiniMax-M3 (3) \\
        B & No executable action & Reply contains no executable action, for example only a text plan & 4 & Kimi-K3 (3), Qwen-CUA (1) \\
        C & Malformed action & Actions are malformed and never execute & 1 & MiniMax-M3 (1) \\
        \multicolumn{5}{@{}l}{\textit{Artifact is malformed}} \\
        D & Data type or geometry error & Wrong field type or geometry dimension, so the result is invalid & 3 & Gemini~3.1~Pro, MiniMax-M3, Qwen-CUA (1 each) \\
        \multicolumn{5}{@{}l}{\textit{Method is inadequate}} \\
        E & Wrong model or prior in place of measurement & The chosen method cannot reach the required accuracy & 12 & GPT-5.6-luna (3), GPT-5.6-terra, Sonnet~5, GPT-6-Astra, Fable~5.1 (2 each), Opus~5 (1) \\
        \multicolumn{5}{@{}l}{\textit{Work stops early}} \\
        F & Low recall with a completion claim & Agent declares the task done after part of the work and does not check recall & 9 & GPT-5.6-terra, Fable~5.1 (2 each), GPT-5.6-luna, Opus~5, Sonnet~5, Kimi-K3, GPT-6-Astra (1 each) \\
        G & Miscount or failed self-check & Wrong count, or a self-check that reaches the wrong conclusion & 4 & GPT-5.6-luna, GPT-5.6-terra, Sonnet~5, Kimi-K3 (1 each) \\
        \multicolumn{5}{@{}l}{\textit{No convergence}} \\
        H & Step limit with no output & Actions take effect, but no artifact is written within 100 steps & 2 & Opus~5, Sonnet~5 (1 each) \\
        \bottomrule
    \end{tabular}
\end{table}

\begin{table}[t]
    \centering
    \footnotesize
    \setlength{\tabcolsep}{5pt}
    \renewcommand{\arraystretch}{1.05}
    \caption{Outcome of every run on the six geoscience tasks (one run per cell, step limit 100). P: successful run. A-H: primary cause of failure as in Table~\ref{tab:geo-taxonomy}, shaded by stage. Pass: number of successful runs. Score: mean partial-credit score in percent.}
    \label{tab:geo-matrix}
    \begin{tabular}{@{}lcccccccc@{}}
        \toprule
        Backbone & GEO\_001 & GEO\_002 & GEO\_003 & GEO\_004 & GEO\_005 & GEO\_006 & Pass & Score \\
        \midrule
        GPT-6-Astra & \cellcolor{geoP}P & \cellcolor{geoP}P & \cellcolor{geoE}E & \cellcolor{geoF}F & \cellcolor{geoP}P & \cellcolor{geoE}E & 3/6 & 69.0 \\
        Fable~5.1 & \cellcolor{geoP}P & \cellcolor{geoP}P & \cellcolor{geoE}E & \cellcolor{geoF}F & \cellcolor{geoF}F & \cellcolor{geoE}E & 2/6 & 68.1 \\
        Opus~5 & \cellcolor{geoP}P & \cellcolor{geoP}P & \cellcolor{geoE}E & \cellcolor{geoF}F & \cellcolor{geoP}P & \cellcolor{geoH}H & 3/6 & 61.6 \\
        GPT-5.6-luna & \cellcolor{geoP}P & \cellcolor{geoF}G & \cellcolor{geoE}E & \cellcolor{geoF}F & \cellcolor{geoE}E & \cellcolor{geoE}E & 1/6 & 48.6 \\
        Sonnet~5 & \cellcolor{geoP}P & \cellcolor{geoF}G & \cellcolor{geoE}E & \cellcolor{geoF}F & \cellcolor{geoE}E & \cellcolor{geoH}H & 1/6 & 37.1 \\
        GPT-5.6-terra & \cellcolor{geoF}G & \cellcolor{geoP}P & \cellcolor{geoE}E & \cellcolor{geoF}F & \cellcolor{geoF}F & \cellcolor{geoE}E & 1/6 & 36.5 \\
        Kimi-K3 & \cellcolor{geoP}P & \cellcolor{geoA}B & \cellcolor{geoA}B & \cellcolor{geoF}F & \cellcolor{geoF}G & \cellcolor{geoA}B & 1/6 & 19.7 \\
        MiniMax-M3 & \cellcolor{geoP}P & \cellcolor{geoA}A & \cellcolor{geoA}C & \cellcolor{geoD}D & \cellcolor{geoA}A & \cellcolor{geoA}A & 1/6 & 16.7 \\
        Gemini~3.1~Pro & \cellcolor{geoD}D & \cellcolor{geoA}A & \cellcolor{geoA}A & \cellcolor{geoA}A & \cellcolor{geoA}A & \cellcolor{geoA}A & 0/6 & 0.0 \\
        Qwen3.7-plus & \cellcolor{geoA}A & \cellcolor{geoA}A & \cellcolor{geoA}A & \cellcolor{geoA}A & \cellcolor{geoA}A & \cellcolor{geoA}A & 0/6 & 0.0 \\
        Qwen-CUA & \cellcolor{geoA}A & \cellcolor{geoA}A & \cellcolor{geoA}B & \cellcolor{geoA}A & \cellcolor{geoA}A & \cellcolor{geoD}D & 0/6 & 0.0 \\
        \midrule
        Pass & 7/11 & 4/11 & 0/11 & 0/11 & 2/11 & 0/11 & 13/66 & 32.5 \\
        \bottomrule
    \end{tabular}
\end{table}

\paragraph{Scope of the analysis.} The analysis has one run per backbone and task, so the counts describe these 66 trajectories and carry no variance estimate. Each failed run receives a single primary cause.
\clearpage

\subsection{How the statistics agents work: the terminal as the default instrument}
\label{sec:app-stat}
\suppressfloats[t]

The statistics image provides a statistician's desktop. R~4.4.3 and Python are installed locally. RStudio~2025.09.1 is installed from its Debian package, and SQLite is available on the path. Nine of the twenty tasks also provide access to SAS through SAS OnDemand for Academics. Before the agent takes its first action, Firefox displays the SAS sign-in page and has the credentials saved. This appendix examines how agents used this environment. The main finding is that they worked in a shell. The dataset contains one run for each of the 240 task-model pairs formed by 20 tasks and 12 models. Among these runs, 214 never opened SAS Studio, RStudio, or a GUI text editor. The two scientific applications central to the domain, SAS Studio and RStudio, were used in only 10 runs.

\paragraph{Classifying trajectories and the limits of the classifier.}
Each \texttt{action} record contains Python code that the harness executes inside the guest. It is therefore important to distinguish running a shell command from using a terminal window. An action that does not call \texttt{pyautogui} never touches the desktop, and 92 of the 240 runs used this direct channel at least once. We scan two channels separately. The first is executed code after removing the model's own \texttt{\#} comments. The second is the model's prose, including comments embedded in its actions.

This procedure has one important limitation. Coordinate-based GUI actions leave no textual trace of the application being controlled. For example, \texttt{pyautogui.click(996,\,469)} is indistinguishable from any other click. An application reached only by clicking can therefore be identified only when the model describes what it sees. Executed code establishes application use in 22 of the 26 runs classified as such. The remaining four runs rely on narration, so we read them in full and adjudicate them manually. As an external check, this process identifies nine runs that reached a live SAS Studio session. Two independent audits of all 108 SAS-task trajectories produced the same count.

\begin{table}[t]
    \centering
    \footnotesize
    \setlength{\tabcolsep}{4.5pt}
    \renewcommand{\arraystretch}{1.05}
    \caption{Applications used by each model across its 20 statistics tasks. \emph{shell}
    denotes a run that opened none of SAS Studio, RStudio, gedit, or VS Code. Such a run may
    have used a terminal window, executed code without a window, or done both. \emph{app.\ GUI}
    denotes a run that drove at least one of the four applications. \emph{declined aloud}
    denotes a run that named an application and argued for using the shell in the same reply.
    Median input tokens are computed per run and include screenshots resent at every step.}
    \label{tab:stat-route-model}
    \resizebox{\textwidth}{!}{%
    \small
    \input{tabs/stat/route_by_model}
    }
\end{table}

\begin{figure}[t]
    \centering
    \includegraphics[width=0.78\textwidth]{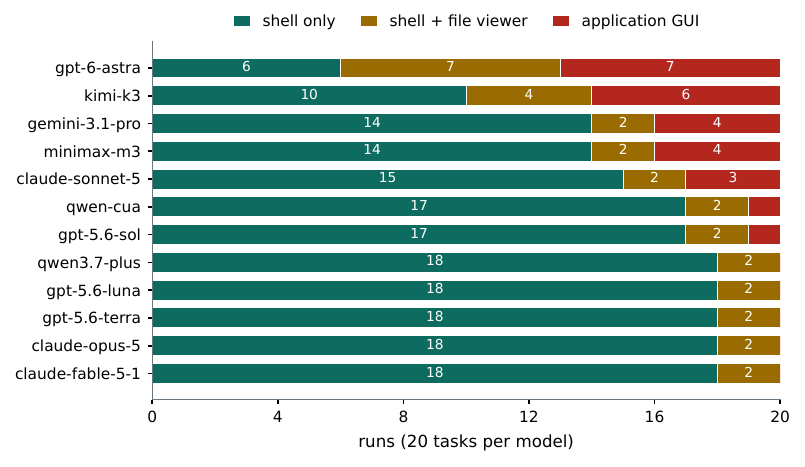}
    \caption{Software routes across all statistics runs. Five of the twelve models never
    opened SAS Studio, RStudio, gedit, or VS Code in their twenty tasks. File-viewer use is
    concentrated in the five tasks with raster inputs.}
    \label{fig:stat-census}
\end{figure}

\paragraph{The census.}
Table~\ref{tab:stat-route-model} reports the per-model breakdown, and Figure~\ref{fig:stat-census} visualises the same distribution. Of the 240 runs, 183 used only a shell. Another 31 also opened a file viewer to inspect a supplied raster or a generated PDF. The remaining 26 drove at least one application: SAS Studio in 9 runs, VS Code in 9, gedit in 9, and RStudio in 1. Two runs used two applications. Agents opened a terminal window in 209 runs. The median run did so in its first step, and 175 runs opened one by step~2.

\begin{table}[t]
    \centering
    \footnotesize
    \setlength{\tabcolsep}{5pt}
    \renewcommand{\arraystretch}{1.05}
    \caption{The same census by task group. \emph{shell\,+ viewer} denotes a shell run that
    also opened an image or PDF viewer, as required by the five tasks with raster inputs.
    \emph{obstacle only} denotes a run in which the desktop or a stray keystroke placed an
    application in front of the agent, which then spent steps closing it. These cases do not
    count as application use.}
    \label{tab:stat-group}
    \input{tabs/stat/by_group}
\end{table}

Five models never opened an application in any of their twenty tasks: \texttt{claude-fable-5-1}, \texttt{claude-opus-5}, \texttt{gpt-5.6-luna}, \texttt{gpt-5.6-terra}, and \texttt{qwen3.7-plus}. This does not mean that the desktop was idle. In total, 86.7\% of the 7{,}238 actions call \texttt{pyautogui}, because typing in a terminal window is itself GUI interaction. Thus, agents used the desktop primarily as a keyboard interface to a shell. Table~\ref{tab:stat-group} presents the same census by task group. File-viewer use is concentrated in the expected group: 27 of the 60 runs on the five tasks with raster inputs.

\begin{figure}[t]
    \centering
    \includegraphics[width=0.72\textwidth]{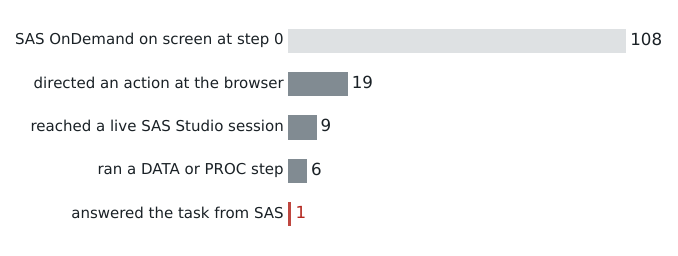}
    \caption{The 108 runs on the nine tasks that offer SAS OnDemand for Academics. The guest
    starts with the sign-in page displayed and the credentials saved, so agents did not need
    to find or launch the browser. Bars are drawn to scale. The final stage contains one run,
    \texttt{kimi-k3} on task \texttt{stat\_liver\_cohort}.}
    \label{fig:stat-funnel}
\end{figure}

\paragraph{The two scientific applications.}
Figure~\ref{fig:stat-funnel} follows the 108 runs on the nine tasks that offer SAS. Every run began with the SAS OnDemand page visible. Nineteen runs directed an action at the browser. Among them, three sent stray keystrokes that opened the sign-in dialog, while one closed Firefox to remove it from the workspace. Nine runs reached a live SAS Studio session, six submitted a \texttt{DATA} or \texttt{PROC} step, and one produced its answer with SAS. Only two of the twelve models reached SAS Studio. \texttt{gpt-6-astra} did so in seven of its nine SAS tasks, and \texttt{kimi-k3} did so in two.

RStudio received even less use. Across all 240 runs, only \texttt{gemini-3.1-pro} typed \texttt{rstudio}. This occurred at step~2 of \texttt{stat\_bernoulli\_causal}. The model clicked three times, pressed Escape, opened a new terminal at step~7, and did not return. The instruction for \texttt{stat\_registry\_dashboard} names RStudio, yet the substring \emph{studio} does not appear in any of its twelve trajectories.

The agents often explained this avoidance. Seventeen runs named an application and rejected it in the same response. Their stated reasons concerned the cost and reliability of reaching the tool, not its suitability for the statistical analysis. At step~1 of \texttt{stat\_ontario\_drug}, \texttt{claude-fable-5-1} wrote, ``\emph{The SAS OnDemand page is loaded but cloud SAS requires uploading/downloading files; R is installed locally and equally acceptable, so I'll do the analysis in R via a terminal.}'' At step~1 of \texttt{stat\_esoph\_casecontrol}, \texttt{kimi-k3} wrote, ``\emph{Given that SAS Studio runs in the cloud (requiring fragile GUI upload/download steps), I'll use the locally installed R for reliability: this is explicitly permitted.}'' On the same task, \texttt{claude-opus-5} wrote, ``\emph{I'll skip SAS and use R locally, which is faster and fully acceptable.}'' After completing its deliverables for \texttt{stat\_qol\_composite}, \texttt{kimi-k3} returned to this decision at step~54: ``\emph{Using SAS OnDemand would add significant risk (cloud upload/download, session management) with no benefit. I have a sound reason not to use SAS.}'' No run argued that R or Python was intrinsically the better statistical instrument.

Changes in instruction wording had little effect. Eight of the nine SAS tasks state, ``use SAS unless you have a reason not to: these questions come from SAS coursework and SAS is the intended tool''. Seven of these eight also state that ``nothing in the grading depends on which you choose''. Every model treated the latter clause as decisive. Two tasks instead remove Python as the most convenient alternative. \texttt{stat\_carseats\_regression} requires the analysis and submitted source to use SAS or R. It permits Python only for desktop automation and file movement. \texttt{stat\_ontario\_drug} includes the SAS preference and also prohibits Python for analysis. All 24 runs across these two tasks used R. None interacted with the SAS page that was already visible. All nine SAS tasks and both tasks that list RStudio in \texttt{related\_apps} name RStudio as an option. Ten of these 11 instructions use the wording ``open a terminal and run R or Rscript, or use RStudio''. The remaining instruction similarly allows the agent to work in R or RStudio, from either the terminal or the IDE.

\begin{table}[t]
    \centering
    \footnotesize
    \setlength{\tabcolsep}{5pt}
    \renewcommand{\arraystretch}{1.05}
    \caption{Outcomes and costs by interaction mode. Raw rows compare all shell-mode and
    application-mode runs.
    Each stratified row computes the \textbf{application-minus-shell difference} within tasks or
    models that contain both modes, then averages the differences weighted by stratum size.
    The number of shell-mode runs is therefore smaller in these rows because strata without an
    application-mode run do not contribute. The $p$-values come from 20{,}000 permutations of
    interaction-mode labels within strata. All entries are means.}
    \label{tab:stat-route-effect}
    \input{tabs/stat/route_effect}
\end{table}

\paragraph{What the interaction mode costs.}
We use \emph{interaction mode} to describe the interface through which an agent completed a task. A run is assigned to shell mode if it did not operate SAS Studio, RStudio, gedit, or VS Code. Otherwise, it is assigned to application mode. Because interaction mode is not randomly assigned, raw differences conflate the chosen mode with task difficulty and model strength. Application-mode runs are concentrated among harder tasks and particular models. Table~\ref{tab:stat-route-effect} therefore reports raw, within-task, and within-model comparisons. The stratified estimates first compute the difference in each stratum containing both modes and then average these differences, weighted by stratum size. Their $p$-values come from permuting mode labels only within strata.

All four outcomes change in the same direction under all three comparisons. Within a fixed task, application-mode use is associated with a score difference of $-0.139$ ($p=0.017$) and 29 additional steps ($p<0.001$). It is also associated with 1.60~M more input tokens per run ($p=0.003$) and 15.4~k more input tokens per step ($p=0.013$). These estimates are not causal. Within a fixed model, the estimated score difference is larger at $-0.299$ ($p<0.001$). The increase in tokens per step reflects the graphical channel, which resends a screenshot at every step.

\begin{figure}[t]
    \centering
    \includegraphics[width=\textwidth]{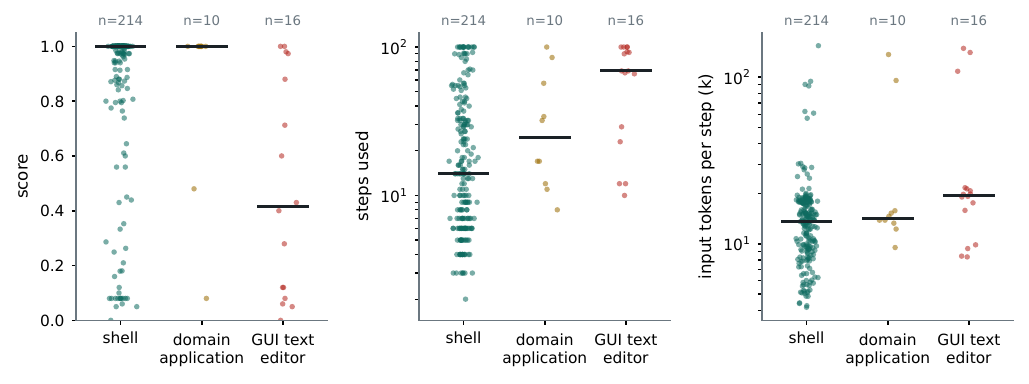}
    \caption{Outcomes and costs by application type. Each point represents one run, and the
    horizontal marker shows the median. Steps and tokens use logarithmic axes. Using SAS
    Studio or RStudio is associated with more steps but no detectable score difference. Using
    a GUI text editor is associated with both more steps and lower scores. The two runs that
    used both a domain application and an editor are assigned to the domain-application group.}
    \label{fig:stat-outcomes}
\end{figure}

However, the aggregate penalty is not concentrated where the framing of this section might suggest. Figure~\ref{fig:stat-outcomes} divides application mode into two more specific modes. Domain-application mode covers SAS Studio and RStudio, while GUI-editor mode covers gedit and VS Code. The two runs that used both are assigned to domain-application mode. The 10 runs in domain-application mode have a median score of 1.000. Within task, they take 16 more steps than shell-mode runs and have an estimated score difference of $-0.069$ ($p=0.158$). This small sample makes the score comparison underpowered, so the non-significant result should not be interpreted as evidence of no effect. In contrast, the 16 runs that used gedit or VS Code have a median score of 0.415. Within task, they take 40 more steps and score $0.228$ lower ($p=0.012$). These runs span six models, with \texttt{kimi-k3} and \texttt{minimax-m3} contributing four each. The measured cost of desktop use is therefore concentrated in authoring code with a GUI text editor. Statistical applications appear slow, but the available data do not show the same score penalty.

\begin{figure}[t]
    \centering
    \begin{minipage}[t]{0.49\textwidth}
        \centering
        \includegraphics[width=\textwidth]{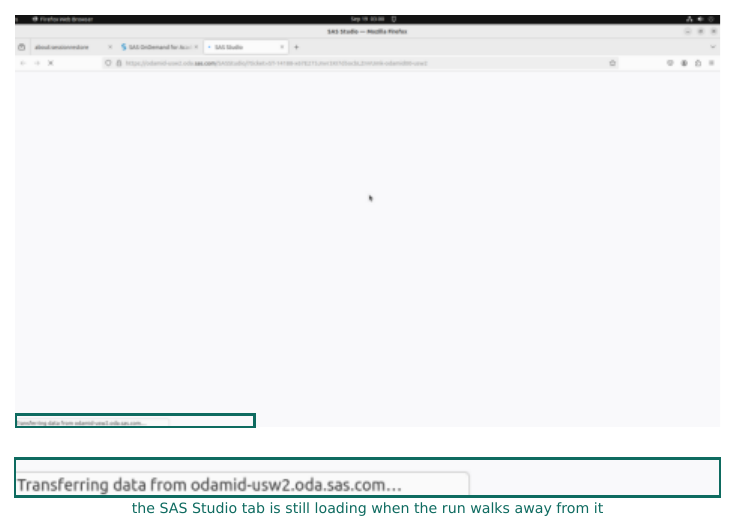}
        \\[2pt]
        {\footnotesize (a) \texttt{gpt-6-astra}, step 4 of 12.}
    \end{minipage}
    \hfill
    \begin{minipage}[t]{0.49\textwidth}
        \centering
        \includegraphics[width=\textwidth]{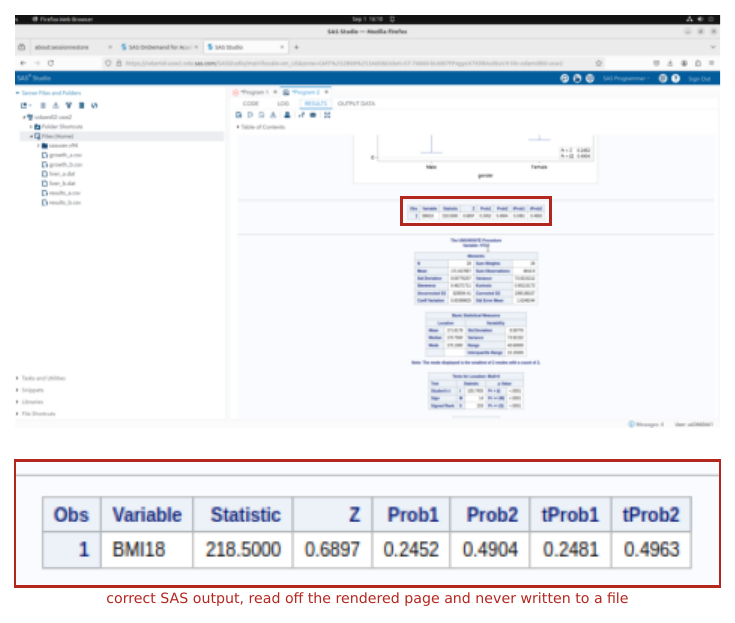}
        \\[2pt]
        {\footnotesize (b) \texttt{kimi-k3}, step 82 of 85.}
    \end{minipage}
    \caption{Two interaction modes for \texttt{stat\_group\_tests}. Each panel shows the run's own
    screenshot, with the relevant region boxed and reproduced below at native resolution.
    (a) \texttt{gpt-6-astra} has signed in and selected Launch, but SAS Studio is still
    loading. The run instead inspects the local files in a terminal and commits to R at
    step~7, earning a score of 1.000. (b) \texttt{kimi-k3} has uploaded both data files, run
    its macro, and read the correct Wilcoxon output from the rendered page. Its budget ends
    three steps later with an empty submission directory. The score is 0.080, awarded only
    for the protected-input gate.}
    \label{fig:stat-case}
\end{figure}

\paragraph{Case study: interface overhead and task completion.}
\texttt{stat\_group\_tests} provides a focused comparison of the two interaction modes. It requires one reusable, parameterised routine that selects the appropriate test from the study design and sample size, applies unchanged to two growth-study files, and reports each statistic to ten significant figures. Both SAS and R are permitted, and the grading does not depend on the choice. Nevertheless, the two runs in Figure~\ref{fig:stat-case} reached sharply different outcomes. One left SAS Studio for local R and completed the task quickly, whereas the other obtained correct statistics in SAS Studio but exhausted its budget before delivering them. The contrast isolates a phenomenon worth studying: interface overhead can turn statistically correct work into an unsuccessful software outcome.

\texttt{gpt-6-astra} first signed in to SAS Studio, but the application was still loading at step~4. It inspected the inputs locally and switched to shell mode at step~7, citing explicit control of tie handling, full-precision serialisation, and avoidance of cloud transfer. It never returned to SAS Studio and scored 1.000 in 12 steps and 2.8 minutes. By contrast, \texttt{kimi-k3} uploaded both files, wrote and submitted a SAS macro, and obtained the correct Wilcoxon statistics shown in panel~(b) at step~82. Reaching that point required recovery from a toolbar misclick, discovery of the correct ODS table names by trial, and a local Python script to patch the SAS source before pasting it again. The run ended during this repair after all 85 steps, 73 minutes, and 11.64~M input tokens. It produced no output file and scored 0.080.

\emph{Takeaway.} This case shows that statistical correctness and successful delivery are distinct capabilities. Under the current benchmark, an instruction that permits several tools rewards the least costly interaction mode, even when it names a scientific application. Tasks intended to measure application operation should therefore make that application necessary, while evaluators should execute or semantically validate the submitted program. The current 300-byte size check for the reusable routine accepted both a 639-byte file containing only comments and SAS-shaped source whose numerical results were computed in SciPy. Stronger task constraints and executable checks are needed to distinguish operating scientific software from merely producing an artefact with the expected form.

\newpage

\definecolor{lingP}{rgb}{0.84,0.94,0.84}
\definecolor{lingX}{rgb}{0.84,0.90,0.98}
\definecolor{lingL}{rgb}{0.99,0.92,0.78}
\definecolor{lingN}{rgb}{0.88,0.88,0.88}
\definecolor{lingD}{rgb}{0.89,0.86,0.96}

\begin{table}[t]
    \centering
    \footnotesize
    \setlength{\tabcolsep}{4.5pt}
    \renewcommand{\arraystretch}{1.08}
    \caption{
        Outcome of every boundary on the two linguistics tasks (one run per cell, step limit 100). Numeric entries report signed placement error in milliseconds, computed as the placed boundary minus the reference. Scores include the 0.1 awarded for returning the audio unchanged.
    }
    \label{tab:ling-matrix}

    \resizebox{0.86\linewidth}{!}{%
        \input{tabs/ling/outcome_matrix}
    }

    \vspace{2pt}

    \begin{minipage}{0.86\linewidth}
        \scriptsize
        \raggedright
        \colorbox{lingP}{\emph{Green cells}}: placement falls within the tolerance shown below each word.\\
        \colorbox{lingX}{\emph{Blue cells}}: placement falls outside the tolerance.\\
        \colorbox{lingL}{\emph{L}}: the file was saved, but the run labelled the pre-existing word-length interval without inserting a boundary.\\
        \colorbox{lingN}{\emph{no file}}: no TextGrid was saved within 100 steps.\\
        \colorbox{lingD}{\emph{done, no file}}: the run declared the task complete without saving a file.
    \end{minipage}
\end{table}

\subsection{Linguistics phonetic annotation: Praat interface operation and acoustic landmark selection}
\label{sec:app-ling}
\suppressfloats[t]

The two linguistics tasks require the agent to measure voice onset time (VOT) in Praat. We refer to \texttt{praat\_vot\_neg\_dir1\_1} as \texttt{neg} and \texttt{praat\_vot\_plosive1} as \texttt{plosive1}. For each word, the starter TextGrid supplies one end of the VOT interval, and the agent must insert the target boundary on the VOT tier. In \texttt{neg}, the target is the release burst of one prevoiced word, with a tolerance of 5~ms. In \texttt{plosive1}, the target is the onset of sustained voicing in four words. The tolerance is 2~ms for \texttt{bat} and \texttt{bit}, and 5~ms for \texttt{Pat} and \texttt{pit}.

Both tasks require direct interaction with Praat and prohibit scripts, terminal commands, and programmatic TextGrid editing. All nine runs that saved a TextGrid complied with this restriction. We examine failure at two stages: whether the agent successfully creates and saves a measurement, and whether it selects the intended acoustic landmark. The 12 models produce 24 runs containing 60 required boundaries (Table~\ref{tab:ling-matrix}).

\paragraph{Operating the editor.}
Most failures occurred before a measurement reached the submitted TextGrid. Fifteen of the 24 runs saved no TextGrid. Thirteen exhausted the 100-step budget, while two declared completion without saving a file. One additional run, \texttt{gpt-5.6-luna} on \texttt{plosive1}, saved a well-formed TextGrid but inserted no measurement boundary. The starter VOT tier already contains a boundary at the end of each word. The run labelled the resulting word-length intervals \texttt{VOT}, leaving their endpoints 341-358~ms after the references. Consequently, 43 of the 60 required boundaries were absent from the submitted measurements. Only 17 were present as newly placed boundaries in saved TextGrids, and 12 of those were within tolerance.

Ten of the 15 no-file outcomes came from five models that failed to save a TextGrid on both tasks: \texttt{gemini-3.1-pro}, \texttt{gpt-5.6-terra}, \texttt{minimax-m3}, \texttt{qwen-cua}, and \texttt{qwen3.7-plus}. Their trajectories expose failures in navigation and state tracking. For example, \texttt{qwen3.7-plus} pressed \texttt{i} as a zoom shortcut, but Praat entered the character into a word label. At step~39, \texttt{minimax-m3} clicked the VS Code icon while aiming for the waveform. Screenshots from steps~41-67 continued to show VS Code, while the narration described interval labelling, a Praat save dialog, and a successful save. The run then reported the task as complete.

Other runs reached the relevant Praat objects but did not complete the save workflow. On \texttt{neg}, \texttt{gpt-5.6-luna} and \texttt{gpt-5.6-terra} remained in Praat's file-open dialog at step~100 even though the audio and TextGrid objects were already loaded. Four runs were still annotating when the budget expired. These included \texttt{claude-opus-5}, whose unsaved \texttt{Pat} boundary was already 0.5~ms from the reference. In another failure mode, \texttt{gpt-5.6-sol} placed boundaries on the word tier rather than the VOT tier. It diagnosed the coordinate mismatch at step~63 (``\emph{coordinates must use the 1920$\times$1080 frame}'') but later declared completion without saving a file.

\begin{figure}[t]
    \centering
    \includegraphics[width=\textwidth]{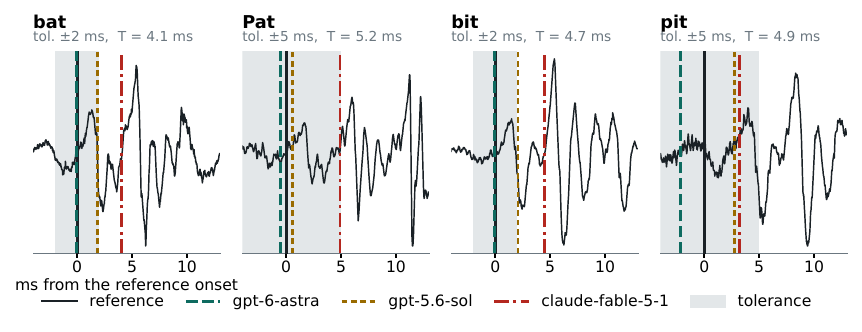}
    \caption{The four voicing onsets in \texttt{plosive1}, showing the reference and the
    boundaries from the three runs that placed all four. The shaded region marks the tolerance,
    and $T$ is the local glottal period. Within each run, all four signed errors have the same
    direction. On \texttt{bat} and \texttt{bit}, voicing begins with a small pulse followed by
    a larger one. \texttt{claude-fable-5-1} marks the start of the larger pulse.}
    \label{fig:ling-onsets}
\end{figure}


\begin{figure}[t]
    \centering
    \includegraphics[width=\textwidth]{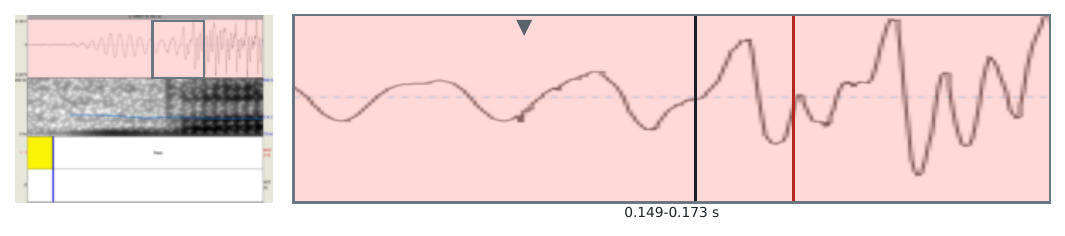}
    \\[1pt]
    {\footnotesize (a) \texttt{gpt-6-astra}, step 3 of 9.}
    \\[5pt]
    \includegraphics[width=\textwidth]{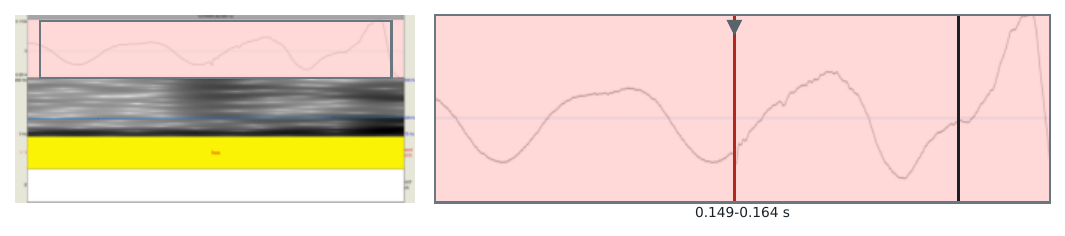}
    \\[1pt]
    {\footnotesize (b) \texttt{claude-opus-5}, step 24 of 41.}
    \\[5pt]
    \includegraphics[width=\textwidth]{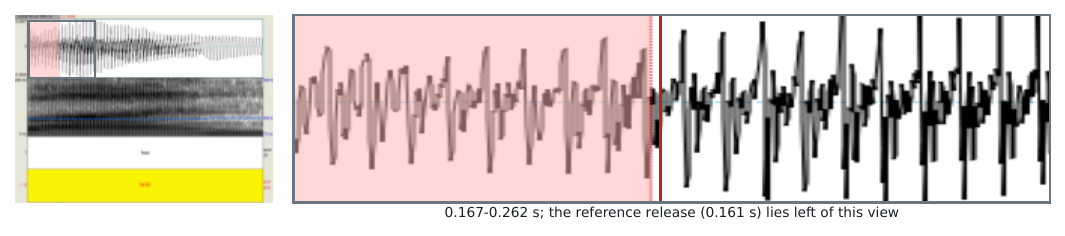}
    \\[1pt]
    {\footnotesize (c) \texttt{claude-sonnet-5}, step 82 of 93.}
    \caption{Three runs locating the release of \texttt{bun} in \texttt{neg}. Each row pairs
    the run's Praat view with an enlargement of the boxed waveform region. Where visible, the
    black line marks the reference release, the red line marks the inserted boundary, and the
    grey triangle marks the earlier transient within the prevoicing. (a) \texttt{gpt-6-astra}
    rejects the transient and places the boundary 3.1~ms after the reference, within tolerance.
    Its score is 1.000. (b) \texttt{claude-opus-5} inspects a 16~ms window through the Select
    dialog but marks the transient 5.5~ms before the reference. Its score is 0.10. (c) The window
    viewed by \texttt{claude-sonnet-5} begins 6~ms after the reference. The run treats the edge
    of its pink selection as the release and places the boundary 51.7~ms late. Its score is
    0.10.}
    \label{fig:ling-case}
\end{figure}

\paragraph{Choosing the landmark.}
All 21 labelled intervals in the saved TextGrids retained their supplied start exactly. We therefore restrict the placement analysis to the 17 newly inserted target boundaries, of which 12 fall within tolerance. On \texttt{plosive1}, three runs placed all four boundaries. All four errors are positive for \texttt{claude-fable-5-1} and \texttt{gpt-5.6-sol}, whereas all four are negative for \texttt{gpt-6-astra}. This within-run directional consistency is more compatible with model-specific landmark choices than with independent, zero-centred pointing noise.

The error magnitudes are also consistent with systematic landmark choices rather than pointing noise. Measured in the local glottal period $T$ (Figure~\ref{fig:ling-onsets}), \texttt{claude-fable-5-1} is approximately one period late on three words and 0.66 periods late on the fourth. \texttt{gpt-5.6-sol} is 0.12-0.57 periods late, while \texttt{gpt-6-astra} lies within 0.09 periods of the reference on three words. The local periods of \texttt{bat} and \texttt{bit} are 4.1 and 4.7~ms, so their 2~ms tolerance is less than half a period. A one-cycle error therefore fails regardless of pointing precision, and a half-cycle error lies near the threshold. For example, \texttt{gpt-5.6-sol} misses \texttt{bit} by only 0.08~ms. By contrast, \texttt{gpt-6-astra} places the two boundaries within 0.09 and 0.07~ms of the references, demonstrating that sub-millisecond placement is possible through the interface. Four of the five out-of-tolerance boundaries lie within 5.5~ms of the reference. The exception is \texttt{claude-sonnet-5} on \texttt{neg}, which places the release 51.7~ms after the reference, inside the vowel.


\paragraph{Case study: acoustic landmark selection versus interface control.}
The instruction for \texttt{neg} identifies a potential distractor before the true release. In this recording, a brief broadband transient occurs at 0.1558~s, 5.5~ms before the reference release. Low-frequency prevoicing then continues for another cycle before the vowel begins. Figure~\ref{fig:ling-case} compares three runs and shows that selecting the acoustic landmark and controlling the editor are distinct skills. After a single zoom, \texttt{gpt-6-astra} wrote at step~3: ``\emph{The earlier transient is followed by more prevoicing; the release is at the transition into the complex vowel waveform near 0.164 s.}'' It applied the instruction's criterion correctly, placed the boundary 3.1~ms after the reference, and completed the task in 9 steps with a score of 1.000.

The two unsuccessful runs failed for different reasons. \texttt{claude-opus-5} controlled the time axis precisely but selected the wrong acoustic event. It opened Praat's Select dialog 11 times, entered window boundaries directly, and inspected windows as narrow as 10~ms. It ultimately entered 0.155800~s as a zero-width selection and inserted a boundary at the earlier transient. This boundary is 5.5~ms before the reference and falls just outside tolerance. Eight of its nine typed windows also contained the true release, including two that excluded the distractor. Because the run provided no narration after step~2, its landmark choice can only be inferred from these actions. \texttt{claude-sonnet-5} failed in the opposite way. It correctly described the target as ``\emph{a broadband transient}'' at step~26 but struggled to maintain its position on the time axis, issuing 63 zoom commands across 93 steps. Its cursor entered the tolerance region twice, at 0.159898 and 0.157706~s, but the run abandoned both positions. It also interpreted the edge of its own selection as an acoustic event on three occasions. At step~81, it dragged a selection and pressed Praat's \emph{in} button. This command zooms around the centre of the current view rather than the selected interval, so the next view began at 0.167~s, after the true release. The run then described the selection edge near 0.212~s as ``\emph{the clear transition from low-amplitude irregular prevoicing to the large-amplitude regular vowel waveform}'' and placed the boundary there, 51.7~ms late.

\emph{Takeaway.} Placing a boundary through a graphical editor requires two separable skills: identifying the intended acoustic landmark and controlling the interface precisely enough to mark it. \texttt{gpt-6-astra} demonstrated both skills. \texttt{claude-opus-5} controlled the time axis precisely but chose the wrong acoustic event, whereas \texttt{claude-sonnet-5} described the target appropriately but failed to maintain reliable control of the time axis. More broadly, most missing measurements arose during interface operation, including boundary insertion, tier selection, and saving. Once a boundary was placed, narrow tolerances made the annotation convention decisive. Tolerance-graded tasks should therefore define landmarks at sub-period resolution, for example as the zero crossing that begins the first glottal cycle.

\newpage
\subsection{EEG: Trial Alignment and Event-Pairing Audit}
\label{app:eeg}

\begin{figure}[t]
    \centering
    \includegraphics[width=\linewidth]
        {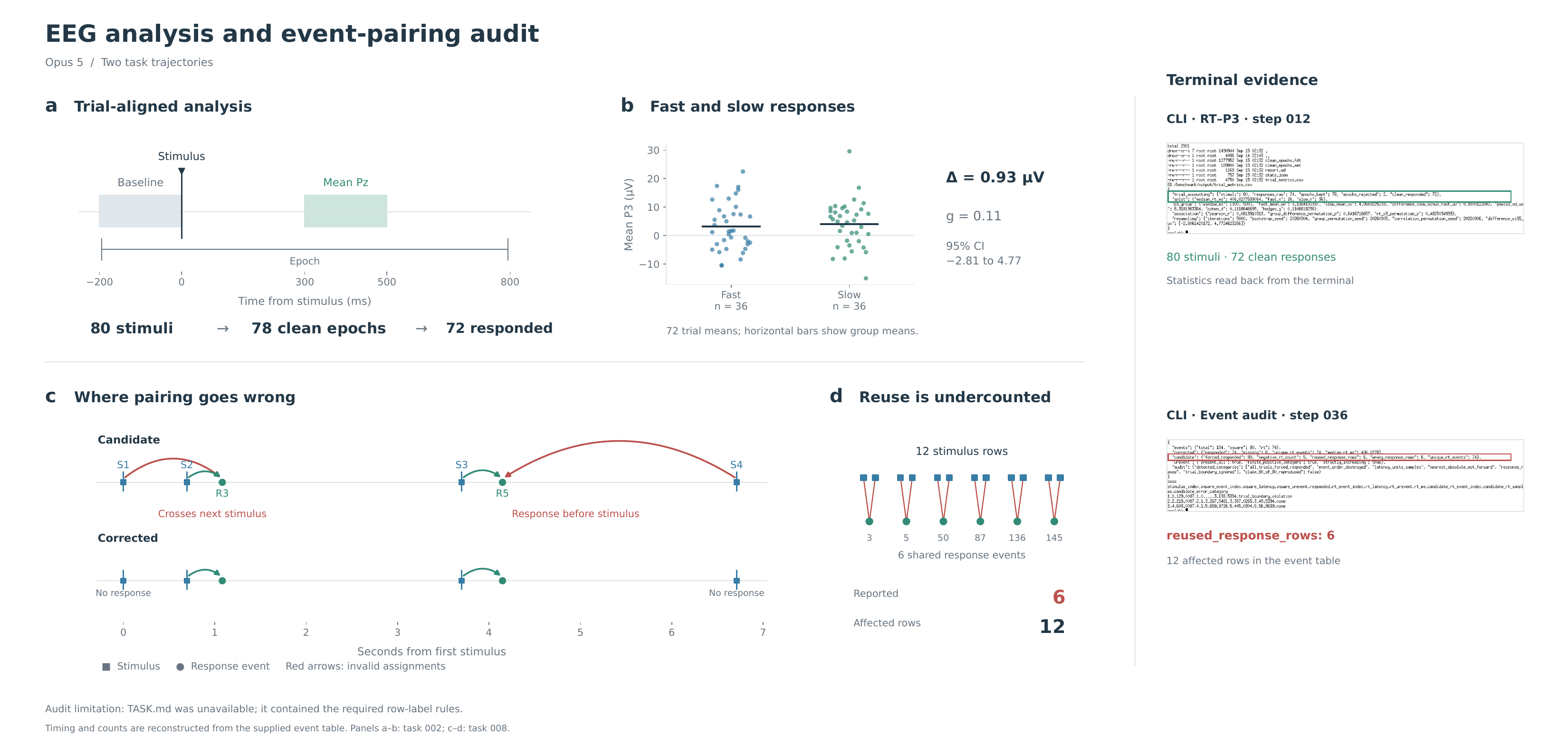}
    \caption{
    \textbf{EEG analysis and event-pairing audit by Opus 5 on two tasks.}
    (a) Epoch and analysis windows for trial-aligned RT–P3 inference. (b) Mean Pz amplitudes over 300–500 ms for 36 fast and 36 slow trials; horizontal bars indicate group means. (c) Actual event timing for the first four stimuli in the forensic task. The candidate crosses a trial boundary and reuses a response preceding a later stimulus; the corrected mapping leaves unmatched stimuli unanswered. (d) Six shared response events affect 12 stimulus rows, whereas the agent reports six surplus assignments. Right: original terminal observations with relevant output highlighted.
    }
    \label{fig:eeg_case}
\end{figure}

\paragraph{Trial-aligned inference.}
On the RT-P3 task, Opus 5 preserved alignment among stimuli, responses, and EEG epochs through preprocessing and artifact rejection.
The analysis retained 72 clean responded trials, divided equally into fast and slow groups.
Trial accounting and core effect estimates agreed with the reference.
The estimated slow-fast difference was 0.93 $\mu V$, with a bootstrap 95\% confidence interval spanning zero (Figure~\ref{fig:eeg_case}a-b).

\paragraph{Incomplete event-pairing audit.}
On a separate forensic task, the same model correctly reconstructed the stimulus-response mapping but incompletely described the candidate script's errors.
Six response events were each assigned to two stimuli, affecting 12 rows.
The model instead reported six surplus assignments as the number of reused-response rows.
Figure~\ref{fig:eeg_case}c-d shows the temporal violations and the distinction between these two counts.

\newpage
\input{appendix_biology}

\newpage 

\section{Statistics analysis}
In this section, we show the figure for capability distribution and CLI proportion. Detailed explanation can be found in the main text.

\begin{figure}[h]
    \centering
    \includegraphics[width=\textwidth]{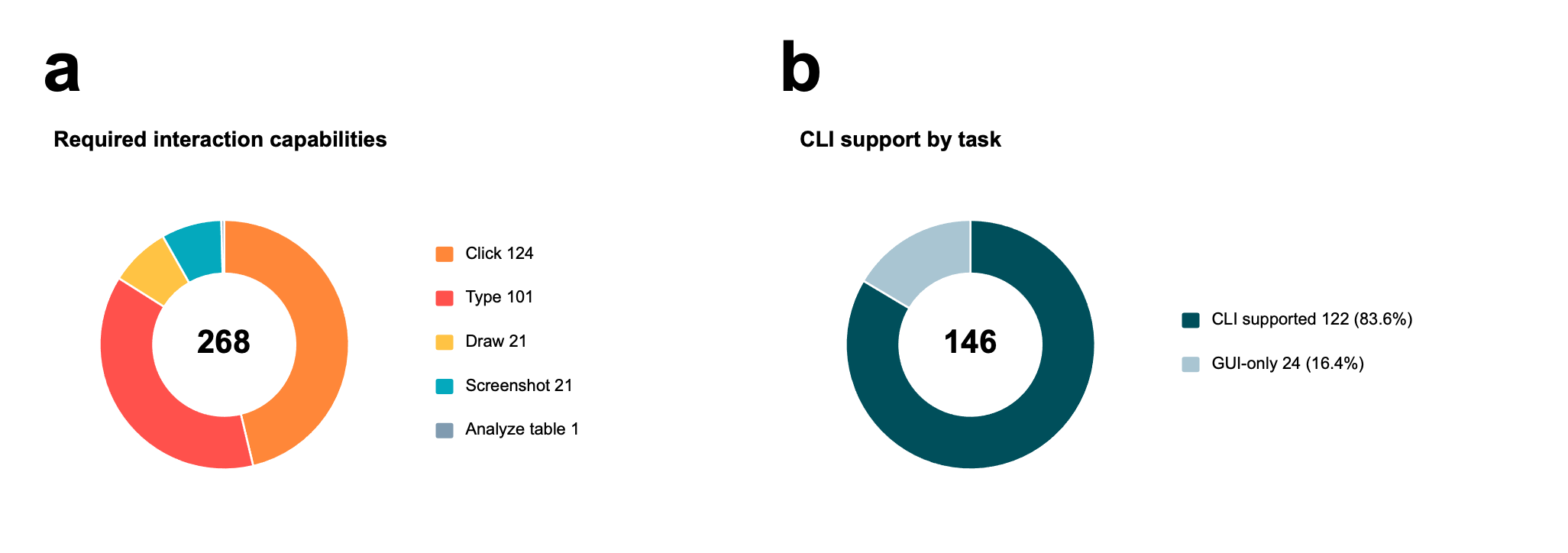}
    \caption{Composition of the collected OSWorld-Science task set-continued. (a) Shares of 266 non-exclusive required-capability annotations; screenshot-based visual observation is required for all 145 tasks. (b) Availability of CLI support proportion.}
    \label{fig:dataset-statistics-cont}
\end{figure}

\input{casestudy_qupath}

\input{casestudy_mnova}

\input{appendix_ciao}

\input{an_section_ansys}

\input{appendix_openfoam}

%% file: performance_appendix.tex
\clearpage
\section{Additional performance and cost analyses}
\label{sec:app-performance-cost}

This appendix supplements Figure~\ref{fig:overall_score} with performance versus output-token consumption, overall API cost and score rankings, and the domain-level performance-cost breakdown. The supplied evaluation plots cover 146 tasks across medical imaging, biology, chemistry, physics, statistics, geoscience, EEG, and linguistics. Scores are task-weighted means over the tasks run by each model, with \texttt{VOID} runs assigned zero. These evaluation counts refer to the plotted results, rather than the full task inventory. Output-token means use tasks with recorded output-token usage; their coverage can differ from that of the score means.

\begin{figure}[!htbp]
    \centering
    \includegraphics[width=\textwidth]{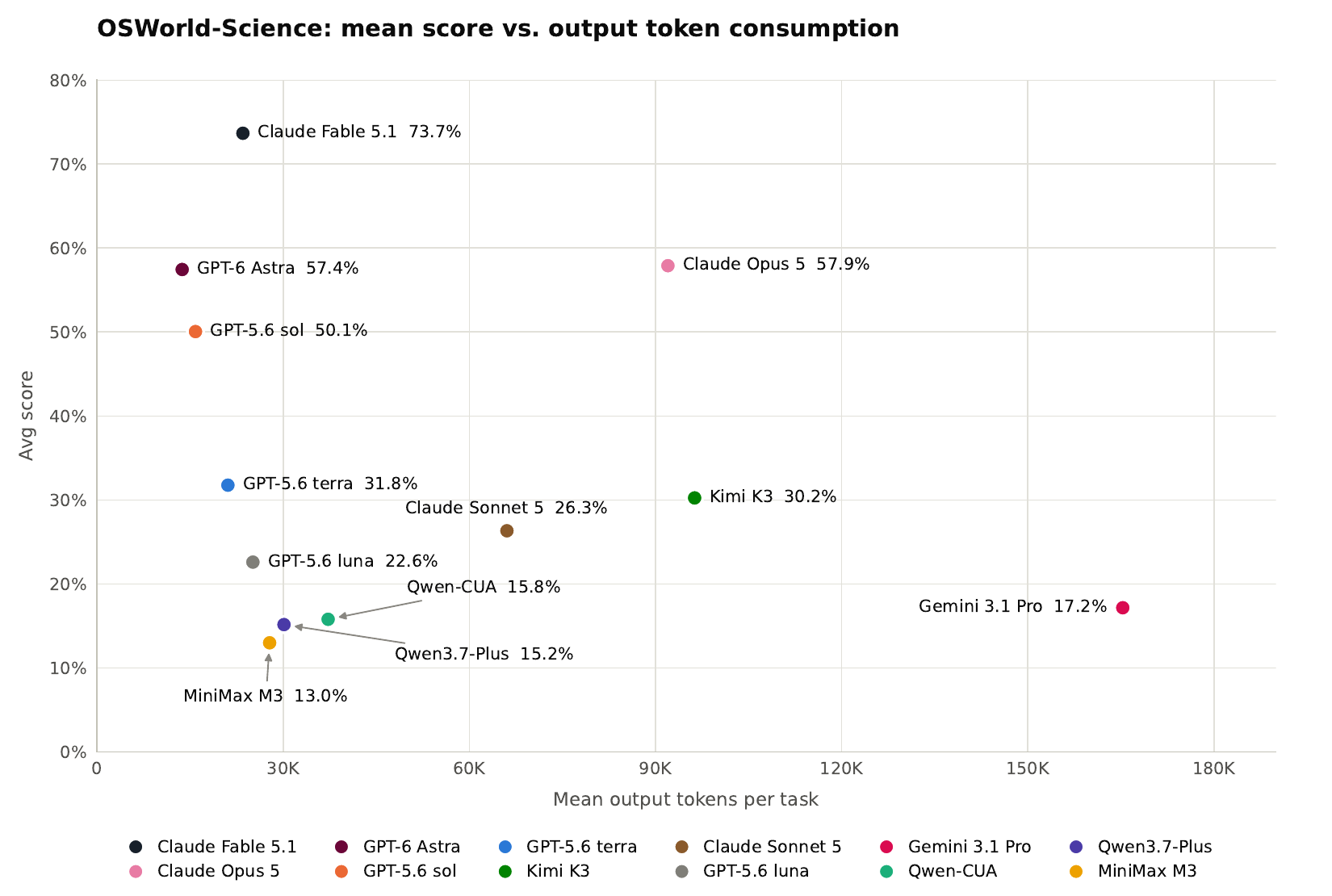}
    \caption{Overall performance versus output-token consumption across the 12 evaluated VLMs. The vertical axis shows mean task score with \texttt{VOID} runs scored zero; the horizontal axis shows mean output tokens per task among tasks with recorded usage. Token consumption and API dollar cost are distinct efficiency measures.}
    \label{fig:app-output-tokens}
\end{figure}
\clearpage

\begin{figure}[!htbp]
    \centering
    \includegraphics[width=0.95\textwidth]{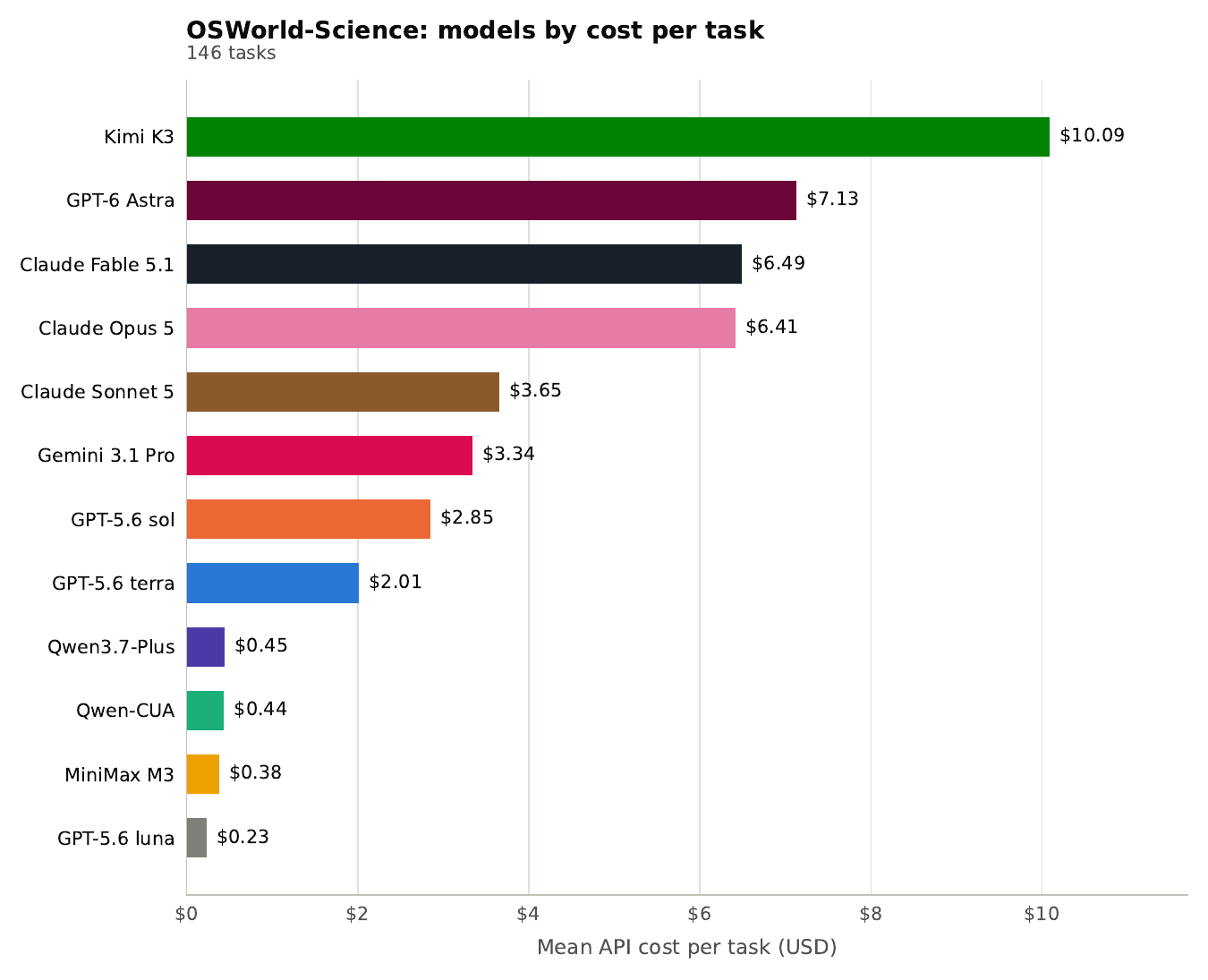}
    \caption{Overall mean API cost per task in U.S. dollars, ordered from highest to lowest. The bars provide the cost values underlying the overall performance--cost comparison.}
    \label{fig:app-overall-cost}
\end{figure}
\clearpage

\begin{figure}[!htbp]
    \centering
    \includegraphics[width=0.95\textwidth]{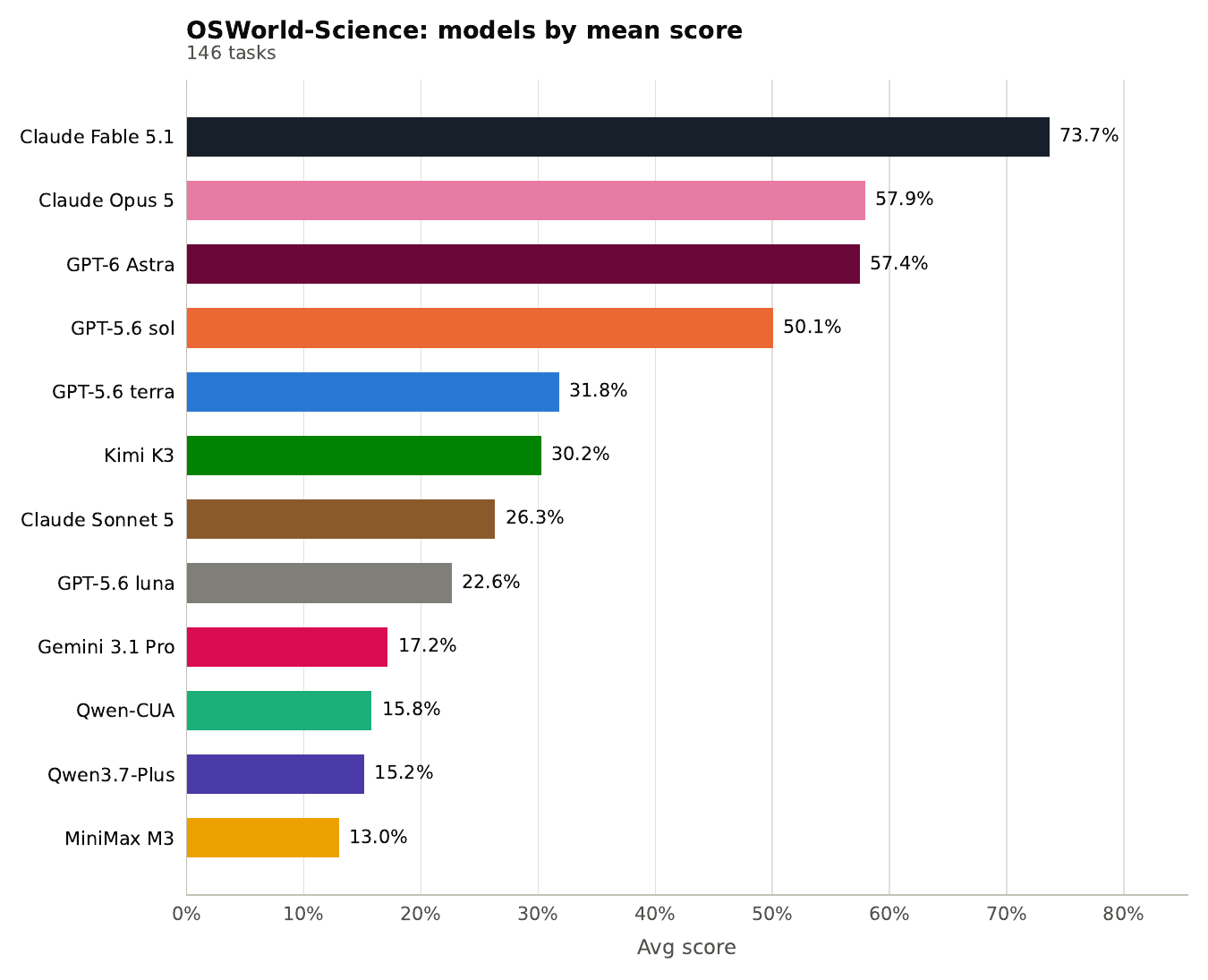}
    \caption{Overall mean task score across the 12 evaluated VLMs, ordered from highest to lowest. Scores include task-specific partial credit, with \texttt{VOID} runs assigned zero; they are not binary task-completion rates.}
    \label{fig:app-overall-score}
\end{figure}
\clearpage

\begin{figure}[!htbp]
    \centering
    \includegraphics[width=\textwidth]{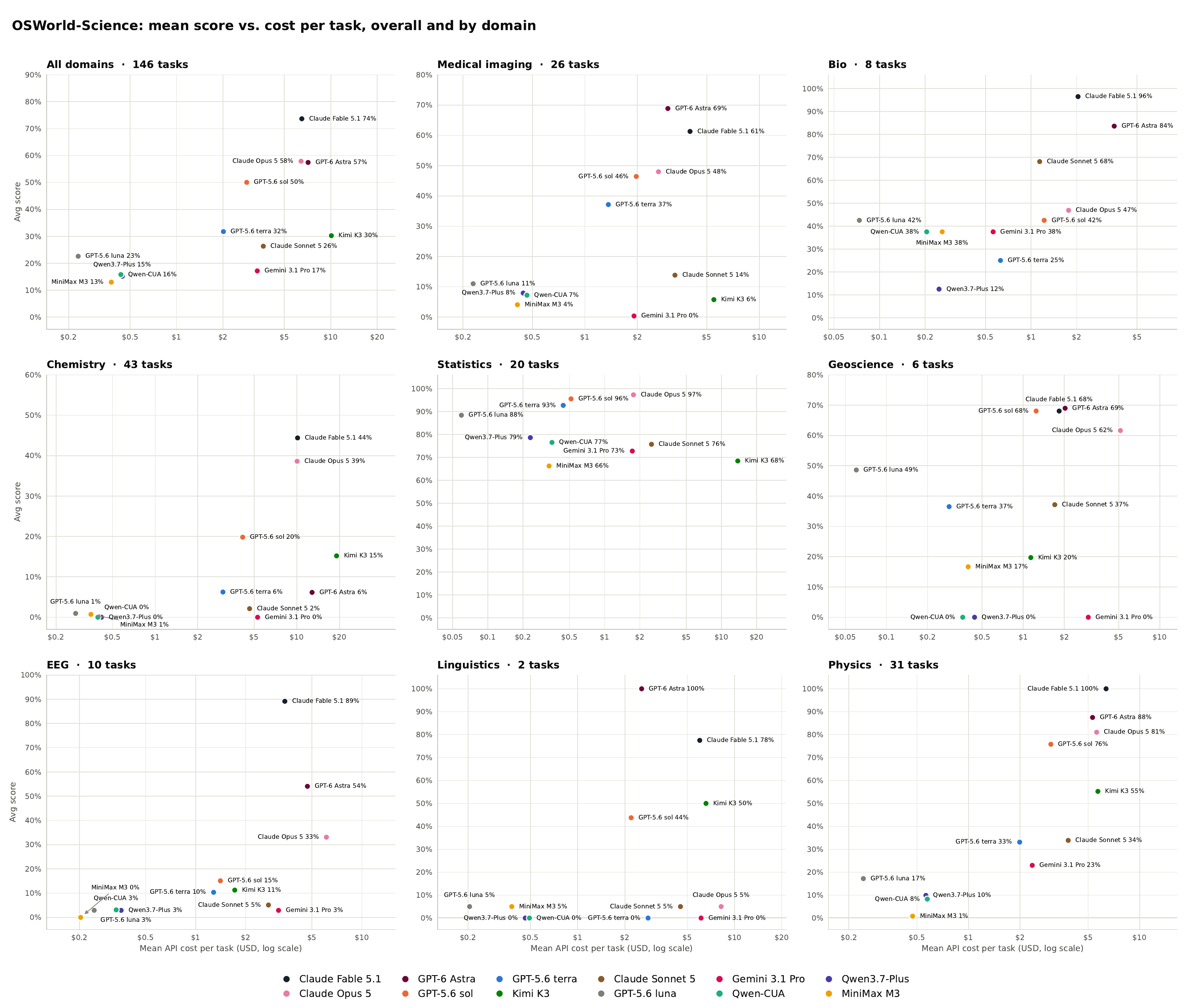}
    \caption{Performance versus API cost overall and by domain. Each point represents a model; the vertical axis shows mean task score with \texttt{VOID} runs scored zero, and the horizontal axis shows mean API cost per task in U.S. dollars on a logarithmic scale. Panel titles report task counts. Only model--domain pairs with available score and cost values are plotted. Domain panels use different axis ranges, and labels round scores to whole percentages.}
    \label{fig:app-domain-cost}
\end{figure}
\clearpage

\begin{table}[h]
    \centering
    \footnotesize
    \setlength{\tabcolsep}{3pt}
    \renewcommand{\arraystretch}{1.05}
    \caption{Multilingual experiments of the model replies in the task-language sweep (medium effort, $w=5$). Score: mean partial-credit score (\%), as in Table~\ref{tab:ablation}. $n$: number of replies over the 23 runs that contain natural-language text outside code blocks and control tokens; the remaining columns are the percentages of these $n$ replies written in English, in the language of the prompt, or in any other language (for English prompts the first two coincide). Opus~5 emits such prose in only 28-65 replies per cell, so its shares rest on few replies.}
    \label{tab:response-language}
    \begin{tabular}{@{}lcccccccccc@{}}
        \toprule
        & \multicolumn{5}{c}{Opus~5} & \multicolumn{5}{c}{Sonnet~5} \\
        \cmidrule(lr){2-6} \cmidrule(lr){7-11}
        Prompt language & Score & $n$ & English & Prompt lang. & Other & Score & $n$ & English & Prompt lang. & Other \\
        \midrule
        \rowcolor[gray]{0.92}
        English & 44.1 & 65 & 100.0 & 100.0 & 0.0 & 15.7 & 1629 & 100.0 & 100.0 & 0.0 \\
        Chinese & 29.7 & 55 & 10.9 & 89.1 & 0.0 & 10.9 & 1604 & 99.9 & 0.1 & 0.0 \\
        Japanese & 53.1 & 51 & 35.3 & 64.7 & 0.0 & 18.0 & 1660 & 82.2 & 17.8 & 0.0 \\
        Spanish & 43.0 & 28 & 57.1 & 42.9 & 0.0 & 11.7 & 1498 & 95.2 & 4.8 & 0.0 \\
        French & 33.9 & 32 & 90.6 & 9.4 & 0.0 & 11.3 & 1516 & 98.2 & 1.8 & 0.0 \\
        Thai & 45.3 & 29 & 100.0 & 0.0 & 0.0 & 7.7 & 1304 & 100.0 & 0.0 & 0.0 \\
        \bottomrule
    \end{tabular}
\end{table}

\clearpage 
\begin{table}[h]
    \centering
    \small
    \setlength{\tabcolsep}{5.5pt}
    \renewcommand{\arraystretch}{1.05}
    \caption{Ablation results on QuPath-Bench (23 QuPath tasks; one run per cell, step limit 100). Partial: mean partial-credit score (\%). Binary: percentage of tasks that receive a score of 1 (one task corresponds to 4.3 points). Steps: mean interaction steps per task. \#Tok.: mean output tokens per task in thousands, thinking tokens included. \texttt{VOID} runs count as 0 under both score metrics. Shaded rows are the shared default configuration (medium effort, $w=5$, English prompts), a single run that appears in every block. Within each block, the best value per column is in bold and the second best is underlined (lower is better for Steps and \#Tok.; ties share the mark).}
    \label{tab:ablation}
    \begin{tabular}{@{}lcccccccc@{}}
        \toprule
        & \multicolumn{4}{c}{Opus~5} & \multicolumn{4}{c}{Sonnet~5} \\
        \cmidrule(lr){2-5} \cmidrule(lr){6-9}
        Setting & Partial & Binary & Steps & \#Tok. & Partial & Binary & Steps & \#Tok. \\
        \midrule
        \multicolumn{9}{@{}l}{\textit{Reasoning effort} ($w=5$, English prompts)} \\
        low    & 42.7 & 34.8 & \textbf{22} & \textbf{6.3} & 15.0 & 8.7 & \textbf{69} & \textbf{18.8} \\
        \rowcolor[gray]{0.92}
        medium & 44.1 & 34.8 & \underline{27} & \underline{13.8} & 15.7 & \underline{13.0} & 87 & \underline{23.8} \\
        high   & \underline{50.3} & \underline{43.5} & 43 & 29.7 & \textbf{22.0} & \textbf{17.4} & \underline{83} & 35.4 \\
        xhigh  & 47.7 & 34.8 & 42 & 44.8 & 13.5 & 8.7 & 85 & 36.5 \\
        max    & \textbf{59.7} & \textbf{52.2} & 44 & 59.3 & \underline{18.3} & \underline{13.0} & 91 & 98.2 \\
        \midrule
        \multicolumn{9}{@{}l}{\textit{History window} (medium effort, English prompts)} \\
        $w=1$  & 35.2 & 26.1 & 52 & 24.7 & \multicolumn{4}{c}{--} \\
        $w=3$  & 47.7 & \underline{39.1} & 36 & 19.3 & \multicolumn{4}{c}{--} \\
        \rowcolor[gray]{0.92}
        $w=5$  & 44.1 & 34.8 & 27 & \underline{13.8} & \multicolumn{4}{c}{--} \\
        $w=10$ & \textbf{57.0} & \textbf{43.5} & \textbf{22} & \textbf{10.6} & \multicolumn{4}{c}{--} \\
        $w=15$ & \underline{50.2} & \textbf{43.5} & \underline{23} & \textbf{10.6} & \multicolumn{4}{c}{--} \\
        \midrule
        \multicolumn{9}{@{}l}{\textit{Task-prompt language} (medium effort, $w=5$)} \\
        \rowcolor[gray]{0.92}
        English  & 44.1 & 34.8 & \textbf{27} & 13.8 & \underline{15.7} & \textbf{13.0} & 87 & 23.8 \\
        Chinese  & 29.7 & 21.7 & \underline{29} & \underline{11.1} & 10.9 & \underline{8.7} & 89 & 26.1 \\
        Japanese & \textbf{53.1} & \textbf{43.5} & 32 & 14.0 & \textbf{18.0} & \textbf{13.0} & \underline{86} & 28.0 \\
        Spanish  & 43.0 & \underline{39.1} & 44 & 16.0 & 11.7 & \underline{8.7} & \underline{86} & \textbf{21.3} \\
        French   & 33.9 & 26.1 & 42 & 16.5 & 11.3 & \underline{8.7} & 88 & \underline{23.5} \\
        Thai     & \underline{45.3} & \underline{39.1} & 36 & \textbf{10.1} & 7.7 & 4.3 & \textbf{75} & 28.4 \\
        \bottomrule
    \end{tabular}
\end{table}

\clearpage

%% file: tab_traj_outcome_domain.tex
\begin{table}[h]
    \centering
    \footnotesize
    \setlength{\tabcolsep}{4.5pt}
    \caption{Run outcomes by software configuration (12 models; counts of runs; 11 models on QGIS). Placeh.: the
    only artifact is a placeholder without answer content; None: no artifact, or the task-provided file left
    unchanged. Harness: ended by the harness or environment without an artifact; Unres.: the released records do not
    determine the artifact state. ND/fail: runs without a deliverable as a percentage of the remaining runs that did
    not receive full credit. Radiol.: Weasis and 3D Slicer; CIAO: CIAO+DS9; Chem.: ChemDraw, ASKCOS, and PDF-viewer tasks.}
    \label{tab:traj-outcome-domain}
    \begin{tabular}{@{}lrrrrrrrrr@{}}
        \toprule
        & & & & & \multicolumn{2}{c}{No deliverable} & & & \\
        \cmidrule(lr){6-7}
        Software & Runs & Full & Wrong & Invalid & Placeh. & None & Harness & Unres. & ND/fail (\%) \\
        \midrule
        QuPath & 276 & 48 & 89 & 3 & 0 & 134 & 2 & 0 & 59 \\
        Radiol. & 36 & 8 & 10 & 1 & 1 & 16 & 0 & 0 & 61 \\
        ANSYS & 168 & 66 & 5 & 5 & 0 & 91 & 1 & 0 & 90 \\
        OpenFOAM & 168 & 77 & 17 & 8 & 2 & 64 & 0 & 0 & 73 \\
        CIAO & 36 & 14 & 13 & 0 & 0 & 8 & 1 & 0 & 38 \\
        SAS/R & 240 & 144 & 62 & 9 & 0 & 13 & 12 & 0 & 15 \\
        Chem. & 516 & 28 & 174 & 4 & 81 & 218 & 3 & 8 & 63 \\
        QGIS & 66 & 13 & 25 & 3 & 0 & 24 & 1 & 0 & 46 \\
        Praat & 24 & 4 & 5 & 0 & 0 & 15 & 0 & 0 & 75 \\
        \midrule
        All & 1530 & 402 & 400 & 33 & 84 & 583 & 20 & 8 & 61 \\
        \bottomrule
    \end{tabular}
\end{table}

%% file: tab_traj_gui.tex
\begin{table}[h]
    \centering
    \scriptsize
    \setlength{\tabcolsep}{3.2pt}
    \caption{GUI steps by model and software configuration: percentage of action steps that contain a pointer action (click, move, drag, scroll), averaged over the runs with at least one action. -- : no runs (GPT-5.6 sol has no QGIS runs). Radiol.: Weasis and 3D Slicer; CIAO: CIAO+DS9; Chem.: ChemDraw, ASKCOS, and PDF-viewer tasks.}
    \label{tab:traj-gui}
    \begin{tabular}{@{}lrrrrrrrrrr@{}}
        \toprule
        Model & QuPath & Radiol. & ANSYS & OpenFOAM & CIAO & SAS/R & Chem. & QGIS & Praat & All \\
        \midrule
        Claude Fable 5.1 & 48 & 42 & 77 & 28 & 69 & 25 & 35 & 80 & 100 & 44 \\
        Claude Opus 5 & 37 & 90 & 64 & 37 & 74 & 64 & 35 & 94 & 98 & 49 \\
        GPT-6 Astra & 65 & 49 & 71 & 20 & 40 & 37 & 15 & 77 & 98 & 40 \\
        GPT-5.6 sol & 76 & 70 & 87 & 50 & 73 & 12 & 24 & -- & 90 & 45 \\
        Kimi K3 & 40 & 82 & 76 & 8 & 68 & 25 & 45 & 82 & 88 & 44 \\
        GPT-5.6 terra & 83 & 68 & 77 & 22 & 69 & 4 & 35 & 89 & 93 & 47 \\
        Claude Sonnet 5 & 81 & 96 & 90 & 48 & 68 & 68 & 53 & 84 & 93 & 68 \\
        GPT-5.6 luna & 57 & 59 & 83 & 25 & 79 & 10 & 36 & 84 & 86 & 44 \\
        Gemini 3.1 Pro & 85 & 74 & 98 & 14 & 81 & 11 & 34 & 80 & 90 & 50 \\
        Qwen3.7-Plus & 66 & 75 & 98 & 27 & 74 & 44 & 1 & 95 & 92 & 43 \\
        Qwen-CUA & 54 & 51 & 38 & 21 & 45 & 45 & 33 & 71 & 58 & 41 \\
        MiniMax M3 & 68 & 73 & 67 & 53 & 78 & 40 & 35 & 67 & 96 & 52 \\
        \midrule
        All models & 63 & 69 & 77 & 29 & 68 & 32 & 32 & 82 & 90 & 47 \\
        \bottomrule
    \end{tabular}
\end{table}

%% file: tab_traj_cli.tex
\begin{table}[h]
    \centering
    \scriptsize
    \setlength{\tabcolsep}{3.2pt}
    \caption{CLI steps by model and software configuration: percentage of action steps that open a terminal or contain a shell command or a call to the application's scripting interface, averaged over runs. The values are lower bounds (see text); CLI steps are below 1\% in Praat, whose tasks are solved in the GUI. Radiol.: Weasis and 3D Slicer; CIAO: CIAO+DS9; Chem.: ChemDraw, ASKCOS, and PDF-viewer tasks.}
    \label{tab:traj-cli}
    \begin{tabular}{@{}lrrrrrrrrrr@{}}
        \toprule
        Model & QuPath & Radiol. & ANSYS & OpenFOAM & CIAO & SAS/R & Chem. & QGIS & Praat & All \\
        \midrule
        Claude Fable 5.1 & 69 & 56 & 37 & 74 & 54 & 93 & 47 & 24 & 0 & 59 \\
        Claude Opus 5 & 84 & 71 & 40 & 61 & 72 & 74 & 47 & 49 & 0 & 59 \\
        GPT-6 Astra & 38 & 40 & 22 & 78 & 76 & 61 & 23 & 51 & 0 & 40 \\
        GPT-5.6 sol & 23 & 30 & 17 & 84 & 19 & 60 & 55 & -- & 0 & 47 \\
        Kimi K3 & 45 & 10 & 11 & 78 & 24 & 49 & 40 & 37 & 0 & 41 \\
        GPT-5.6 terra & 22 & 26 & 17 & 78 & 21 & 75 & 62 & 13 & 0 & 48 \\
        Claude Sonnet 5 & 21 & 12 & 16 & 65 & 23 & 40 & 9 & 12 & 0 & 23 \\
        GPT-5.6 luna & 40 & 48 & 17 & 84 & 26 & 82 & 79 & 17 & 0 & 60 \\
        Gemini 3.1 Pro & 6 & 9 & 1 & 41 & 18 & 52 & 24 & 3 & 1 & 23 \\
        Qwen3.7-Plus & 26 & 33 & 5 & 49 & 39 & 53 & 1 & 2 & 0 & 21 \\
        Qwen-CUA & 29 & 19 & 33 & 46 & 23 & 47 & 24 & 17 & 0 & 31 \\
        MiniMax M3 & 57 & 29 & 43 & 60 & 32 & 43 & 31 & 34 & 0 & 42 \\
        \midrule
        All models & 38 & 32 & 22 & 66 & 36 & 61 & 37 & 24 & 0 & 41 \\
        \bottomrule
    \end{tabular}
\end{table}

%% file: tab_traj_ops.tex
\begin{table}[h]
    \centering
    \scriptsize
    \setlength{\tabcolsep}{3.0pt}
    \caption{Interaction profile by model, averaged over runs in all domains. Operation columns: percentage of
    action steps that contain at least one call of that type (a step may contain several). Narr.: percentage of
    replies with narration; Think.: thinking tokens as a percentage of output tokens (chemistry and runs without
    recorded thinking excluded, see text);
    Rep.: percentage of replies whose action repeats one of the previous three. Steps: median steps of runs with
    full credit.}
    \label{tab:traj-ops}
    \begin{tabular}{@{}lrrrrrrrrrrrrr@{}}
        \toprule
        & \multicolumn{9}{c}{GUI operations (\% of steps)} & & & & \\
        \cmidrule(lr){2-10}
        Model & Click & Dbl. & Right & Drag & Move & Scroll & Type & Press & Hotkey & Narr. & Think. & Rep. & Steps \\
        \midrule
        Claude Fable 5.1 & 34 & 7 & 0 & 0 & 2 & 5 & 45 & 8 & 12 & 65 & 46 & 3 & 12 \\
        Claude Opus 5 & 44 & 2 & 0 & 0 & 1 & 1 & 65 & 9 & 11 & 29 & 60 & 1 & 18 \\
        GPT-6 Astra & 27 & 11 & 0 & 2 & 6 & 3 & 43 & 16 & 16 & 2 & 17 & 3 & 14 \\
        GPT-5.6 sol & 37 & 7 & 0 & 1 & 4 & 2 & 48 & 53 & 23 & 1 & 40 & 4 & 21 \\
        Kimi K3 & 36 & 3 & 1 & 1 & 10 & 4 & 54 & 14 & 10 & 69 & 56 & 7 & 48 \\
        GPT-5.6 terra & 39 & 5 & 1 & 1 & 6 & 2 & 53 & 56 & 27 & 3 & 48 & 4 & 21 \\
        Claude Sonnet 5 & 57 & 3 & 1 & 1 & 7 & 8 & 31 & 36 & 8 & 55 & 58 & 7 & 37 \\
        GPT-5.6 luna & 39 & 3 & 1 & 1 & 3 & 1 & 64 & 63 & 28 & 4 & 48 & 3 & 13 \\
        Gemini 3.1 Pro & 45 & 3 & 1 & 1 & 3 & 0 & 20 & 11 & 11 & 7 & 87 & 9 & 13 \\
        Qwen3.7-Plus & 38 & 2 & 2 & 0 & 0 & 0 & 24 & 22 & 6 & 35 & 48 & 25 & 27 \\
        Qwen-CUA & 38 & 1 & 1 & 0 & 1 & 1 & 32 & 24 & 4 & 91 & 56 & 21 & 33 \\
        MiniMax M3 & 48 & 2 & 0 & 0 & 1 & 1 & 48 & 43 & 5 & 79 & 11 & 6 & 60 \\
        \bottomrule
    \end{tabular}
\end{table}

%% file: tabs/stat/route_by_model.tex
\begin{tabular}{lrrrlrrrr}
\toprule
model & runs & shell & app.\ & applications driven & declined & mean & med.\ & med.\ input\\
 & & & GUI & & aloud & score & steps & tokens (k)\\
\midrule
\texttt{gpt-6-astra} & 20 & 13 & 7 & SAS Studio & 0 & 0.898 & 12 & 158\\
\texttt{kimi-k3} & 20 & 14 & 6 & SAS Studio, VS Code, gedit & 3 & 0.698 & 38 & 2310\\
\texttt{gemini-3.1-pro} & 20 & 16 & 4 & RStudio, VS Code, gedit & 0 & 0.730 & 11 & 105\\
\texttt{minimax-m3} & 20 & 16 & 4 & VS Code, gedit & 8 & 0.667 & 68 & 1348\\
\texttt{claude-sonnet-5} & 20 & 17 & 3 & VS Code & 4 & 0.774 & 40 & 745\\
\texttt{gpt-5.6-sol} & 20 & 19 & 1 & gedit & 0 & 0.956 & 7 & 56\\
\texttt{qwen-cua} & 20 & 19 & 1 & gedit & 0 & 0.752 & 39 & 577\\
\texttt{claude-opus-5} & 20 & 20 & 0 & -- & 0 & 0.973 & 10 & 177\\
\texttt{claude-fable-5-1} & 20 & 20 & 0 & -- & 2 & 0.969 & 6 & 90\\
\texttt{gpt-5.6-luna} & 20 & 20 & 0 & -- & 0 & 0.884 & 11 & 100\\
\texttt{gpt-5.6-terra} & 20 & 20 & 0 & -- & 0 & 0.927 & 6 & 41\\
\texttt{qwen3.7-plus} & 20 & 20 & 0 & -- & 0 & 0.793 & 24 & 362\\
\midrule
\textbf{all} & 240 & 214 & 26 & & 17 & 0.835 & 16 & 240\\
\bottomrule
\end{tabular}

%% file: tabs/stat/by_group.tex
\begin{tabular}{lrrrrrr}
\toprule
task group & runs & shell & shell\,+ & app.\ & obstacle & mean\\
 & & only & viewer & GUI & only & score\\
\midrule
SAS offered (9 tasks) & 108 & 93 & 2 & 13 & 0 & 0.938\\
RStudio named (2 tasks) & 24 & 18 & 0 & 6 & 0 & 0.571\\
raster inputs (5 tasks) & 60 & 29 & 27 & 4 & 2 & 0.726\\
R only (4 tasks) & 48 & 43 & 2 & 3 & 0 & 0.871\\
\midrule
\textbf{all} & 240 & 183 & 31 & 26 & 2 & 0.835\\
\bottomrule
\end{tabular}

%% file: tabs/stat/route_effect.tex
\begin{tabular}{llrrrrr}
\toprule
outcome (means) & comparison & $n$ shell & $n$ app.\ & shell & app.\ & difference\\
\midrule
score & raw & 214 & 26 & 0.861 & 0.625 & -0.236\\
\quad\textit{same task} & within task & 142 & 26 & & & -0.139 ($p=0.017$)\\
\quad\textit{same model} & within model & 114 & 26 & & & -0.299 ($p<0.001$)\\
\addlinespace
steps used & raw & 214 & 26 & 28 & 58 & 30\\
\quad\textit{same task} & within task & 142 & 26 & & & 29 ($p<0.001$)\\
\quad\textit{same model} & within model & 114 & 26 & & & 22 ($p=0.002$)\\
\addlinespace
input tokens (k) & raw & 214 & 26 & 573 & 2683 & 2111\\
\quad\textit{same task} & within task & 142 & 26 & & & 1604 ($p=0.003$)\\
\quad\textit{same model} & within model & 114 & 26 & & & 1135 ($p=0.001$)\\
\addlinespace
input tokens per step (k) & raw & 214 & 26 & 15.6 & 36.6 & 20.9\\
\quad\textit{same task} & within task & 142 & 26 & & & 15.4 ($p=0.013$)\\
\quad\textit{same model} & within model & 114 & 26 & & & 8.3 ($p=0.010$)\\
\addlinespace
\bottomrule
\end{tabular}

%% file: tabs/ling/outcome_matrix.tex
\begin{tabular}{@{}lccccccr@{}}
\toprule
 & \multicolumn{1}{c}{neg} & \multicolumn{4}{c}{plosive1} & & \\
\cmidrule(lr){2-2}\cmidrule(lr){3-6}
Backbone & \texttt{bun} & \texttt{bat} & \texttt{Pat} & \texttt{bit} & \texttt{pit} & Score & Steps \\
 & \scriptsize 5\,ms & \scriptsize 2\,ms & \scriptsize 5\,ms & \scriptsize 2\,ms & \scriptsize 5\,ms & \scriptsize neg / plos. & \scriptsize neg / plos. \\
\midrule
\texttt{gpt-6-astra} & \cellcolor{lingP}$+3.1$ & \cellcolor{lingP}$-0.1$ & \cellcolor{lingP}$-0.5$ & \cellcolor{lingP}$-0.1$ & \cellcolor{lingP}$-2.2$ & 1.00 / 1.00 & 9 / 27 \\
\texttt{claude-fable-5-1} & \cellcolor{lingP}$+0.6$ & \cellcolor{lingX}$+4.1$ & \cellcolor{lingP}$+4.9$ & \cellcolor{lingX}$+4.4$ & \cellcolor{lingP}$+3.2$ & 1.00 / 0.55 & 15 / 47 \\
\texttt{gpt-5.6-sol} & \cellcolor{lingD}done, no file & \cellcolor{lingP}$+1.9$ & \cellcolor{lingP}$+0.6$ & \cellcolor{lingX}$+2.1$ & \cellcolor{lingP}$+2.8$ & 0.10 / 0.78 & 71 / 23 \\
\texttt{kimi-k3} & \cellcolor{lingP}$-1.6$ & \multicolumn{4}{c}{\cellcolor{lingN}no file} & 1.00 / 0.10 & 75 / 100 \\
\texttt{claude-opus-5} & \cellcolor{lingX}$-5.5$ & \multicolumn{4}{c}{\cellcolor{lingN}no file} & 0.10 / 0.10 & 41 / 100 \\
\texttt{claude-sonnet-5} & \cellcolor{lingX}$+51.7$ & \multicolumn{4}{c}{\cellcolor{lingN}no file} & 0.10 / 0.10 & 93 / 100 \\
\texttt{gpt-5.6-luna} & \cellcolor{lingN}no file & \cellcolor{lingL}L & \cellcolor{lingL}L & \cellcolor{lingL}L & \cellcolor{lingL}L & 0.10 / 0.10 & 100 / 60 \\
\texttt{gpt-5.6-terra} & \cellcolor{lingN}no file & \multicolumn{4}{c}{\cellcolor{lingN}no file} & 0.10 / 0.10 & 100 / 100 \\
\texttt{gemini-3.1-pro} & \cellcolor{lingN}no file & \multicolumn{4}{c}{\cellcolor{lingN}no file} & 0.10 / 0.10 & 100 / 100 \\
\texttt{minimax-m3} & \cellcolor{lingD}done, no file & \multicolumn{4}{c}{\cellcolor{lingN}no file} & 0.10 / 0.10 & 67 / 100 \\
\texttt{qwen-cua} & \cellcolor{lingN}no file & \multicolumn{4}{c}{\cellcolor{lingN}no file} & 0.10 / 0.10 & 100 / 100 \\
\texttt{qwen3.7-plus} & \cellcolor{lingN}no file & \multicolumn{4}{c}{\cellcolor{lingN}no file} & 0.10 / 0.10 & 100 / 100 \\
\midrule
Within tolerance & 3/12 & 2/12 & 3/12 & 1/12 & 3/12 & & \\
\bottomrule
\end{tabular}

%% file: appendix_biology.tex
\subsection{Structural biology and NMR: application gates and coordinate frames}
\label{app:bio}

This appendix analyzes the trajectories on the eight structural-biology and spectroscopy tasks of
the biology domain; its EEG tasks are analyzed in Appendix~\ref{app:eeg}. Five tasks work on
protein structures. \texttt{align} superposes an EGFR structure (8SC7) onto another (2ITX) in PyMOL
and saves the result. \texttt{af-egfr}, \texttt{af-her2} and \texttt{af-igf1r} ask for the AlphaFold~DB
model of a receptor tyrosine kinase, its predicted aligned error (PAE) matrix, a file holding only
the kinase domain, and twelve values derived from them. \texttt{mutation} introduces L858R into
2ITX with PyMOL's mutagenesis wizard. \texttt{rcsb} is a faceted search of the RCSB PDB (EGFR,
X-ray, 2.0--2.5\,\AA, released 2005--2009) that asks for the result set, the top hit's mmCIF file and
eight values read from it. \texttt{synergy} audits two blocks of a drug-combination matrix on the
SynergyFinder web server. \texttt{nmr} asks for a full $^{1}$H assignment of 6-bromoindole from the
raw Bruker FID in Mnova (MestReNova). All tasks are graded pass or fail except \texttt{nmr}, which
is scored on a 48-point rubric.

We evaluate twelve backbones with one run per task, a limit of 100 interaction steps, and a screen
resolution of 1920\,$\times$\,1080, which gives 96 runs. Twelve are not scored. The \texttt{af-her2} and \texttt{af-igf1r} runs of Qwen3.7-plus,
MiniMax-M3 and Kimi-K3 were graded by an evaluator version that passed only the answer file to the
metric. Six attempts were lost to infrastructure faults. Qwen-CUA's model endpoint refused every
request on \texttt{af-her2} and \texttt{af-igf1r} and returned server errors from step~84 of its
\texttt{rcsb} run. The \texttt{synergy} runs of Qwen3.7-plus, MiniMax-M3 and Kimi-K3, all made on one
day, met a SynergyFinder server that dropped every session: from their third, ninth and fourth step
on, the page reports that it is disconnected from the server. Of the 84 scored runs, 57 succeed, and
the mean partial-credit score is 71.3\%. Of the 27 failed runs, 4 receive partial credit, 5 deliver an
incorrect answer, 1 declares the task complete without a valid deliverable, 1 gives up explicitly,
and 16 are VOID. We read each trajectory and assign each failed run one primary cause. The causes
are ordered along the execution path (Table~\ref{tab:bio-causes}), and Table~\ref{tab:bio-outcomes}
gives the outcome of every run. Appendix~\ref{app:bio-case} contrasts two \texttt{nmr} trajectories.

\paragraph{Successful runs.}
The 57 successes follow the tasks more closely than the backbones. Every backbone solves
\texttt{align}. Eleven do so with a fetch--align--save script run in headless PyMOL, and Kimi-K3
types the same commands into PyMOL's command line. The three extractions and \texttt{mutation} are
solved in 34 of their 40 scored runs, and Opus~5, Sonnet~5 and Fable~5.1 solve every task except
\texttt{nmr}. The two tasks that hinge on GUI dialogs separate the backbones: \texttt{synergy} is
solved three times and \texttt{nmr} never, and 18 of the 27 failures fall on these two tasks.

\paragraph{Interaction-layer failures.}
In 13 of the 27 failed runs, the actions do not take effect where the agent intends (A). In eleven of
them, the clicks follow a coordinate frame other than the 1920\,$\times$\,1080 screen. GPT-5.6-sol,
GPT-5.6-luna and GPT-5.6-terra click in a frame 1.41 times smaller, the 1365\,$\times$\,768 at which
they see the screenshot. Gemini~3.1~Pro, Qwen3.7-plus and MiniMax-M3 click in a 0--1000 normalized
frame, and Qwen-CUA in a frame 1.5 times smaller, as Appendix~\ref{sec:app-astro} documents for
these four. Scaled into those frames, the clicks hit the controls the agents describe. On
\texttt{synergy}, 100 of GPT-5.6-luna's 101 clicks land on the security dialog's checkbox or its OK
button once scaled, and none does on the screen. Across the six \texttt{nmr} runs of this kind, no
click lands on a button of Mnova's license dialogs, whereas 140 do once scaled. The mismatch decides
the outcome only where the task has no terminal route: GPT-5.6-sol and GPT-5.6-luna solve all six
structure tasks, which they work from the terminal, and fail both tasks that hinge on dialogs. In
three runs, GPT-5.6-sol and GPT-5.6-terra state the mismatch in their own reasoning, as GPT-5.6-sol
does at step~58 of \texttt{synergy}: ``PyAutoGUI uses a 1920$\times$1080 coordinate space while the
observation is 1365$\times$768.'' They correct a few clicks, which is enough to upload the workbook or to
open Mnova's License Manager, and then fall back to the image frame. The other two A runs are on
\texttt{mutation}. Gemini~3.1~Pro's PyMOL commands go into a find dialog that holds the keyboard
focus, and it declares the task complete at step~38. Qwen3.7-plus's input lands in the wrong window,
and at step~100 it is still trying to write its PyMOL script to a file.

\paragraph{Synergy: the bot check.}
SynergyFinder opens with a security dialog that carries a Google reCAPTCHA, and the dialog appears
in every run. It can be passed without the reCAPTCHA: its text asks users with limited Google access
to simply press OK. Opus~5 and Sonnet~5 press OK and are through at once, and Fable~5.1 ticks the
checkbox first. Five of the six failures are the frame mismatches above (A). GPT-5.6-luna never
dismisses the dialog. Gemini~3.1~Pro dismisses it at step~33 with a click in screen coordinates;
for the next 60 steps its clicks on the upload button land on the data-format menu, and it uploads
the workbook at step~94, six steps before the limit. Qwen-CUA dismisses the dialog from the keyboard
at step~19, and its clicks on the upload button land on empty page until the limit. GPT-5.6-sol and
GPT-5.6-terra dismiss the dialog from the keyboard, state the scaling, and upload the workbook, at
steps~59 and~29. Each then returns to the image frame, reloads the page when the controls stop
responding (steps~77 and~60), and meets the dialog again; GPT-5.6-sol does not pass it before the
limit, and GPT-5.6-terra gives up at step~66. GPT-6-Astra passes the dialog at step~17 and completes
the analysis, but it is still reading block means back from its own exported figures when the
budget ends, with no answer file written (F). The three runs made during the server outage are not
scored; Kimi-K3 traces that outage through the browser's developer tools to the server application
exiting on every new session. A task that depends on a live third-party service also depends on the
service's availability, and a locally hosted instance would remove that dependence.

\paragraph{Structure search and extraction: the value behind the number.}
Six failures fall on the four database tasks. On \texttt{rcsb}, Qwen3.7-plus filters resolution with
the search API's range operator, which excludes both limits unless told otherwise. It therefore loses
the two entries at the limits, 2RGP at 2.00\,\AA\ and 2EB2 at 2.50\,\AA, and its count, identifiers
and top hit are all wrong (C). MiniMax-M3 ends its 100 steps still rewriting search-API queries that
the service rejects as malformed (C). Kimi-K3 builds the right query and the right top hit and passes
9 of 10 checks, but reports the deposited sequence length of 2RGP as 216 instead of 315, and that
value is one of the gated checks (D). On the AlphaFold tasks, the answer files of GPT-5.6-terra and
Qwen3.7-plus for \texttt{af-egfr} report the reference domain boundaries, pLDDT means and matrix
dimensions. Their failures lie in the delivered structure files, which the archived trajectories do
not keep, so the failing check cannot be named (D). GPT-5.6-terra's \texttt{af-egfr} run submits
after three steps without opening any of the files it wrote, and its \texttt{af-igf1r} domain file
contains no residue numbered inside the kinase domain (D).

\begin{figure}[t]
\centering
\includegraphics[width=\textwidth]{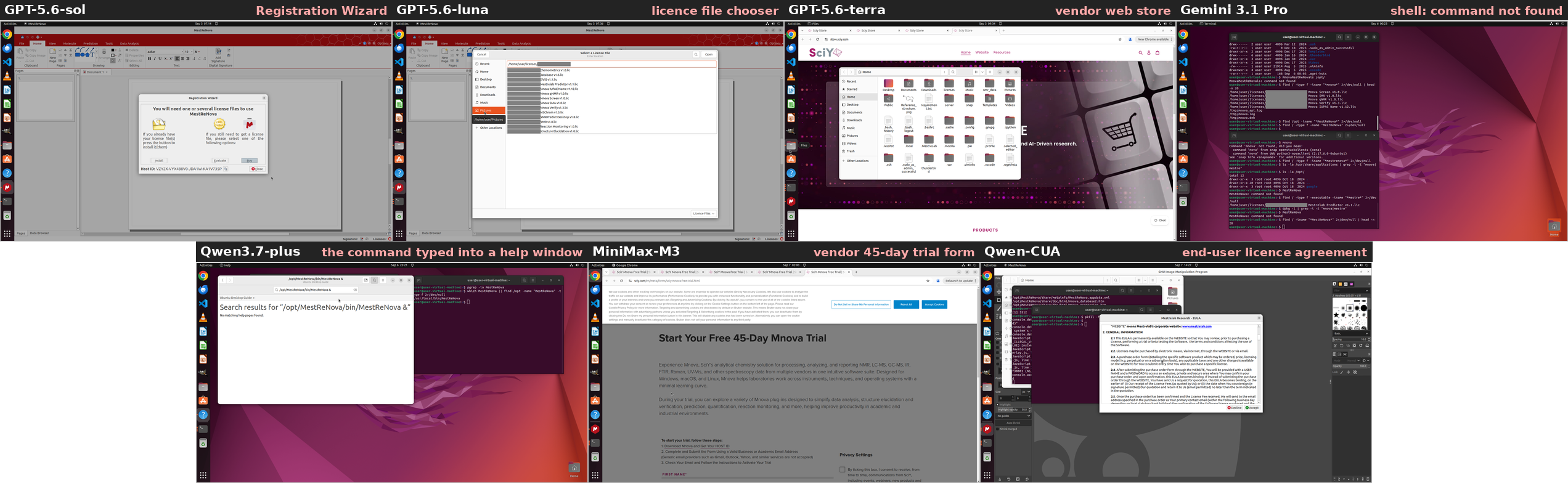}
\caption{The final screen of the seven \texttt{nmr} runs that never open a spectrum (one run per
backbone). Six click Mnova's license agreement and Registration Wizard in a scaled coordinate frame
(A), and Gemini~3.1~Pro never installs Mnova (B). They end on the Registration Wizard, the license
file chooser, the license agreement, the vendor's web store, the vendor's 45-day trial form, a shell
reporting \texttt{command not found}, or the launch command typed into a help window, with the
institutional license files in \texttt{/home/user/licenses/} throughout. Institution names are
redacted.}
\label{fig:bio-nmr-ends}
\end{figure}

\paragraph{NMR: the application gate.}
Mnova opens Bruker data only after three GUI steps: accepting its license agreement, choosing
Install in its Registration Wizard, and importing a license file. In these runs the staging step that
installs Mnova did not take effect, although the instruction states that Mnova is installed. Eleven of
the twelve runs therefore install it themselves from the staged package. Gemini~3.1~Pro never does,
and it ends with \texttt{MestReNova: command not found} (B). Five runs pass all three steps and open
the spectrum. The other six never import the license (A): their clicks on the agreement and the
wizard follow a scaled frame, as described above. GPT-5.6-luna clicks one point on the empty canvas
beside the wizard 39 times; scaled by 1.41, that point is the wizard's Install button
(Figure~\ref{fig:mnova-case-studies}, c1--c2). The seven runs that never open a spectrum end on the Registration Wizard,
the license file chooser, the license agreement, the vendor's web store, the vendor's 45-day trial
form, a shell reporting \texttt{command not found}, and the launch command typed into a help window
(Figure~\ref{fig:bio-nmr-ends}).

\paragraph{NMR: what passing the gate buys.}
The instruction asks the agent to write the answer file as soon as Mnova reports a multiplet table,
typically from its automatic multiplet analysis, and allows it to refine the file afterwards. It also
asks for overlapping aromatic signals to be handled explicitly. All five runs that open the spectrum
write their answer from the automatic analysis. The analysis groups the 7.30--7.45\,ppm region into
one two-proton multiplet where the reference assigns two protons, C2-H at 7.37\,ppm and C4-H at
7.40\,ppm. The peak list that Mnova displays above the spectrum resolves the pair, yet the pair stays
merged in every final answer (E). Four runs submit and receive partial credit: Opus~5
(0.75, 25 steps), Sonnet~5 (0.75, 56), Fable~5.1 (0.72, 22) and GPT-6-Astra (0.69, 37). They stop
with 44 to 78 steps left, and in each of them the merged pair costs 7.9 of the 48 rubric points. The
remaining differences come from one further step each. Fable~5.1 swaps the assignments of C4-H and
C7-H. GPT-6-Astra re-processes the FID (zero filling to 128K points, referencing to DMSO), checks the
acquisition parameters and states the overlap explicitly, but it gives the merged signal a confidence
of 0.9 and is penalized as confidently wrong. Kimi-K3 writes the same kind of answer at step~98 and
is checking it at step~100, when its harness ends the run with FAIL. The environment scores the run~0
without reading the file; the evaluator gives that file 0.75 when applied to it afterwards (F).

\paragraph{Scope of the analysis.}
The analysis has one run per backbone and task, so the counts describe these 84 scored trajectories
and carry no variance estimate. Each failed run receives a single primary cause. The coordinate
frames are inferred from where the clicks land relative to the controls the agents describe; they fit
most clicks, not all. The system prompt of these runs, that of the OSWorld PromptAgent for pyautogui
actions, does not state the screen resolution. Qwen-CUA's scored \texttt{synergy} run lost 4 of its
100 steps to gateway errors. Kimi-K3 runs under its vendor's agent, which uses
tool calling; we cap its screenshot history at 8 images (the vendor default is 100) because the
provider rejects larger requests. That agent ends a run with FAIL at the step limit, and the
environment then scores the run 0 without reading the delivered files. The \texttt{nmr} runs were
made with Mnova absent at the start (see above), so they are not directly comparable with runs of
the released task, in which staging installs Mnova. The \texttt{nmr} reference is a best-effort
assignment measured from the shipped 1D data and is intended for expert and 2D review. The failing
check of the two D runs on \texttt{af-egfr} cannot be recovered. The \texttt{af-igf1r} run of
GPT-5.6-terra was graded by an evaluator version that raised an exception on its empty domain; the
current evaluator fails the same range check.

\begin{table}[t]
\centering
\caption{Primary causes of the 27 failed runs on the structural-biology and NMR tasks, ordered along
the execution path. Each failed run has exactly one cause. The last column lists the affected
backbones with their number of runs.}
\label{tab:bio-causes}
\small
\setlength{\tabcolsep}{4pt}
\begin{tabular}{@{}p{0.19\linewidth} p{0.38\linewidth} c p{0.29\linewidth}@{}}
\toprule
Cause & Criterion & Runs & Backbones (runs) \\
\midrule
\multicolumn{4}{@{}l}{\textit{Action does not take effect}} \\
A\enspace Action misses & Clicks follow a coordinate frame other than the 1920\,$\times$\,1080 screen, or input goes to another window, so the step has no effect & 13 & GPT-5.6-sol, GPT-5.6-luna, GPT-5.6-terra, Gemini 3.1 Pro, Qwen3.7-plus, Qwen-CUA (2 each), MiniMax-M3 \\
\multicolumn{4}{@{}l}{\textit{Blocked before the task can start}} \\
B\enspace Application not installed & Mnova is never installed, so no spectrum is opened & 1 & Gemini 3.1 Pro \\
\multicolumn{4}{@{}l}{\textit{Method or content is wrong}} \\
C\enspace Query wrong & The database query excludes valid entries or never runs & 2 & Qwen3.7-plus, MiniMax-M3 \\
D\enspace Delivered value or file wrong & A reported value or a delivered structure file fails a check & 4 & GPT-5.6-terra (2), Qwen3.7-plus, Kimi-K3 \\
E\enspace Tool default kept & Mnova's automatic multiplet grouping is written into the answer and never revised, so the overlapping pair stays merged & 4 & Opus 5, Sonnet 5, Fable 5.1, GPT-6-Astra \\
\multicolumn{4}{@{}l}{\textit{No convergence}} \\
F\enspace Steps run out & The step limit arrives before the run submits a deliverable & 3 & Kimi-K3 (2), GPT-6-Astra \\
\bottomrule
\end{tabular}
\end{table}

\begin{table}[t]
\centering
\footnotesize
\setlength{\tabcolsep}{3.5pt}
\renewcommand{\arraystretch}{1.05}
\caption{Outcome of every run on the eight structural-biology and NMR tasks (one run per cell, step
limit 100). P: successful run. A--F: primary cause of failure as in Table~\ref{tab:bio-causes},
shaded by stage with the colors of Table~\ref{tab:geo-matrix}. A superscript V marks a VOID run,
which reaches the step limit without DONE, submits nothing, and scores 0. For the partially credited
\texttt{nmr} runs the score is given in parentheses. Pass: number of successful runs. Score: mean
partial-credit score in percent over the scored tasks.}
\label{tab:bio-outcomes}
\resizebox{\linewidth}{!}{%
\begin{tabular}{@{}lcccccccccc@{}}
\toprule
Backbone & \texttt{align} & \texttt{af-egfr} & \texttt{af-her2} & \texttt{af-igf1r} & \texttt{mutation} & \texttt{rcsb} & \texttt{synergy} & \texttt{nmr} & Pass & Score \\
\midrule
Sonnet~5 & \cellcolor{geoP}P & \cellcolor{geoP}P & \cellcolor{geoP}P & \cellcolor{geoP}P & \cellcolor{geoP}P & \cellcolor{geoP}P & \cellcolor{geoP}P & \cellcolor{geoE}E (0.75) & 7/8 & 96.9 \\
Opus~5 & \cellcolor{geoP}P & \cellcolor{geoP}P & \cellcolor{geoP}P & \cellcolor{geoP}P & \cellcolor{geoP}P & \cellcolor{geoP}P & \cellcolor{geoP}P & \cellcolor{geoE}E (0.75) & 7/8 & 96.9 \\
Fable~5.1 & \cellcolor{geoP}P & \cellcolor{geoP}P & \cellcolor{geoP}P & \cellcolor{geoP}P & \cellcolor{geoP}P & \cellcolor{geoP}P & \cellcolor{geoP}P & \cellcolor{geoE}E (0.72) & 7/8 & 96.5 \\
GPT-6-Astra & \cellcolor{geoP}P & \cellcolor{geoP}P & \cellcolor{geoP}P & \cellcolor{geoP}P & \cellcolor{geoP}P & \cellcolor{geoP}P & \cellcolor{geoH}F$^{V}$ & \cellcolor{geoE}E (0.69) & 6/8 & 83.6 \\
GPT-5.6-luna & \cellcolor{geoP}P & \cellcolor{geoP}P & \cellcolor{geoP}P & \cellcolor{geoP}P & \cellcolor{geoP}P & \cellcolor{geoP}P & \cellcolor{geoA}A$^{V}$ & \cellcolor{geoA}A$^{V}$ & 6/8 & 75.0 \\
GPT-5.6-sol & \cellcolor{geoP}P & \cellcolor{geoP}P & \cellcolor{geoP}P & \cellcolor{geoP}P & \cellcolor{geoP}P & \cellcolor{geoP}P & \cellcolor{geoA}A$^{V}$ & \cellcolor{geoA}A$^{V}$ & 6/8 & 75.0 \\
Gemini~3.1~Pro & \cellcolor{geoP}P & \cellcolor{geoP}P & \cellcolor{geoP}P & \cellcolor{geoP}P & \cellcolor{geoA}A & \cellcolor{geoP}P & \cellcolor{geoA}A$^{V}$ & \cellcolor{geoD}B$^{V}$ & 5/8 & 62.5 \\
MiniMax-M3 & \cellcolor{geoP}P & \cellcolor{geoP}P & -- & -- & \cellcolor{geoP}P & \cellcolor{geoE}C$^{V}$ & -- & \cellcolor{geoA}A$^{V}$ & 3/5 & 60.0 \\
Qwen-CUA & \cellcolor{geoP}P & \cellcolor{geoP}P & -- & -- & \cellcolor{geoP}P & -- & \cellcolor{geoA}A$^{V}$ & \cellcolor{geoA}A$^{V}$ & 3/5 & 60.0 \\
GPT-5.6-terra & \cellcolor{geoP}P & \cellcolor{geoE}D & \cellcolor{geoP}P & \cellcolor{geoE}D & \cellcolor{geoP}P & \cellcolor{geoP}P & \cellcolor{geoA}A & \cellcolor{geoA}A$^{V}$ & 4/8 & 50.0 \\
Kimi-K3 & \cellcolor{geoP}P & \cellcolor{geoP}P & -- & -- & \cellcolor{geoH}F$^{V}$ & \cellcolor{geoE}D & -- & \cellcolor{geoH}F$^{V}$ & 2/5 & 40.0 \\
Qwen3.7-plus & \cellcolor{geoP}P & \cellcolor{geoE}D & -- & -- & \cellcolor{geoA}A$^{V}$ & \cellcolor{geoE}C & -- & \cellcolor{geoA}A$^{V}$ & 1/5 & 20.0 \\
\midrule
Pass & 12/12 & 10/12 & 8/8 & 7/8 & 9/12 & 8/11 & 3/9 & 0/12 & 57/84 & 71.3 \\
\bottomrule
\end{tabular}}
\end{table}

%% file: casestudy_qupath.tex
\subsection{Medicine Analysis}
\begin{figure*}[ht]
    \centering
    \includegraphics[width=0.92\textwidth]{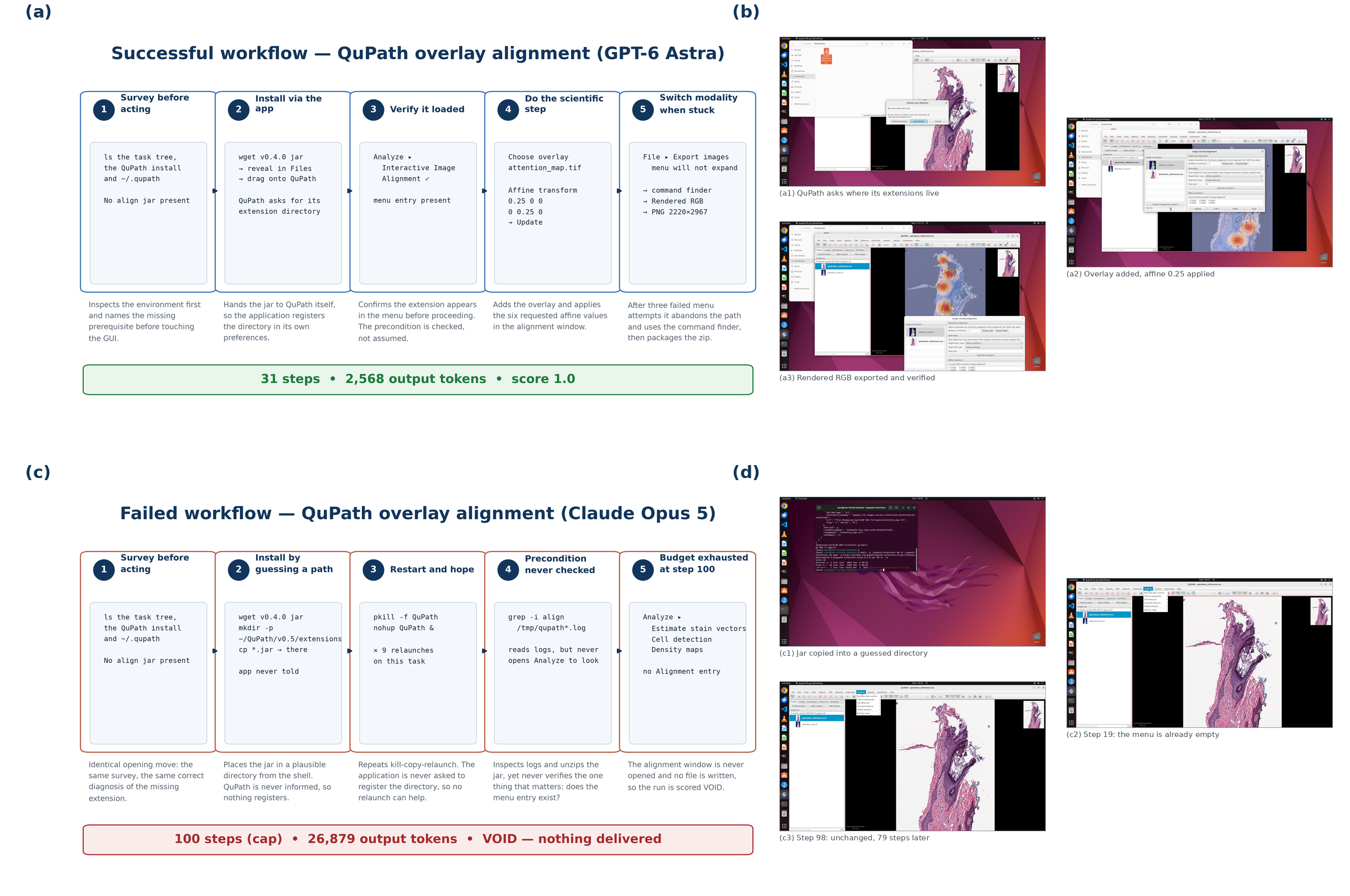}
    \caption{Contrasting workflows on \texttt{QP\_004\_T1} (register an attention map onto a reference slide; requires installing the Interactive Image Alignment extension). \textbf{Top:} GPT-6~Astra installs through the application, confirms the menu entry exists, applies the affine transform, and switches to the command finder when menus stall --- 31 steps, score~1.0. \textbf{Bottom:} Claude Opus~5 copies the jar into a guessed directory from the shell and relaunches QuPath nine times. Panels (c2) and (c3) show the same empty \texttt{Analyze} menu at step~19 and step~98: the disconfirming evidence was available 79 steps before the budget ran out. The run is scored \texttt{VOID}. The decisive difference is verifying the precondition through the interface that owns it, not trajectory length or effort.}
    \label{fig:qupath-case-studies}
\end{figure*}

\textbf{Selective case study.} QuPath's overlay-alignment tasks (\texttt{QP\_004\_T1}, \texttt{QP\_004\_T2}) ask the agent to register a model attention map onto a reference slide. Unlike every other task in the suite, the required command does not exist when the run starts: the Interactive Image Alignment extension must first be fetched from the project's release page and installed into QuPath. The task therefore separates the scientific step from a prerequisite that the application itself controls, and it is where the backbones diverge most sharply --- ten of the twelve models deliver nothing at all on \texttt{QP\_004\_T1}, while GPT-6~Astra and Fable~5.1 both score~1.0 (Figure~\ref{fig:qupath-case-studies}).

\emph{Successful trajectory.} GPT-6~Astra surveyed the environment first, named the missing jar, and downloaded release v0.4.0. It then handed the file to QuPath rather than to the filesystem: it revealed the download in the file manager and dropped it onto the running window, at which point QuPath asked where its extension directory should be and registered it. Before doing any science, the run opened \texttt{Analyze} and confirmed that \emph{Interactive Image Alignment} had appeared; only then did it add \texttt{attention\_map.tif} as the overlay and enter the six requested affine values. When the \texttt{File\,$\to$\,Export images} submenu twice refused to expand, it abandoned menu navigation after the third attempt and reached the same command through QuPath's command finder, exported the rendered RGB image at the required $2220\times2967$, and packaged the submission zip. The evaluator scored~1.0 after 31 steps and 2{,}568 output tokens --- the cheapest solved run on the task.

\emph{Failed trajectory.} Claude Opus~5 opened identically: the same survey, the same correct diagnosis of the missing extension. It then installed the jar by \emph{guessing where it belonged}, creating \texttt{\textasciitilde/QuPath/v0.5/extensions} and copying the file there from the shell. QuPath was never told, so nothing was registered, and no amount of restarting could change that; the run relaunched the application nine times on this task alone. The decisive detail is that the evidence was already visible: at step~19 the run opened \texttt{Analyze} and the menu contained only the six stock entries, exactly as it still did at step~98. Instead of treating the empty menu as a refutation of its install, the run went back to reading \texttt{/tmp/qupath*.log} and re-unzipping the jar to inspect its manifest, and repeated the same install-and-relaunch cycle for another 79 steps. It never opened the alignment window, never wrote a file, and was scored \texttt{VOID} at the 100-step cap after 26{,}879 output tokens --- ten times Astra's spend for zero delivery.

\emph{Takeaway.} The failure is not a reasoning failure about pathology, and not a matter of trajectory length: both runs understood the task and Opus~5 had six times the budget. It is a failure to treat an application's own state as authoritative. Registering an extension is a transaction that only QuPath can perform; writing the jar into a plausible directory produces a filesystem that looks correct and an application that is unchanged. Fable~5.1 is the informative middle case: it spent roughly thirty steps on the same shell-copy strategy as Opus~5 and relaunched thirteen times, but then typed the path into QuPath's own extension-directory prompt, which registered it, and finished at step~84. What separates the three runs is not planning quality or persistence but whether the agent ever asked the application to confirm the precondition --- and, having been shown an empty menu, whether it let that observation overturn its belief. For GUI-driven scientific software, agents should verify a prerequisite through the interface that owns it before consuming budget on the work that depends on it, and should treat a negative check as disconfirming evidence rather than as a retry.

%% file: casestudy_mnova.tex
\section{Biology Analysis}
\label{app:bio-case}
\begin{figure*}[h]
    \centering
    \includegraphics[width=0.92\textwidth]{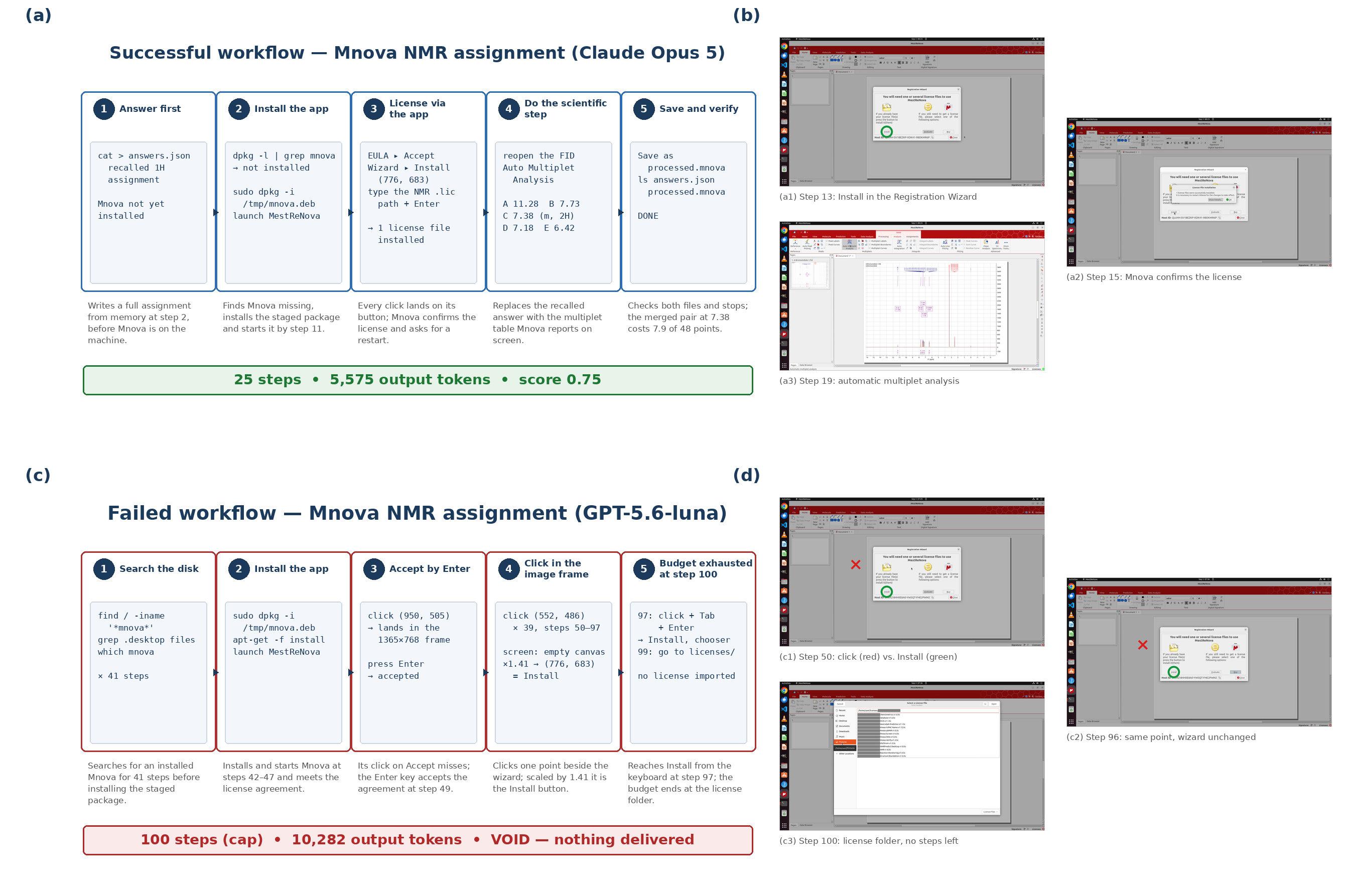}
    \caption{Contrasting workflows on \texttt{nmr\_6bromoindole\_assignment} (assign the $^{1}$H spectrum of 6-bromoindole from the raw FID in Mnova; requires accepting Mnova's license agreement, choosing Install in its Registration Wizard and importing a license file). \textbf{Top:} Claude Opus~5 installs Mnova, clicks Accept and Install where they are, and Mnova itself confirms the license (a2) before the run analyzes the spectrum (a3) --- 25 steps, score~0.75. \textbf{Bottom:} GPT-5.6-luna clicks one point beside the Registration Wizard 39 times; scaled by 1.41, the point is the Install button (red cross: the click; green circle: Install). Panels (c1) and (c2) show the same unchanged wizard at step~50 and step~96; the run turned to the keyboard only at step~97, three steps before the budget ran out. The run is scored \texttt{VOID}. The decisive difference is whether the clicks land in the screen's coordinate frame, not knowledge of NMR. Institution names in the license file names are redacted.}
    \label{fig:mnova-case-studies}
\end{figure*}

\textbf{Selective case study.} Mnova's NMR task (\texttt{nmr\_6bromoindole\_assignment}) asks the agent to assign the $^{1}$H spectrum of 6-bromoindole from the raw Bruker FID. Unlike the other structural-biology tasks, the analysis cannot begin until the application is licensed: Mnova opens Bruker data only after the agent accepts its license agreement, chooses Install in its Registration Wizard, and imports a license file, three steps that only Mnova's own dialogs offer. The task therefore separates the scientific step from a prerequisite that the application itself controls, and it is where the backbones diverge most sharply --- seven of the twelve models never open the spectrum, while the five that pass the gate all deliver an assignment (Figure~\ref{fig:mnova-case-studies}; Appendix~\ref{app:bio}).

\emph{Successful trajectory.} Claude Opus~5 wrote a complete assignment from memory at step~2, before Mnova was even installed, and only then turned to the application. It found Mnova missing, installed the staged package, and started it at step~11. It accepted the license agreement at step~12, chose Install in the Registration Wizard at step~13, and typed the path of the NMR license file into the file dialog at step~14; each click landed on its button, and Mnova itself confirmed the installed license and asked for a restart. The run restarted Mnova directly on the FID, ran the automatic multiplet analysis at step~18, and at step~19, as the instruction asks, replaced its recalled answer with the five multiplets on screen. It saved the processed document, confirmed that both output files existed, and stopped. The evaluator scored~0.75 after 25 steps and 5{,}575 output tokens. Reading the spectrum lifted the score above the 0.70 that the recalled answer would have earned; most of the points still missing are on the overlapping C2-H/C4-H pair, which Mnova's analysis reports as one two-proton signal and the run kept.

\emph{Failed trajectory.} GPT-5.6-luna spent its first 41 steps searching the disk for an installed Mnova, then installed the staged package and started it at steps~42--47. Its click on Accept in the license agreement landed short of the button, and the Enter key accepted the agreement at step~49. It then aimed at the wizard's Install button and missed it. From step~50 to step~97, 39 of its 48 steps click within two pixels of (552,~486), a point on the empty canvas beside the wizard; scaled by 1.41, the ratio between the 1920$\times$1080 screen and the 1365$\times$768 image in which the model sees it, that point is (776,~683), exactly where Opus~5 clicked Install. The decisive detail is that the evidence was on screen at every step: the wizard looks the same at step~50 and at step~96, and seven of the other nine steps only switch or close windows before the run clicks the same point again. Only at step~97 did it add Tab and Enter to the click, which activated Install from the keyboard. The license chooser opened, the run reached the license folder at step~99, and it was scored \texttt{VOID} at the 100-step cap after 10{,}282 output tokens --- nearly twice Opus~5's output for no delivery.

\emph{Takeaway.} The failure is not a failure of spectroscopy, and not a matter of budget: GPT-5.6-luna solves all six structure tasks of the domain, and it had four times the steps that Opus~5 needed. It is a failure at the interaction layer, at a step that only the application offers: the license becomes active only through Mnova's dialogs, and the dialogs respond only to clicks that land on them. GPT-5.6-sol is the informative middle case. On the same task it states the mismatch in its own reasoning at step~65 (``the desktop is scaled from 1920$\times$1080 to the 1366$\times$768 observation''), corrects its clicks, and opens Mnova's License Manager, but it later falls back to the image frame and ends on the Registration Wizard at step~100. What separates the three runs is not knowledge of NMR or persistence but whether the clicks land in the screen's coordinate frame --- and, once an unchanged dialog shows that they do not, whether the agent lets that observation change its coordinates. For GUI-driven scientific software, harnesses should check each backbone's clicks against the screen frame, or rescale them, and tasks that measure analysis should install and license the application during staging, as the released version of this task does. Passing the gate does not settle the analysis either: every run that opened the spectrum kept Mnova's grouping of the overlapping pair, which the rubric's overlap criterion is there to detect.

%% file: appendix_ciao.tex
\subsection{Physics Analysis I: CIAO in Astronomy}
\label{sec:app-astro}
\suppressfloats[t]

\definecolor{astA}{rgb}{0.98,0.85,0.82}
\definecolor{astB}{rgb}{0.99,0.92,0.78}
\definecolor{astC}{rgb}{0.89,0.86,0.96}
\definecolor{astD}{rgb}{0.84,0.90,0.98}
\definecolor{astG}{rgb}{0.88,0.88,0.88}
\definecolor{astP}{rgb}{0.84,0.94,0.84}

This appendix analyzes the trajectories on the three astronomy tasks, which run in SAOImageDS9 with the CIAO analysis menus on Chandra ACIS data. The first task, \texttt{netcounts} (ObsID 13858), asks for the background-subtracted counts of a quasar in a 5$''$ source circle, with a 20--60$''$ annulus as the background region. The reference is 1424.92 counts. An answer scores 1 within 0.5 counts and 0.25 within 1.5 counts. Both tolerances are smaller than the 2.08-count gap between the net counts and the raw counts (1427), so the raw value scores 0. The second task, \texttt{lightcurve} (ObsID 4479), asks for four quantities of a background-subtracted light curve: the time of the peak 5000-s bin, the time of the first 5000-s bin above 0.015 counts\,s$^{-1}$, and the peak rates with 5000-s and 1000-s bins to four significant figures (tolerance $10^{-5}$ counts\,s$^{-1}$). Each quantity is worth 0.25. The third task, \texttt{streak} (ObsID 88), asks for the physical coordinates of a point on the ACIS readout streak of a bright source, at least 3~arcmin from that source. The streak is tilted $6.1^{\circ}$ from vertical. A point in the scored part of the streak scores 1 within 3 pixels of the streak line and 0.25 within 8 pixels.

We evaluate twelve backbones with one run per task and a limit of 100 interaction steps (150 for eight backbones on \texttt{lightcurve}, where no run uses more than 100), which gives 36 runs. Fourteen runs succeed, and the mean partial-credit score is 45.1\%. Of the 22 failed runs, 13 deliver an incorrect answer, 8 are \texttt{VOID}, and 1 declares the task complete with an empty answer. We read each trajectory and assign each failed run one primary cause. The causes are ordered along the execution path (Table~\ref{tab:astro-taxonomy}), and Table~\ref{tab:astro-matrix} gives the outcome of every run.

\paragraph{Successful runs.} The 14 successful runs are concentrated in a few backbones. Opus~5 and Fable~5.1 pass all three tasks, GPT-5.6-sol, GPT-6-Astra, and Kimi-K3 pass two, and no other backbone passes more than one. Compared with failed runs on the same tasks, the successful runs share one practice in how they obtain the final number. Each takes it from the full-precision output of a CIAO tool, computes it from the data, or checks it directly against the image. The data are either the event list or the raw counts and areas that the tool reports. The length of a trajectory says little about its outcome. Successful runs take 6 to 77 steps, and failed runs that deliver an answer take 5 to 100.

\paragraph{Interaction-layer failures.} In 8 of the 22 failed runs, the actions do not take effect where the agent intends (A). The system prompt states the $1920\times1080$ pixel frame. Gemini~3.1~Pro, Qwen3.7-plus, and MiniMax-M3 give coordinates in a 0--1000 normalized frame, and the clicks of Qwen-CUA mostly fit a frame 1.5 times smaller (about $1280\times720$). Their clicks land away from the intended terminal, menu item, or dialog button. In five runs, clicks meant to focus the terminal land on DS9 or on the desktop. For most of the run, the commands typed next go there instead of to the shell. These are Qwen3.7-plus on \texttt{netcounts} and \texttt{lightcurve}, Qwen-CUA on \texttt{netcounts} and \texttt{streak}, and Gemini~3.1~Pro on \texttt{streak}. Gemini~3.1~Pro on \texttt{netcounts} never types a command, and its clicks miss DS9's menu items throughout its 100 steps. On \texttt{streak}, Qwen3.7-plus fails to hit a button of a DS9 dialog for 83 steps. MiniMax-M3 opens other applications by mistake and loses the DS9 window. Seven of these runs are \texttt{VOID}. Gemini~3.1~Pro on \texttt{streak} declares the task complete with an empty answer. Gemini~3.1~Pro, Qwen-CUA, and MiniMax-M3 also lose steps in this way on \texttt{lightcurve}, but there it does not decide their outcome. Gemini~3.1~Pro recovers after 15 ineffective steps and passes, and Qwen-CUA and MiniMax-M3 fail because of their method (D, below).

\paragraph{Net counts: setting up the measurement.} On \texttt{netcounts}, nine runs have actions that take effect, and eight of them define both regions with the given center and radii. MiniMax-M3 omits the arcsecond mark (B), so CIAO reads the radii as 5, 20, and 60 physical pixels. \texttt{dmstat} reports 80 pixels in its source aperture, about a quarter of the 326 pixels of a 5$''$ circle. The run still describes the aperture as a 5$''$ circle and submits 1394.34.

\paragraph{Net counts: reading the result.} For the other eight runs, the difficulty is reading the value. DS9's Net Counts output does not show the correct value on the screen in any run. When the annulus is not tagged as background, the table lists the raw counts of each region (1427 and 265) under the header \texttt{NET\_COUNTS}. When it is tagged, the output window opens scrolled to the bottom. There, \texttt{COUNTS}, \texttt{BG\_ERR}, and the path of the output file are visible, and the \texttt{NET\_COUNTS} row is above the visible text. No run scrolls a result window up. The five successful runs obtain the value in other ways. GPT-5.6-sol treats the untagged table as raw counts and scales the annulus by the area ratio 1/128, and the 1424.93 of Sonnet~5 matches the same scaling. Opus~5 computes the subtraction in the terminal from \texttt{dmstat} pixel areas. Fable~5.1 and GPT-6-Astra read \texttt{NET\_COUNTS} from the output file (1424.9204178903). Every successful answer is within 0.0096 counts of the reference. The two failures at this stage (C) enter a number from the screen, or the difference of two such numbers. GPT-5.6-luna received the same task instruction as GPT-5.6-sol and saw the same untagged table. It enters 1427 as the background-subtracted value. GPT-5.6-terra enters $1427-\texttt{BG\_ERR}=1410.7212$ from the tagged window (see the case study below). Kimi-K3 produces the correct measurement in DS9 at step~90 but never writes it to the answer file (G). It spends steps 7--22 and 93--100 reading the source code of DS9's analysis tools.

\paragraph{Light curve: replacing the tool's normalization.} On \texttt{lightcurve}, eleven runs deliver answers. The two times are almost always right: the first-crossing time in 11 of 11 runs and the peak time in 10 of 11. The two rates are right in 6 of 11, and one practice separates these six runs from the others: they keep the exposure and background normalization of the CIAO tools. Four of them run \texttt{dmextract} with \texttt{opt=ltc1} and keep its default normalization (GPT-5.6-sol, Opus~5, Fable~5.1, and Gemini~3.1~Pro). GPT-6-Astra drives DS9's analysis backend from the terminal. Kimi-K3 uses DS9's light-curve tool and is the only run that measures through the interface. Every run that delivers wrong rates replaces or overrides this normalization (D). GPT-5.6-luna and GPT-5.6-terra compute the rates in Python and divide by the nominal bin width instead of the dead-time-corrected exposure. Both submit 0.05406 and 0.06775. These values are low by exactly the dead-time factor of 0.98734, and their errors are 69 and 87 times the tolerance. Both runs see a \texttt{dmextract} warning that suggests \texttt{opt=ltc1}. GPT-5.6-luna even switches to that option and prints the \texttt{EXPOSURE} column before it divides by the nominal width. Sonnet~5 passes the area ratio as \texttt{bkgnorm} on top of the area scaling that \texttt{dmextract} already applies. The background is therefore scaled twice, and in effect it is not subtracted (0.05509 and 0.06887). MiniMax-M3 and Qwen-CUA anchor the bins on the first event instead of the start of the observation, and they also divide by the nominal width. For MiniMax-M3, the shift moves the peak into the next bin, so its peak time is also wrong.

\paragraph{Streak: assuming the geometry.} On \texttt{streak}, eight runs deliver a point, and three of them pass, each within 0.41 pixels of the streak line. Four of the five failed deliveries place the point almost straight below the source (E), 70.8--111.1 pixels from the streak. Their X coordinates are within 23 pixels of the X coordinate of the source. GPT-5.6-luna and Sonnet~5 describe the streak as vertical. GPT-5.6-sol and GPT-5.6-terra give no reasoning but place their points in the same way. At the rows of these points, the streak line lies 63--109 pixels to the left of the source. A vertical line through the source is at least 38.8 pixels from the streak everywhere in the scored range, so this placement can never score. The point of GPT-5.6-luna also lies below the lower end of the streak. Sonnet~5 reports a source point and a streak point with the same X coordinate. It states that the cursor is on the streak, although the magnifier shows no line under it. GPT-6-Astra notes the tilt. Like Fable~5.1, it works at the default zoom, where one screen pixel covers 4.27 physical pixels. Unlike Fable~5.1, it does not check the magnifier, and its point is 8.60 pixels from the line, just outside the 8-pixel tolerance (F). Fable~5.1 notes that the streak leans down and to the left. It checks that the line is centered in the magnifier before it reports a point 7~arcmin from the source. Opus~5 and Kimi-K3 compute the line from the events. Opus~5 first selects the events in the CCD column of the source. It then switches to an angle scan in sky coordinates, which finds the line $5.85^{\circ}$ from vertical and gives its submitted point. Kimi-K3 fits a line to the events in that column ($6.09^{\circ}$). It replaces an earlier candidate, 16.5 pixels off the line, with the event closest to the fit.

\begin{figure}[t]
    \centering
    \includegraphics[width=\textwidth]{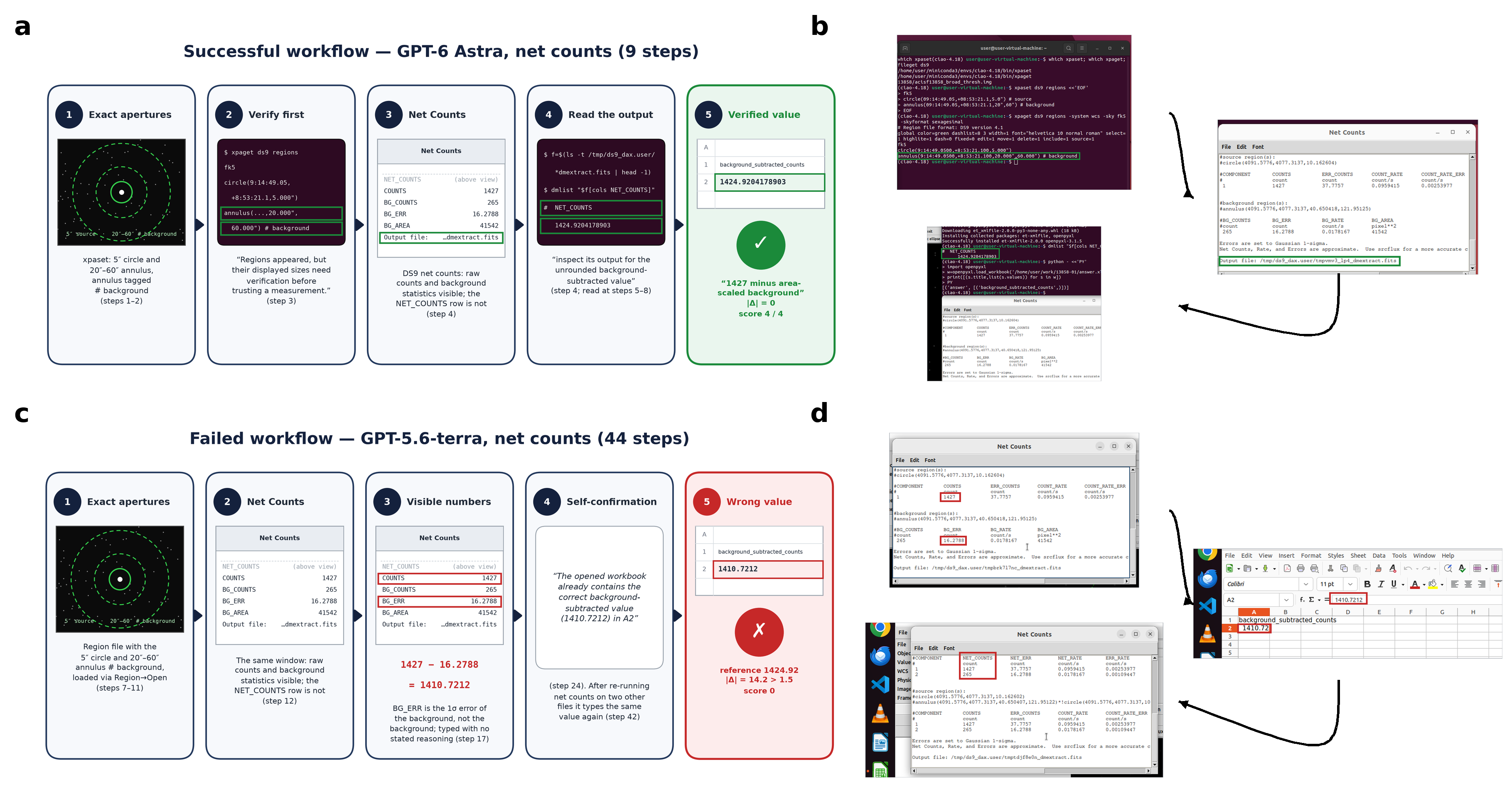}
    \caption{Contrasting GPT-6-Astra and GPT-5.6-terra workflows on \texttt{netcounts}, with representative raw trajectory frames shown beside each workflow diagram. (a) and (b): the successful run reads its regions back from DS9. It then reads \texttt{NET\_COUNTS} from the output file named at the bottom of the Net Counts window (green boxes) and writes 1424.9204178903. (c) and (d): the failed run sees the same window with the \texttt{NET\_COUNTS} row scrolled out of view. It enters $1427-\texttt{BG\_ERR}=1410.7212$ (red boxes) and later cites its own entry as the correct value. The decisive difference is the origin of the reported number, not the setup, the route, or the trajectory length.}
    \label{fig:astro-case}
\end{figure}

\paragraph{Selective case study.} We contrast a GPT-6-Astra and a GPT-5.6-terra trajectory on \texttt{netcounts} (Figure~\ref{fig:astro-case}). Both runs defined the same two regions, tagged the annulus as background, and saw the same Net Counts window, but they had opposite outcomes. The comparison illustrates that success depends less on the interface route or the number of steps than on where the agent takes the final number from: the tool's full-precision output or numbers visible on the screen.

\emph{Successful trajectory.} GPT-6-Astra confirmed from a terminal that the XPA tools reached the running DS9. It sent both regions in FK5 coordinates, with explicit arcsecond marks and the annulus tagged as \texttt{background}. Before it used any result, it noted that the regions' ``\emph{displayed sizes need verification before trusting a measurement}''. It read them back with \texttt{xpaget}, which returned the requested radii. After it ran DS9's Net Counts, it did not transcribe a number from the result window but planned to ``\emph{inspect its output for the unrounded background-subtracted value}''. The window named the FITS table that the analysis had written. The run read that table with CIAO's \texttt{dmlist} and noted ``\emph{NET\_COUNTS = 1424.9204178903 (1427 minus area-scaled background)}''. It wrote the value to A2 with \texttt{openpyxl}, reloaded the workbook, and asserted that the stored value matched. The evaluator found the answer correct (score~1.0 after 9 steps). The productive pattern was short but layered. Each check resolved a named assumption before the next step relied on it: first the regions, then the origin of the value, then the saved cell.

\emph{Failed trajectory.} GPT-5.6-terra wrote the same tagged regions to a file from the terminal, loaded the file through DS9's Region menu, and ran Net Counts. The window it saw at step~12 matched the one GPT-6-Astra saw. \texttt{COUNTS} 1427 and \texttt{BG\_ERR} 16.2788 were in view, and the \texttt{NET\_COUNTS} row was scrolled above them. At step~17, before it had written any reasoning text, it typed 1410.7212 into A2 of the answer workbook and saved it. This value equals $1427-16.2788$ to the last digit. It is the raw counts minus the $1\sigma$ uncertainty of the background, not minus the area-scaled background of 2.08 counts. The run then ran Net Counts again. Its attempt to enter a different value, 1407.7788, did not reach the workbook. At step~24 it validated the saved entry against itself: ``\emph{The opened workbook already contains the correct background-subtracted value (1410.7212) in A2}''. Later it loaded two other data files and ran Net Counts on the event file. The new table listed 1427 under \texttt{NET\_COUNTS}, and at step~42 the run typed 1410.7212 again. The evaluator found the answer 14.2 counts from the reference, against a tolerance of 1.5 (score~0.0 after 44 steps). Once 1410.7212 was in the workbook, the run checked the cell against itself rather than against the definition of the quantity. The later measurements did not change the saved value.

\emph{Takeaway.} A number read from a GUI result window is therefore not sufficient evidence until its definition has been checked. The window can show the right label over the wrong content, as the untagged table does, or the right content out of view, as the scrolled window does. As in the chemistry case study (Figure~\ref{fig:opus5-case-studies}), a check that shares the value's own origin establishes consistency but not correctness. Here that check is a saved cell compared with itself. The same pattern appears on the other two tasks, where the failed runs replace the tool's normalization (D) or assume a vertical streak (E). For quantitative GUI measurements, agents should trace each reported number to the tool's full-precision output, or recompute it from inputs whose meaning they have verified, before they commit it.

\begin{table}[t]
    \centering
    \footnotesize
    \setlength{\tabcolsep}{4pt}
    \renewcommand{\arraystretch}{1.05}
    \caption{Primary causes of the 22 failed runs on the astronomy tasks, ordered along the execution path. Each failed run has exactly one cause. The last column lists the affected backbones with their number of runs.}
    \label{tab:astro-taxonomy}
    \begin{tabular}{@{}l@{\hspace{5pt}}p{0.17\textwidth}p{0.37\textwidth}cp{0.25\textwidth}@{}}
        \toprule
        & Cause & Criterion & Runs & Backbones (runs) \\
        \midrule
        \multicolumn{5}{@{}l}{\textit{Action does not take effect}} \\
        A & Coordinate-frame mismatch & Click coordinates do not follow the $1920\times1080$ frame stated in the system prompt, so clicks land away from the intended terminal, menu item, or dialog button & 8 & Qwen3.7-plus (3), Gemini~3.1~Pro, Qwen-CUA (2 each), MiniMax-M3 (1) \\
        \multicolumn{5}{@{}l}{\textit{Measurement is set up wrong}} \\
        B & Wrong aperture unit & Region radii given without the arcsecond mark are read as pixels & 1 & MiniMax-M3 (1) \\
        \multicolumn{5}{@{}l}{\textit{Result is misread}} \\
        C & Visible number taken as the result & A number shown on the screen, or the difference of two such numbers, is entered as the background-subtracted value & 2 & GPT-5.6-luna, GPT-5.6-terra (1 each) \\
        \multicolumn{5}{@{}l}{\textit{Method is inadequate}} \\
        D & Tool normalization replaced & Rates computed by hand, or with an overriding parameter, instead of with the tool's exposure and background normalization & 5 & GPT-5.6-luna, GPT-5.6-terra, Sonnet~5, MiniMax-M3, Qwen-CUA (1 each) \\
        E & Geometric prior in place of observation & The point is placed almost straight below the source, as if the streak were vertical & 4 & GPT-5.6-sol, GPT-5.6-luna, GPT-5.6-terra, Sonnet~5 (1 each) \\
        F & Unverified placement & The point is placed at the default zoom without a magnifier check and lands just outside the tolerance & 1 & GPT-6-Astra (1) \\
        \multicolumn{5}{@{}l}{\textit{No convergence}} \\
        G & Step limit with no output & The measurement is produced, but the steps run out before the answer is written & 1 & Kimi-K3 (1) \\
        \bottomrule
    \end{tabular}
\end{table}

\begin{table}[t]
    \centering
    \footnotesize
    \setlength{\tabcolsep}{5pt}
    \renewcommand{\arraystretch}{1.05}
    \caption{Outcome of every run on the three astronomy tasks (one run per cell). P: successful run. A--G: primary cause of failure as in Table~\ref{tab:astro-taxonomy}, shaded by stage. A superscript V marks a \texttt{VOID} run, which reaches the step limit without \texttt{DONE}, submits nothing, and scores 0. For each failed \texttt{lightcurve} run that delivers an answer, the partial-credit score is given in parentheses. Pass: number of successful runs. Score: mean partial-credit score in percent.}
    \label{tab:astro-matrix}
    \begin{tabular}{@{}lccccc@{}}
        \toprule
        Backbone & \texttt{netcounts} & \texttt{lightcurve} & \texttt{streak} & Pass & Score \\
        \midrule
        Opus~5 & \cellcolor{astP}P & \cellcolor{astP}P & \cellcolor{astP}P & 3/3 & 100.0 \\
        Fable~5.1 & \cellcolor{astP}P & \cellcolor{astP}P & \cellcolor{astP}P & 3/3 & 100.0 \\
        GPT-5.6-sol & \cellcolor{astP}P & \cellcolor{astP}P & \cellcolor{astD}E & 2/3 & 66.7 \\
        GPT-6-Astra & \cellcolor{astP}P & \cellcolor{astP}P & \cellcolor{astD}F & 2/3 & 66.7 \\
        Kimi-K3 & \cellcolor{astG}G$^{\mathrm{V}}$ & \cellcolor{astP}P & \cellcolor{astP}P & 2/3 & 66.7 \\
        Sonnet~5 & \cellcolor{astP}P & \cellcolor{astD}D (0.5) & \cellcolor{astD}E & 1/3 & 50.0 \\
        Gemini~3.1~Pro & \cellcolor{astA}A$^{\mathrm{V}}$ & \cellcolor{astP}P & \cellcolor{astA}A & 1/3 & 33.3 \\
        GPT-5.6-luna & \cellcolor{astC}C & \cellcolor{astD}D (0.5) & \cellcolor{astD}E & 0/3 & 16.7 \\
        GPT-5.6-terra & \cellcolor{astC}C & \cellcolor{astD}D (0.5) & \cellcolor{astD}E & 0/3 & 16.7 \\
        Qwen-CUA & \cellcolor{astA}A$^{\mathrm{V}}$ & \cellcolor{astD}D (0.5) & \cellcolor{astA}A$^{\mathrm{V}}$ & 0/3 & 16.7 \\
        MiniMax-M3 & \cellcolor{astB}B & \cellcolor{astD}D (0.25) & \cellcolor{astA}A$^{\mathrm{V}}$ & 0/3 & 8.3 \\
        Qwen3.7-plus & \cellcolor{astA}A$^{\mathrm{V}}$ & \cellcolor{astA}A$^{\mathrm{V}}$ & \cellcolor{astA}A$^{\mathrm{V}}$ & 0/3 & 0.0 \\
        \midrule
        Pass & 5/12 & 6/12 & 3/12 & 14/36 & 45.1 \\
        \bottomrule
    \end{tabular}
\end{table}

\paragraph{Scope of the analysis.} The analysis has one run per backbone and task, so the counts describe these 36 trajectories and carry no variance estimate. Each failed run receives a single primary cause.
\clearpage

%% file: an_section_ansys.tex
\subsection{Physics Analysis II: Ansys Fluent in FLUID DYNAMICS}
\label{sec:app-ansys}
\suppressfloats[t]

\definecolor{ansA}{rgb}{0.98,0.85,0.82}
\definecolor{ansD}{rgb}{0.99,0.92,0.78}
\definecolor{ansE}{rgb}{0.84,0.90,0.98}
\definecolor{ansF}{rgb}{0.89,0.86,0.96}
\definecolor{ansH}{rgb}{0.88,0.88,0.88}
\definecolor{ansP}{rgb}{0.84,0.94,0.84}

This appendix analyzes the trajectories on the 14 ANSYS tasks, which run in Ansys Fluent 2026~R1~\citep{ansys_software} on a Windows~10 desktop. Four tasks derive from the Fluent tutorials: continuing a volume-of-fluid inkjet simulation (FT-005c), reading out a cavitating nozzle flow (FT-006b), and two stages of an adjoint shape optimization of a cylinder in cross-flow (FT-008a, FT-008c). The other ten form five pairs of verification cases: a laminar channel (FV-001), a turbulent channel with the $k$--$\varepsilon$ model (FV-002), a heated channel (FV-003), a shear-thinning power-law liquid (FV-004), and steady conduction in a square slab (FV-005). In each pair, the \emph{a} task sets up and solves the case from a supplied mesh, and the \emph{b} task extracts quantities from a supplied converged solution. Every instruction pins the solver settings and asks for two to six named quantities in a JSON report. A quantity earns its weight if it lies within 0.1\% of the reference, and the adjoint iteration count of FT-008a must match exactly. We evaluate twelve backbones with one run per task, a limit of 100 interaction steps, and a screen resolution of $1920\times1080$, which gives 168 runs. Sixty-six runs succeed, and the mean partial-credit score is 40.0\%. Of the 102 runs without full credit, 85 are \texttt{VOID} (82 reach the step limit and 3 the wall-clock limit), 10 deliver a report that is wrong or does not parse (three of them earn partial credit), and 7 stop early without a report. We read each trajectory and assign each of these runs one primary cause. The causes are ordered along the execution path, as in the geoscience analysis (Table~\ref{tab:an-taxonomy}). Table~\ref{tab:an-matrix} gives the outcome of every run.

\paragraph{Successful runs.} Four backbones account for 48 of the 66 successful runs: Fable~5.1 solves all 14 tasks, GPT-6-Astra and Opus~5 solve 12 each, and GPT-5.6-sol solves 10. The successes take a median of 44.5 steps (range 14 to 100) and mix two ways of working. Sixty-two of them operate Fluent's graphical interface and also type at least one text command into its console. Two (Kimi-K3 on FT-008a and FV-005b) use the panels only, and two (Sonnet~5 and Fable~5.1 on FT-008c) run Fluent without a window from a PowerShell terminal. Among the 117 runs that work only in an interactive Fluent window, typed console commands make up a median of 15\% of the actions in the 64 successes and 3\% in the 53 failures. On FV-003b and FV-005b, whose answers come from a cell-centred ASCII export of the field, the route to the export decides the outcome. The nine runs that produce the export with the console command \texttt{/file/export/ascii} all succeed. Seven runs use the Export dialog instead, whose quantity list filters its entries as one types. Four of these seven fail: three never obtain a usable export, and Kimi-K3 obtains one only at step 85 of FV-003b and runs out of steps. One of the three that pass (GPT-5.6-luna on FV-005b) evaluates the analytical series solution, which the instruction forbids; it differs from the discrete solution by at most $3\times10^{-5}$ relative, well inside the band. Setting up a case is harder than reading one out: the \emph{a} tasks are solved in 19 of 60 runs and the \emph{b} tasks in 27 of 60.

\paragraph{Interaction-layer failures.} In 46 of the 102 runs (45\%), the intended actions never take effect. The system prompt states the $1920\times1080$ frame, but four backbones give coordinates in a scaled frame (A\textsubscript{1}, 39 runs). Between 92\% and 99.6\% of the click actions of Gemini~3.1~Pro, Qwen3.7-plus, MiniMax-M3, and Qwen-CUA fall inside the top-left $1000\times1000$ pixels, against 19--85\% for the other eight backbones. Gemini~3.1~Pro clicks $(580,377)$ for the 2D option of the Fluent Launcher, which sits at $(1113,408)$, and 22 of the 28 runs of Gemini~3.1~Pro and Qwen3.7-plus never get past the launcher. Qwen-CUA clicks $(699,347)$ for an OK button at $(1049,522)$, which is the $1280\times720$ frame. The four backbones score 0 in all 56 of their runs, so, as on the geoscience tasks, the cause follows the backbone and not the task. A second mechanism (A\textsubscript{2}, 5 runs) affects stronger backbones: the keystrokes reach the wrong widget or are lost. On FT-005c and FT-008a, Sonnet~5 types its first console command into the find box of Fluent's console, and none of the commands it types in these two runs appears in the Fluent transcript. On FV-003b and FV-005b, the Export dialog drops characters typed into its quantity filter, so GPT-5.6-luna, GPT-5.6-terra, and Sonnet~5 never produce a usable export. One further run issues no executable action (B), and one repeats a single action block hundreds of times (C). No run of Fable~5.1, GPT-6-Astra, Opus~5, or GPT-5.6-sol fails at this stage.

\paragraph{Runs that exhaust the budget.} Cause H covers 38 runs whose actions take effect but which write no gradable report, 36 at the step limit and 2 at the wall-clock limit. None of them runs out while a solve is in progress. Eight have every graded value on screen. GPT-5.6-sol (FT-006b, FV-004b), GPT-5.6-terra (FT-006b), and Sonnet~5 (FV-002a, FV-002b) recompute values they already hold, and Kimi-K3 spends its last steps on consistency checks (FT-006b, FV-002b, FV-004a). Six are still extracting values when the budget ends; one of them (GPT-5.6-luna on FV-001b) is already trying to write two wrong values. Twelve run out while setting up an \emph{a} task, and eleven of them never iterate the solver. Seven of these twelve reopen or retype settings they have already applied, and they include all four failures of GPT-6-Astra and Opus~5. GPT-6-Astra completes the setup of FV-001a by step 54 and spends the remaining 46 steps re-checking it without pressing Calculate; on FV-004a it re-checks for 63 steps. Opus~5 re-enters console dialogues that the Fluent transcript shows were already accepted (FV-002a, FV-005a). The agent sees only its last five steps, so a setup completed earlier drops out of view and is done again. The remaining twelve runs are MiniMax-M3 and Qwen-CUA runs that drive Fluent through batch journals or scripts, which stall on invalid commands, processes that never return, or licence and Python errors. Waiting for the solver costs steps but does not by itself exhaust the budget. The runs issue 521 \texttt{WAIT} replies, 72\% of them on the three tasks with solves of several minutes (FT-005c, FT-008c, FV-003a). Four successful runs finish within four steps of the limit, two of them after long waits: Sonnet~5 on FT-008c after 59 \texttt{WAIT} replies and Fable~5.1 on FV-003a after 65.

\paragraph{Delivery, verification, and method.} All ten runs that deliver a report scoring below 1 come from the three GPT-5.6 variants; every other backbone delivers either a fully correct report or none. Writing a small JSON file on Windows is itself unreliable. In at least 15 runs that hold correct values, the first attempt produces no file, an empty file, or invalid JSON. The causes include a Notepad dialog that swallows the typed text, a Linux terminal shortcut that opens nothing, and \texttt{cmd echo} quoting that writes literal backslashes. Nine of these runs open or list the file afterwards and repair it, and one succeeds after about thirty blind attempts. The other five are the D runs, and GPT-5.6-sol even has the truncated line on screen at step 90 of FT-005c and saves it. Method errors (E, 9 runs) and failed self-checks (G, 2 runs) come mostly from GPT-5.6-luna and GPT-5.6-terra (10 of the 11). Most are slips in single dialogs: a force observable without a zone, the flow equation switched off, an inlet velocity typed as 11 instead of 1 or never set, the temperature-dependent \emph{power-law} viscosity chosen for a shear-thinning liquid, and a report taken on the wrong surface. In seven of these runs the screen shows a physically implausible number: zero drag on the cylinder, all-zero adjoint residuals at the first iteration, a drag curve flat at its pre-solve value, a peak velocity of 16.2~m/s in a channel fed at 1~m/s, an inlet gauge pressure of 0~Pa, a station pressure equal to the inlet mean, and zero wall heat rates. None of the seven finds the setup error behind the number, and GPT-5.6-luna instead divides its FV-001a peak velocity by ten. The remaining E run is MiniMax-M3, which deletes the supplied mesh on FV-005a while cleaning the working directory. Qwen-CUA twice reports invented values as a finished task (F).

\begin{table}[t]
    \centering
    \footnotesize
    \setlength{\tabcolsep}{4pt}
    \renewcommand{\arraystretch}{1.05}
    \caption{Primary causes of the 102 ANSYS runs without full credit (99 failed and 3 partially credited runs), ordered along the execution path. Each run has exactly one cause. The last column lists the affected backbones with their number of runs.}
    \label{tab:an-taxonomy}
    \begin{tabular}{@{}l@{\hspace{5pt}}p{0.17\textwidth}p{0.37\textwidth}cp{0.25\textwidth}@{}}
        \toprule
        & Cause & Criterion & Runs & Backbones (runs) \\
        \midrule
        \multicolumn{5}{@{}l}{\textit{Action does not take effect}} \\
        A\textsubscript{1} & Coordinate-frame mismatch & Clicks follow a scaled frame (normalized or $1280\times720$) instead of the $1920\times1080$ frame stated in the system prompt, so launcher and dialog clicks miss their target & 39 & Gemini~3.1~Pro (14), Qwen3.7-plus (14), MiniMax-M3 (6), Qwen-CUA (5) \\
        A\textsubscript{2} & Input lost in a widget & Typed text lands in the console's find box or is dropped by the Export dialog's quantity filter, so commands and selections never take effect & 5 & Sonnet~5 (3), GPT-5.6-luna, GPT-5.6-terra (1 each) \\
        B & No executable action & Replies contain no executable action until the stall limit & 1 & MiniMax-M3 (1) \\
        C & Malformed action & Replies chain hundreds of identical action blocks, so no step makes progress & 1 & Qwen-CUA (1) \\
        \multicolumn{5}{@{}l}{\textit{Artifact is malformed}} \\
        D & Report lost at the file write & Correct values are on screen, but the report is missing, truncated, or invalid JSON & 5 & GPT-5.6-terra (3), GPT-5.6-sol, GPT-5.6-luna (1 each) \\
        \multicolumn{5}{@{}l}{\textit{Method is inadequate}} \\
        E & Setup or extraction error & Wrong boundary condition, material law, equation set, observable zone, or report surface, or a deleted input file, so the solution or the reported quantity is wrong & 9 & GPT-5.6-luna (6), GPT-5.6-terra (2), MiniMax-M3 (1) \\
        \multicolumn{5}{@{}l}{\textit{Result is not verified}} \\
        F & Invented values with a completion claim & Agent declares the task done with values that no solver produced & 2 & Qwen-CUA (2) \\
        G & Failed self-check & A check or the screen exposes the error, but the agent concludes that its setup or state is correct & 2 & GPT-5.6-terra (2) \\
        \multicolumn{5}{@{}l}{\textit{Budget runs out}} \\
        H & Step or wall-clock limit with no report & Actions take effect, but no gradable report is written within the budget & 38 & Sonnet~5 (7), Kimi-K3, MiniMax-M3, Qwen-CUA (6 each), GPT-5.6-sol, GPT-5.6-luna, GPT-5.6-terra (3 each), Opus~5, GPT-6-Astra (2 each) \\
        \bottomrule
    \end{tabular}
\end{table}

\begin{table}[t]
    \centering
    \footnotesize
    \setlength{\tabcolsep}{3pt}
    \renewcommand{\arraystretch}{1.05}
    \caption{Outcome of every run on the 14 ANSYS Fluent tasks (one run per cell, step limit 100; nine successful runs used an earlier limit of 60 to 80 steps). P: successful run. A--H: primary cause of failure as in Table~\ref{tab:an-taxonomy}, shaded by stage; a superscript gives the partial-credit score in percent. Pass: number of successful runs. Score: mean partial-credit score in percent.}
    \label{tab:an-matrix}
    \resizebox{\textwidth}{!}{%
    \begin{tabular}{@{}lcccccccccccccccc@{}}
        \toprule
        & \multicolumn{4}{c}{FT} & \multicolumn{10}{c}{FV} & & \\
        \cmidrule(lr){2-5}\cmidrule(lr){6-15}
        Backbone & 005c & 006b & 008a & 008c & 001a & 001b & 002a & 002b & 003a & 003b & 004a & 004b & 005a & 005b & Pass & Score \\
        \midrule
        Fable~5.1 & \cellcolor{ansP}P & \cellcolor{ansP}P & \cellcolor{ansP}P & \cellcolor{ansP}P & \cellcolor{ansP}P & \cellcolor{ansP}P & \cellcolor{ansP}P & \cellcolor{ansP}P & \cellcolor{ansP}P & \cellcolor{ansP}P & \cellcolor{ansP}P & \cellcolor{ansP}P & \cellcolor{ansP}P & \cellcolor{ansP}P & 14/14 & 100.0 \\
        GPT-6-Astra & \cellcolor{ansP}P & \cellcolor{ansP}P & \cellcolor{ansP}P & \cellcolor{ansP}P & \cellcolor{ansH}H & \cellcolor{ansP}P & \cellcolor{ansP}P & \cellcolor{ansP}P & \cellcolor{ansP}P & \cellcolor{ansP}P & \cellcolor{ansH}H & \cellcolor{ansP}P & \cellcolor{ansP}P & \cellcolor{ansP}P & 12/14 & 85.7 \\
        Opus~5 & \cellcolor{ansP}P & \cellcolor{ansP}P & \cellcolor{ansP}P & \cellcolor{ansP}P & \cellcolor{ansP}P & \cellcolor{ansP}P & \cellcolor{ansH}H & \cellcolor{ansP}P & \cellcolor{ansP}P & \cellcolor{ansP}P & \cellcolor{ansP}P & \cellcolor{ansP}P & \cellcolor{ansH}H & \cellcolor{ansP}P & 12/14 & 85.7 \\
        GPT-5.6-sol & \cellcolor{ansD}D & \cellcolor{ansH}H & \cellcolor{ansP}P & \cellcolor{ansP}P & \cellcolor{ansP}P & \cellcolor{ansP}P & \cellcolor{ansP}P & \cellcolor{ansP}P & \cellcolor{ansP}P & \cellcolor{ansP}P & \cellcolor{ansP}P & \cellcolor{ansH}H & \cellcolor{ansH}H & \cellcolor{ansP}P & 10/14 & 71.4 \\
        Kimi-K3 & \cellcolor{ansP}P & \cellcolor{ansH}H & \cellcolor{ansP}P & \cellcolor{ansP}P & \cellcolor{ansH}H & \cellcolor{ansP}P & \cellcolor{ansH}H & \cellcolor{ansH}H & \cellcolor{ansP}P & \cellcolor{ansH}H & \cellcolor{ansH}H & \cellcolor{ansP}P & \cellcolor{ansP}P & \cellcolor{ansP}P & 8/14 & 57.1 \\
        Sonnet~5 & \cellcolor{ansA}A\textsubscript{2} & \cellcolor{ansH}H & \cellcolor{ansA}A\textsubscript{2} & \cellcolor{ansP}P & \cellcolor{ansH}H & \cellcolor{ansP}P & \cellcolor{ansH}H & \cellcolor{ansH}H & \cellcolor{ansH}H & \cellcolor{ansP}P & \cellcolor{ansH}H & \cellcolor{ansP}P & \cellcolor{ansH}H & \cellcolor{ansA}A\textsubscript{2} & 4/14 & 28.6 \\
        GPT-5.6-terra & \cellcolor{ansD}D & \cellcolor{ansH}H & \cellcolor{ansF}G\textsuperscript{33} & \cellcolor{ansD}D & \cellcolor{ansH}H & \cellcolor{ansD}D & \cellcolor{ansP}P & \cellcolor{ansE}E & \cellcolor{ansP}P & \cellcolor{ansA}A\textsubscript{2} & \cellcolor{ansF}G & \cellcolor{ansE}E\textsuperscript{56} & \cellcolor{ansH}H & \cellcolor{ansP}P & 3/14 & 27.8 \\
        GPT-5.6-luna & \cellcolor{ansP}P & \cellcolor{ansP}P & \cellcolor{ansE}E\textsuperscript{33} & \cellcolor{ansE}E & \cellcolor{ansE}E & \cellcolor{ansH}H & \cellcolor{ansH}H & \cellcolor{ansD}D & \cellcolor{ansE}E & \cellcolor{ansA}A\textsubscript{2} & \cellcolor{ansE}E & \cellcolor{ansH}H & \cellcolor{ansE}E & \cellcolor{ansP}P & 3/14 & 23.8 \\
        Gemini~3.1~Pro & \cellcolor{ansA}A\textsubscript{1} & \cellcolor{ansA}A\textsubscript{1} & \cellcolor{ansA}A\textsubscript{1} & \cellcolor{ansA}A\textsubscript{1} & \cellcolor{ansA}A\textsubscript{1} & \cellcolor{ansA}A\textsubscript{1} & \cellcolor{ansA}A\textsubscript{1} & \cellcolor{ansA}A\textsubscript{1} & \cellcolor{ansA}A\textsubscript{1} & \cellcolor{ansA}A\textsubscript{1} & \cellcolor{ansA}A\textsubscript{1} & \cellcolor{ansA}A\textsubscript{1} & \cellcolor{ansA}A\textsubscript{1} & \cellcolor{ansA}A\textsubscript{1} & 0/14 & 0.0 \\
        Qwen3.7-plus & \cellcolor{ansA}A\textsubscript{1} & \cellcolor{ansA}A\textsubscript{1} & \cellcolor{ansA}A\textsubscript{1} & \cellcolor{ansA}A\textsubscript{1} & \cellcolor{ansA}A\textsubscript{1} & \cellcolor{ansA}A\textsubscript{1} & \cellcolor{ansA}A\textsubscript{1} & \cellcolor{ansA}A\textsubscript{1} & \cellcolor{ansA}A\textsubscript{1} & \cellcolor{ansA}A\textsubscript{1} & \cellcolor{ansA}A\textsubscript{1} & \cellcolor{ansA}A\textsubscript{1} & \cellcolor{ansA}A\textsubscript{1} & \cellcolor{ansA}A\textsubscript{1} & 0/14 & 0.0 \\
        MiniMax-M3 & \cellcolor{ansA}A\textsubscript{1} & \cellcolor{ansA}A\textsubscript{1} & \cellcolor{ansH}H & \cellcolor{ansA}B & \cellcolor{ansH}H & \cellcolor{ansA}A\textsubscript{1} & \cellcolor{ansH}H & \cellcolor{ansA}A\textsubscript{1} & \cellcolor{ansA}A\textsubscript{1} & \cellcolor{ansH}H & \cellcolor{ansA}A\textsubscript{1} & \cellcolor{ansH}H & \cellcolor{ansE}E & \cellcolor{ansH}H & 0/14 & 0.0 \\
        Qwen-CUA & \cellcolor{ansA}C & \cellcolor{ansF}F & \cellcolor{ansA}A\textsubscript{1} & \cellcolor{ansA}A\textsubscript{1} & \cellcolor{ansA}A\textsubscript{1} & \cellcolor{ansH}H & \cellcolor{ansH}H & \cellcolor{ansH}H & \cellcolor{ansH}H & \cellcolor{ansH}H & \cellcolor{ansH}H & \cellcolor{ansF}F & \cellcolor{ansA}A\textsubscript{1} & \cellcolor{ansA}A\textsubscript{1} & 0/14 & 0.0 \\
        \midrule
        Pass & 5/12 & 4/12 & 5/12 & 6/12 & 3/12 & 6/12 & 4/12 & 4/12 & 6/12 & 5/12 & 3/12 & 5/12 & 3/12 & 7/12 & 66/168 & 40.0 \\
        Score & 41.7 & 33.3 & 47.2 & 50.0 & 25.0 & 50.0 & 33.3 & 33.3 & 50.0 & 41.7 & 25.0 & 46.3 & 25.0 & 58.3 & & \\
        \bottomrule
    \end{tabular}}
\end{table}

\begin{figure*}[ht]
    \centering
    \includegraphics[width=0.9\textwidth]{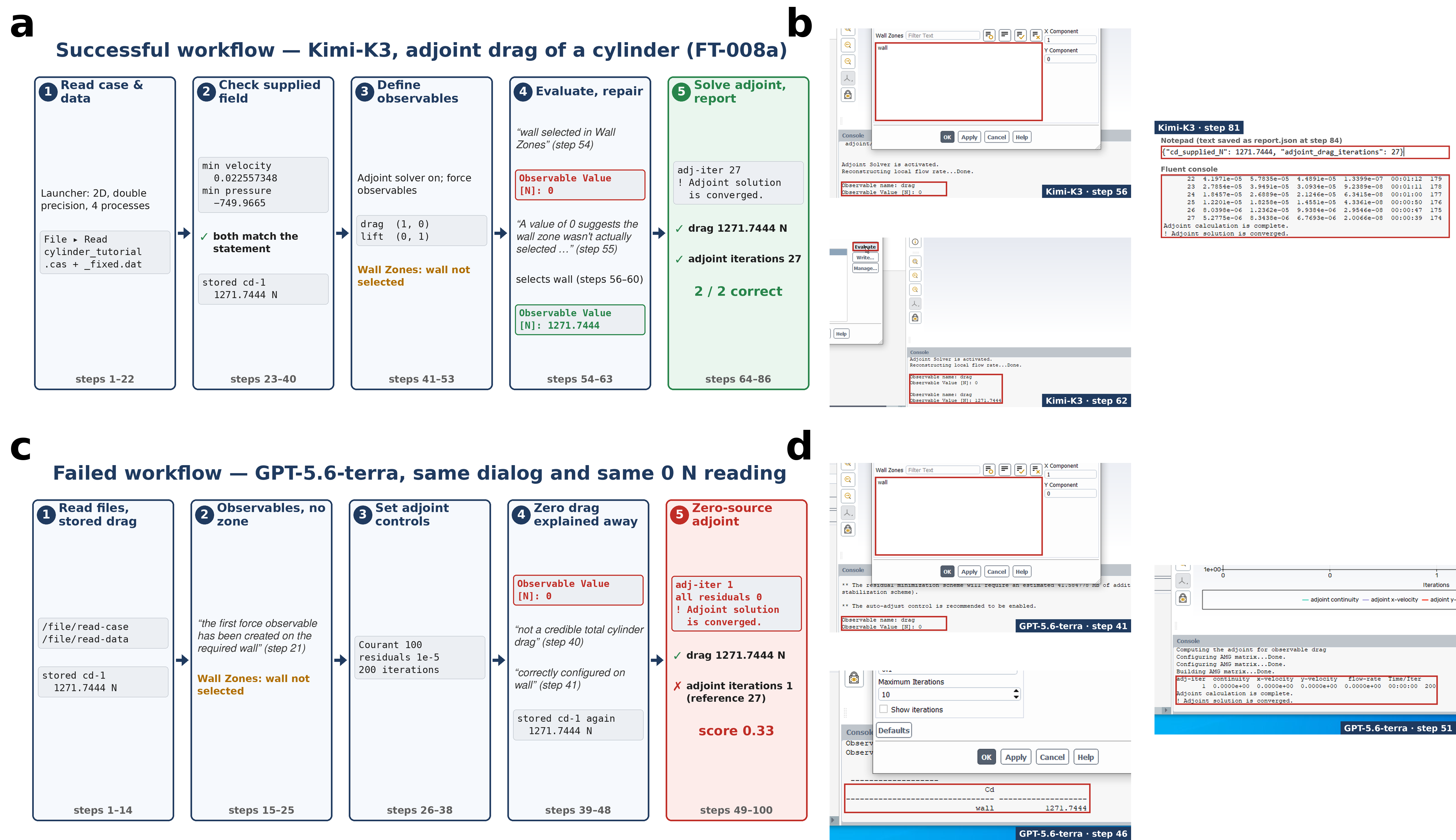}
    \caption{Contrasting two trajectories on the adjoint task FT-008a, with raw frames from each run shown to the right of its workflow diagram. (a) and (b): Kimi-K3 evaluates its drag observable, reads 0~N, selects the wall zone it had left unselected, re-evaluates to 1271.7444~N, and runs the adjoint to convergence at iteration 27 (score 1.0, 86 steps). (c) and (d): GPT-5.6-terra reaches the same dialog with the same 0~N reading, concludes that the observable is correctly configured, obtains 1271.7444~N from the stored drag report without reconciling the two values, and runs a zero-source adjoint that stops at iteration 1 (score 0.33, 100 steps). Each frame is the run's own screenshot with the relevant region boxed. The decisive difference is whether an implausible value may overrule the agent's reading of its setup, not the trajectory length or the number of checks.}
    \label{fig:an-case}
\end{figure*}

\paragraph{Case study: a zero-valued observable on the adjoint task.} FT-008a asks for the drag on a cylinder in a supplied flow at Re${}=40$ and for the number of iterations that the drag adjoint needs to converge (27, two thirds of the weight). In these runs the count turns on defining a force observable on the cylinder wall. The dialog lists the zone \texttt{wall} but does not select it. An observable without a zone evaluates to 0~N and gives the adjoint a zero source, which Fluent reports as converged after one iteration. GPT-5.6-luna, GPT-5.6-terra, and Kimi-K3 all create their observables without selecting the zone. Figure~\ref{fig:an-case} contrasts Kimi-K3 (score 1.0, 86 steps) with GPT-5.6-terra (0.33, 100 steps). Both reach the same screen, the observable dialog with \texttt{wall} unselected above a console line that reads \texttt{Observable Value [N]: 0} (Kimi-K3 at step~56, GPT-5.6-terra at step~41), and they respond in opposite ways.

\emph{Successful trajectory.} Kimi-K3 reads the legacy case and data through the File menu (steps 9--22), checks the minimum velocity and pressure against the values in the statement (steps 33 and 36), computes the stored drag report, activates the adjoint solver, and creates the drag and lift observables (steps 39--52). Like GPT-5.6-terra, it reads the unselected list as selected (``wall selected in Wall Zones'', step~54); it then evaluates the drag observable, and the console prints 0~N. Its visible replies at steps 54 to 59 are empty, but its reasoning at step~55 notes that ``A value of 0 suggests the wall zone wasn't actually selected in the observable'', and its actions reopen the observable manager (step~55) and select \texttt{wall} for both observables (steps 56--60). The new evaluation (step~61) prints 1271.7444~N, the value of the stored drag report. It then sets the adjoint controls and monitors, runs the adjoint, which converges at iteration 27, and writes both values (steps 64--86). Both quantities are within the band.

\emph{Failed trajectory.} GPT-5.6-terra reads the files and computes the stored drag of 1271.7444~N in the text console (steps 5--11). It creates the two observables and notes that the first ``has been created on the required wall'' (step~21). After evaluating the drag observable, it reads 0~N and calls it ``not a credible total cylinder drag'' (step~40). At step~41, facing the dialog with the zone unselected, it concludes that the observable ``is correctly configured on wall'' and that ``the loaded flow field must be checked''. It recomputes the stored report, obtains 1271.7444~N again (step~45), and starts the adjoint without reconciling the two values. The adjoint stops at iteration 1 with all residuals zero (step~51). The run writes the count 1, which it has just called inconsistent, then quits, relaunches Fluent, spends steps 58--89 reading the files again, and reaches the step limit while re-creating the observable. The drag is correct and the count is not (1 against 27), for a score of 0.33. The third run with the omission, GPT-5.6-luna, never evaluates its observable and reports 2 iterations. All five successful runs on this task evaluate the drag observable before running the adjoint and read 1271.7444~N, Kimi-K3 after its first reading of 0~N.

\emph{Takeaway.} This case shows that a check helps only if its result is allowed to overrule the agent's reading of its own setup. Both runs misread the same unselected list as selected. Kimi-K3 lets the 0~N reading overturn that belief within one step. GPT-5.6-terra runs the right check, recognizes the value as implausible, and holds two numbers that contradict each other, yet it interprets both through the belief that the observable is attached to the wall. As in the chemistry case study (Figure~\ref{fig:opus5-case-studies}), the error survives because each check is read through the assumption it should test. For simulation tasks, agents should treat a physically impossible value, such as zero drag on a bluff body or zero residuals at the first iteration, as evidence against the setup, and benchmarks should ask for quantities, such as an iteration count, that expose silent setup errors.

\paragraph{Scope of the analysis.} The analysis has one run per backbone and task, so the counts describe these 168 trajectories and carry no variance estimate. Each run without full credit receives a single primary cause; where a run shows several faults, we assign the one that determined the outcome. Attempts lost to infrastructure faults were repeated, and only the run of record is analyzed. An earlier batch of GPT-6-Astra with the same instructions covers nine of the tasks. Leaving aside one attempt lost to an infrastructure fault, it agrees with the runs of record on six of eight tasks and flips two: FV-004a (solved earlier, failed here) and FV-005a (failed earlier, solved here). Single outcomes of long runs should therefore be read with caution. Nine successful runs were recorded under an earlier per-task limit of 60 to 80 steps and finished within 45 steps, so the lower limit did not bind. Ten of Fable~5.1's runs end with a DONE written as prose, which stops the run after five replies without an action; the report is already written, and the scores are unaffected. Kimi-K3 runs its own tool-calling loop under the same step, image, and token budgets, so its trajectory shapes are only partly comparable. Some graded values can in principle be obtained without the solver, as the analytical series on FV-005b shows. All 66 successful runs opened a Fluent session, but a success alone does not prove that every reported value came from Fluent.

%% file: appendix_openfoam.tex
\subsection{Physics Analysis III: Openfoam in FLUID DYNAMICS}
\label{sec:app-cfd}
\suppressfloats[t]

\definecolor{cfdA}{rgb}{0.98,0.85,0.82}
\definecolor{cfdB}{rgb}{0.99,0.92,0.78}
\definecolor{cfdE}{rgb}{0.84,0.90,0.98}
\definecolor{cfdG}{rgb}{0.89,0.86,0.96}
\definecolor{cfdH}{rgb}{0.88,0.88,0.88}
\definecolor{cfdP}{rgb}{0.84,0.94,0.84}

This appendix analyzes the trajectories on the 14 computational fluid dynamics tasks, which run in OpenFOAM v2512 on an Xfce desktop. Each task supplies a coarse and a fine mesh and asks for the converged cell-center solution on both, delivered as two CSV matrices. The tasks are Taylor--Couette flow (OF-001), natural convection in a heated square cavity (OF-002), turbulent plane channel flow (OF-004), laminar flow through a porous block (OF-005), flow around a cylinder in a channel (OF-007), laminar pipe flow (OF-009), a power-law fluid in a plane channel (OF-010), compressible flow through a converging--diverging duct (OF-011), a turbulent flat-plate boundary layer (OF-012), Joule heating of a solid conductor (OF-013), turbulent flow over a wall-mounted bump (OF-015), one blade passage of a mixer in a rotating frame (OF-016), a planar sudden expansion (OF-017), and a transitional flat-plate boundary layer (OF-018). OF-013 prescribes the discretization in full and leaves the solver free, so a run may solve it with its own code. Four gates score each run. G1 checks the format, the match of every row to a mesh cell, and at least 8 printed significant digits. G2 restarts our own solver setup from the delivered fields and checks that they are a fixed point. G3 computes quantities of interest from both matrices and checks that they converge between the two meshes and lie within a reference band on each mesh. G4 checks for shortcuts, for example a field interpolated from the other mesh. A run scores 1 when all four gates pass and 0.0625 per passed gate otherwise. We evaluate twelve backbones with one run per task, a limit of 100 interaction steps and 3~h of wall-clock time, and a screen resolution of $1920\times1080$, which gives 168 runs. Seventy-seven runs succeed, and the mean partial-credit score is 47.3\%. Of the 91 failed runs, 53 are \texttt{VOID}, 25 deliver both matrices (22 of them earn partial credit), and 13 end with an incomplete or empty submission. A run is \texttt{VOID} when it reaches the step or wall-clock limit without issuing \texttt{DONE} and submits nothing, and it scores 0. We read each trajectory, replay the run's own case where the cause is in doubt, and assign each failed run one primary cause (Table~\ref{tab:cfd-taxonomy}). Table~\ref{tab:cfd-matrix} gives the outcome of every run. Just over half of the failed runs leave at least one required mesh without a single solver iteration, and a fifth have a correct case whose solution is unfinished or lost at export.

\paragraph{Successful runs.} Four backbones account for 48 of the 77 successful runs: Fable~5.1 passes all 14 tasks, GPT-6-Astra 13, GPT-5.6-sol 11, and Opus~5 10. OF-009 and OF-013 are passed most often (9 runs each), and OF-016 least often (Fable~5.1 and GPT-5.6-sol only). A successful run takes a median of 23 steps. A search of the commands of every run (OF-013 excluded) shows what the passes share and what they do not. They reach a running solver early, at a median of step~7, against step~14 for the failed runs that launch at all. Once the solver runs, the passes are not more patient than the failures. Commands that sleep for at least 30~s appear in 55 of the 68 passes (81\%) and in 34 of the 37 failures with causes F--I (92\%). The median number of \texttt{WAIT} replies, the idle reply that pauses for 2~s and uses one of the 100 steps, is zero among both the passes and the failures. On the large meshes of OF-015, OF-017, and OF-018, 11 of the 14 passes decompose the case and solve it in parallel. Consulting the tutorials does not separate the groups either: commands that name the tutorial directory appear in 31 of the 68 passes (46\%) and in 46 of the 88 failures (52\%). In short, a typical pass reaches a running solver quickly and, on the large meshes, solves in parallel.

\paragraph{Most failures never solve both meshes.} Forty-nine of the 91 failed runs (54\%) have causes A--D, in which at least one required mesh never reaches a solver iteration. Actions that miss their target decide few of them, unlike on the geoscience tasks (Appendix~\ref{sec:app-geo}), because the agents drive OpenFOAM through a terminal. Only one of the five A failures is a misclick (Qwen-CUA on OF-013). The other four lose a complete case generator or solver before it reaches the disk. Three lose it to a syntax error from nested quoting, which stops the action code before any keystroke, and one to a here-document that is closed with nothing typed into it. None of them sends it again in one piece. Our harness returns only a screenshot after each action, so the agent does not see such a syntax error. Five successful runs hit the same kind of error and resend the generator within one or two steps. In 14 runs (C), the solver is never launched on at least one mesh, and the steps go to reading. Sonnet~5 accounts for seven of these. On OF-002, it has a nearly correct case by step~41 and spends the remaining 59 steps listing and re-reading its own files. Kimi-K3 spends 81, 69, and 86 of 100 steps reading the OpenFOAM installation on OF-002, OF-016, and OF-017. In 26 runs (D), every launch on at least one mesh stops at a start-up error before the first iteration, and the run often ends in a read-only loop, or after a last fix that is never relaunched. The cause of failure follows the backbone. Fable~5.1, GPT-6-Astra, GPT-5.6-sol, and Opus~5 have no failure with causes A--D, and all eight of their failures occur after the solver has iterated on both meshes. MiniMax-M3, GPT-5.6-luna, Qwen-CUA, Qwen3.7-plus, and Sonnet~5 pass 11 of their 70 runs and account for 40 of the 49 early failures.

\paragraph{Solving a different problem.} Among the runs that compute a solution, the largest class is a setting that contradicts the statement (F, 20 runs). Six are dictionary entries that switch off part of the set-up without stopping the solver: a block named \texttt{SIMPLEC} that the solver never reads (three runs), a forcing relaxation of 0 that holds the driving pressure gradient at zero, an \texttt{fvOptions} file without a header, and \texttt{consistent} placed inside \texttt{PIMPLE}. Five use a scheme or limiter other than the one the statement prescribes, three of them one that the statement explicitly excludes. Four apply the porous resistance of OF-005 explicitly, although the statement requires the pressure-implicit treatment. Three set a wrong boundary condition, and two keep the tolerances of a tutorial. Five of the 20 runs diverge, and the other 15 converge to a different solution. In eight of the 20, a replay of the agent's own case with one word or line changed scores 1. On OF-015, for example, Kimi-K3 writes \texttt{linearUpwindV}, the vector-limited variant that the statement explicitly excludes (``not the vector-limited variant''). It prints \texttt{fvSchemes} six times in its last 24 steps, but its small terminal leaves the line on the screen in only two of these views, and it never questions the line. Replacing that one word scores 1. In most F runs, the reply states the right physics while the dictionary it types disagrees. Four more runs (E) deliver a closed-form profile or a finite-difference result of their own instead of the simulation, and three of them are by GPT-5.6-luna.

\paragraph{Correct cases that never reach the grader.} Eighteen failures (20\%) have a correct case whose solution is unfinished (G, 7 runs) or lost at export (H 5, I 6). GPT-5.6-terra (4), Opus~5, Gemini~3.1~Pro, and Qwen-CUA (3 each), and GPT-5.6-sol (2) account for 15 of them. For 12 of the 18, a scored replay with a single change restores a score of 1: letting the solve finish, running the exporter, or printing the native fields with 12 digits. Four of the seven unfinished solves are paced by the solver. Opus~5 on OF-012, OF-015, and OF-017 and Gemini~3.1~Pro on OF-015 issue 76, 85, 51, and 34 \texttt{WAIT} replies while the fine mesh runs on a single core. None of the four G failures on OF-015 and OF-017 (these three and GPT-5.6-luna on OF-017) runs in parallel. Opus~5 writes a four-subdomain decomposition dictionary on OF-015 and never uses it. Its single-core fine solve finishes only 26~s before the step limit ends the run, and nothing is exported. The other three G runs fail differently. GPT-5.6-terra stops OF-011 at a fixed 900 iterations with residuals still near $10^{-4}$. GPT-5.6-luna declares \texttt{DONE} on OF-017 with 32 steps left while its fine solve is still running. Qwen-CUA exports nothing after both OF-007 solves converge. At export, the route matters for the lossy exports (I): all six build their matrices from \texttt{foamToVTK} output. Five are Float32 arrays printed with 6 digits, written by GPT-5.6-sol on OF-004 and OF-010 and by GPT-5.6-terra on OF-002, OF-004, and OF-009. In the sixth, point data overwrite the cell data of the same name. Excluding OF-013, \texttt{foamToVTK} or \texttt{pvpython} appears in 14 of the 68 passes (21\%) and in 14 of the 24 failures that deliver both matrices (58\%). Running the converter is not the defect in itself, since the successful run in the case study below also runs \texttt{foamToVTK}. Delivering its output is.

\paragraph{Stopping without a value-level check.} Thirty-one failures end with the agent's own \texttt{DONE} or \texttt{FAIL}, with a median of 66 steps unused. In 18 of them (12 F, 6 I), the agent declares \texttt{DONE} over a wrong setting or a lossy export that it never detects. In 8 of the 25 failures that deliver both matrices, evidence of the defect is on the screen before \texttt{DONE}, such as 6-digit rows, final residuals of order $10^{-4}$, or rows from time 0. The checks that let these runs through are row counts and file shapes, which catch a missing or truncated file but not a wrong value. Such checks do rescue two successful runs: a line count shows Qwen-CUA a one-line CSV on OF-009, and Gemini~3.1~Pro notices header-only files on OF-001. A value-level check rescues a third. On OF-001, Kimi-K3 prints the pressure at the reference location on both meshes, finds that the fine case kept the reference-pressure cell of the coarse mesh, and re-runs and re-exports the fine mesh at step~96 of 100. The fine matrix it had exported at step~71 would have scored 0.

\begin{figure}[t]
    \centering
    \includegraphics[width=\textwidth]{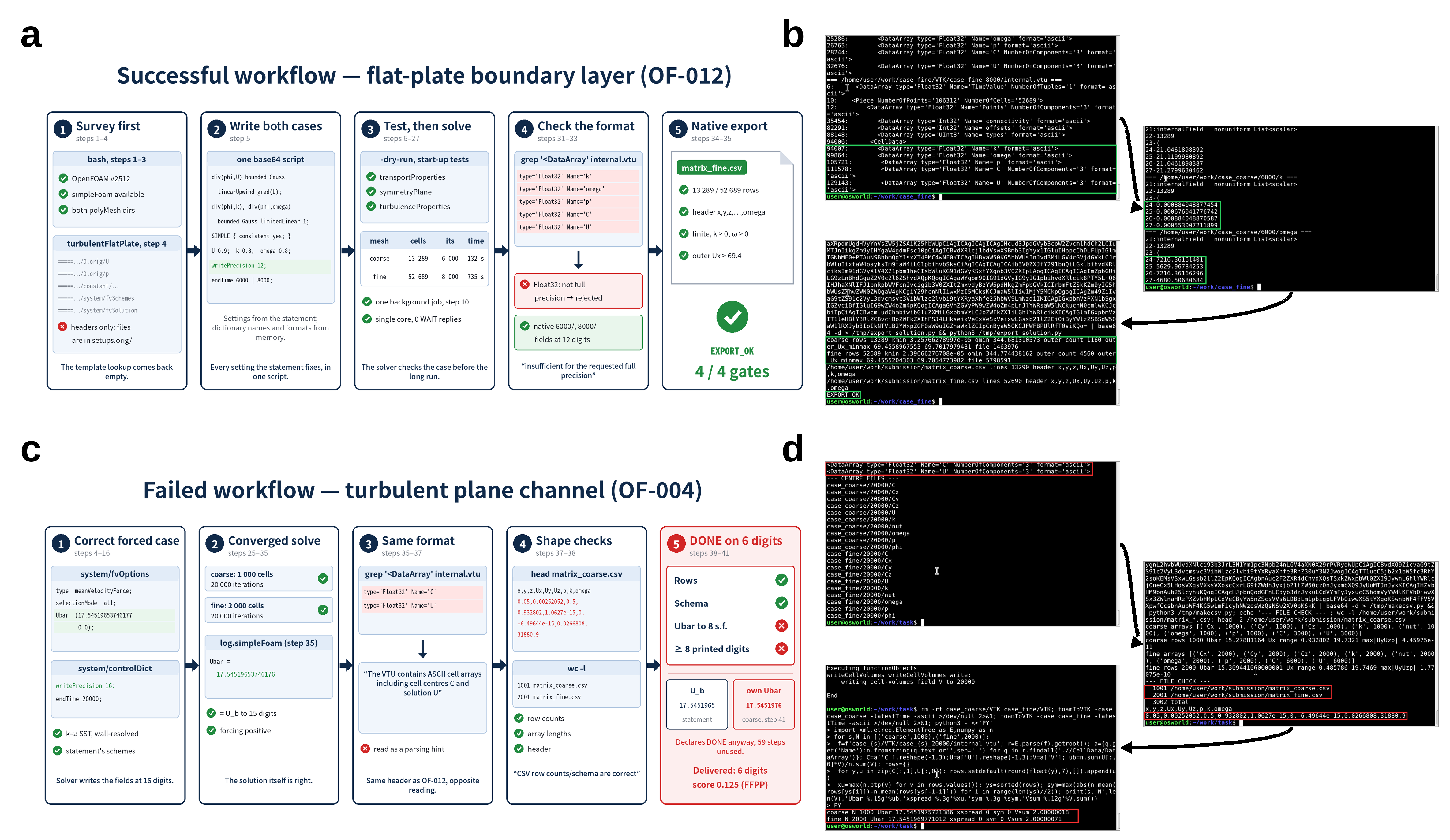}
    \caption{Contrasting GPT-5.6-sol OpenFOAM workflows, with representative raw terminal frames beside each workflow diagram (cropped to the terminal window; a step number refers to the screen the model saw before its action at that step). (a) and (b): the successful flat-plate run (OF-012) inspects the array declarations in its VTK (\texttt{.vtu}) output, rejects the Float32 arrays as short of full precision (step~33), reads the 12-digit native field files (step~34), and checks row counts and physical bounds before \texttt{EXPORT\_OK} (score 1 after 35 steps). (c) and (d): the failed plane-channel run (OF-004) sees the same kind of declaration (step~37), builds its CSV files from 6-digit ASCII VTK arrays and checks their line counts, array lengths, and header (step~38), and declares \texttt{DONE} although its own volume-weighted bulk velocity differs from $U_b$ in the 8th significant figure (step~41; score 0.125 after 41 steps). The string FFPP in (c) lists the gates G1--G4 in order, F for fail and P for pass. Green boxes mark the evidence that the successful run acted on, and red boxes mark what the failed run accepted and delivered. The decisive difference is validation of the delivered precision against the stated requirement, not the correctness of the solve or the number of checks.}
    \label{fig:cfd-case}
\end{figure}

\paragraph{Case study: correct physics, lost digits.} We contrast two GPT-5.6-sol trajectories on wall-resolved $k$--$\omega$ SST tasks (Figure~\ref{fig:cfd-case}). Both computed the correct converged solution, and they ended with opposite outcomes. Both statements require at least 8 printed significant digits in every value. The comparison shows that success depended less on setting up and solving the case than on whether the agent held an intermediate format, and the checks built on it, against that requirement.

\emph{Successful trajectory.} On the flat-plate boundary-layer task (OF-012), GPT-5.6-sol first confirmed the installation and the meshes. Its attempt to print the dictionaries of the \texttt{turbulentFlatPlate} tutorial returned only file names, because this version keeps them in a subdirectory that it did not search. Its next reply nonetheless called the inspection complete, and at step~5 it wrote both cases from memory in one script, with the statement's schemes, limiters, and relaxation factors and \texttt{writePrecision 12}. It then let the solver check them. A dry run and short start-up runs exposed two dictionaries written under the names used by the openfoam.org distribution (\texttt{physicalProperties} and \texttt{momentumTransport}, where v2512 reads \texttt{transportProperties} and \texttt{turbulenceProperties}) and a \texttt{symmetry} condition on a \texttt{symmetryPlane} patch, which it fixed one per step (steps~7--9). Once a start-up test ran stably, it solved both meshes in one background job (6000 and 8000 iterations, 132~s and 735~s on one core). Before exporting, it inspected the array declarations in its VTK (\texttt{.vtu}) output and rejected the Float32 arrays at step~33: ``\emph{The VTU writer reduced values to Float32, which is insufficient for the requested full precision; I will inspect the native ASCII field layout and parse the OpenFOAM internalField values directly instead.}'' It parsed the 12-digit native fields instead and asserted the row counts, the CSV header, finite values, positive $k$ and $\omega$, and the outer-flow velocity before printing \texttt{EXPORT\_OK}. The evaluator passed all four gates (56 of 56 checks; score~1.0 after 35 steps). The productive pattern was to test each tool before relying on it: the solver checked the dictionaries, and the run checked the export format against the statement's precision requirement.

\emph{Failed trajectory.} On the turbulent plane-channel task (OF-004), the same backbone wrote the forced case with \texttt{meanVelocityForce} at the prescribed $U_b = 17.54519653746177$ and \texttt{writePrecision 16} and ran both meshes for 20{,}000 iterations. The solver logs on the screen at step~35 reported \texttt{uncorrected Ubar = 17.54519653746176} on the coarse mesh and \texttt{17.54519653746169} on the fine mesh, and a replay of the run's dictionaries reproduced both logs line for line. At step~37 Float32 array declarations of the same kind were on the screen, and the run read them as a parsing hint: ``\emph{The VTU contains ASCII cell arrays including cell centres C and solution U; convert both meshes to the exact requested CSV schema, then print validation statistics.}'' The statement permits a single-precision path as long as at least 8 digits are printed, but \texttt{foamToVTK -ascii} prints 6, whatever \texttt{writePrecision} is set to, and rewriting the arrays with \texttt{\%.12g} could not restore the lost digits. The run then checked line counts, array lengths, and the CSV header (``\emph{CSV row counts/schema are correct}''). Finally it computed from the same arrays the volume-weighted bulk velocity, which the statement calls the authoritative check. On the coarse mesh this value, 17.5451975721, differed from $U_b$ in the 8th significant figure (relative difference $5.9\times10^{-8}$; the fine-mesh value differed only in the 9th). The run nonetheless declared \texttt{DONE} at step~41 with 59 steps unused. The evaluator failed G1 on the printed digits (6 against 8) and G2 on the first-iteration residuals of the rounded fields, producing a score of~0.125. The printed precision alone decided this score: GPT-5.6-terra's run on this task converged to the same solution, and a replay of its case scored 1 when the fields were printed with 12 digits and 0.125 when they were printed with 6. Once the 6-digit arrays entered the export, every downstream step (reformatting, row counting, and the schema check) faithfully preserved the loss.

\emph{Takeaway.} A correct solve and several validation actions are therefore insufficient when the result of a check is never compared with the tolerance it is meant to test. The row counts and the schema could not see the lost digits, and the check that did, the bulk velocity, was never held against the 8 significant figures that the statement requires. For simulation workflows, agents should compare the delivered numbers, and not only the shape of the files, with the tolerance that the task states, and should treat derived export formats as lossy until shown otherwise. Checking the delivered precision against the stated requirement, and acting on the result, rather than the self-consistency of the files, is what separates the two cases.

\begin{table}[t]
    \centering
    \footnotesize
    \setlength{\tabcolsep}{4pt}
    \renewcommand{\arraystretch}{1.05}
    \caption{Primary causes of the 91 failed runs on the OpenFOAM tasks, ordered along the execution path. Each failed run has exactly one cause: the defect that decides its outcome, and the earliest one along the path when there are several. The path is judged per mesh: a run with a mesh that never iterates gets one of the causes A--D, with two exceptions. The cause is E when the delivered matrices come from a substitute for the simulation (four runs), and F when the agent never reached that mesh because a wrong setting derailed the solve on the other mesh and was never fixed (five runs). The last column lists the affected backbones with their number of runs.}
    \label{tab:cfd-taxonomy}
    \begin{tabular}{@{}l@{\hspace{5pt}}p{0.17\textwidth}p{0.37\textwidth}cp{0.25\textwidth}@{}}
        \toprule
        & Cause & Criterion & Runs & Backbones (runs) \\
        \midrule
        \multicolumn{5}{@{}l}{\textit{Action does not take effect}} \\
        A & Input never reaches the case & Keystrokes land outside the terminal, the action code fails before typing (a syntax error from nested quoting), or the agent's own here-document closes around nothing, and the lost case or solver is never re-sent & 5 & GPT-5.6-luna (2), GPT-5.6-terra, Qwen-CUA, MiniMax-M3 (1 each) \\
        \multicolumn{5}{@{}l}{\textit{No solution is computed}} \\
        B & Gives up before solving & The agent issues \texttt{FAIL} or \texttt{DONE} with steps left, before a solver has run on every required mesh & 4 & Gemini~3.1~Pro (2), GPT-5.6-luna, MiniMax-M3 (1 each) \\
        C & Solver never launched & The step limit ends the run before a solver is launched on at least one required mesh, and this missing launch, not a start-up error, is the decisive defect; the steps go to reading source code, tutorials or the agent's own files & 14 & Sonnet~5 (7), Kimi-K3 (3), MiniMax-M3 (2), Qwen3.7-plus, GPT-5.6-luna (1 each) \\
        D & Launched but never iterates & Every launch on at least one required mesh stops at a start-up error before the first iteration (missing or headerless dictionary, missing scheme entry, wrong patch type), and the run ends with such an error unresolved & 26 & MiniMax-M3 (9), Qwen-CUA (6), Qwen3.7-plus (5), Kimi-K3, GPT-5.6-luna (2 each), Gemini~3.1~Pro, Sonnet~5 (1 each) \\
        \multicolumn{5}{@{}l}{\textit{The simulation solves a different problem}} \\
        E & Substitute for the simulation & The delivered matrices come from a closed-form profile or the agent's own finite-difference code instead of the required solve & 4 & GPT-5.6-luna (3), GPT-5.6-terra (1) \\
        F & Setting contradicts the statement & The solver iterates, but a scheme, limiter, boundary condition, porosity treatment or dictionary entry differs from the statement, so the run diverges or converges to another solution & 20 & Qwen3.7-plus (4), GPT-5.6-terra, GPT-5.6-luna (3 each), Gemini~3.1~Pro, Qwen-CUA (2 each), GPT-6-Astra, GPT-5.6-sol, Opus~5, Kimi-K3, Sonnet~5, MiniMax-M3 (1 each) \\
        \multicolumn{5}{@{}l}{\textit{The solution is not finished}} \\
        G & Unconverged or unfinished solve & The case is correct, but a solve is still running when the run ends or finishes too late for any export to follow, or it is stopped and exported before the statement's convergence criterion & 7 & Opus~5 (3), Gemini~3.1~Pro, GPT-5.6-terra, Qwen-CUA, GPT-5.6-luna (1 each) \\
        \multicolumn{5}{@{}l}{\textit{The solution is lost at export}} \\
        H & Converged solution not exported & A converged solution exists for every mesh and an export is attempted after the solves end, but the exporter crashes or is never run, or the case is overwritten, so a deliverable is missing or empty & 5 & Gemini~3.1~Pro (2), Kimi-K3, Qwen3.7-plus, Qwen-CUA (1 each) \\
        I & Lossy export & Both matrices come from a correct converged state, but the values are degraded on export, for example 6-digit Float32 VTK arrays or point data written as cell data & 6 & GPT-5.6-terra (3), GPT-5.6-sol (2), Qwen-CUA (1) \\
        \bottomrule
    \end{tabular}
\end{table}

\begin{table}[t]
    \centering
    \scriptsize
    \setlength{\tabcolsep}{2.6pt}
    \renewcommand{\arraystretch}{1.05}
    \caption{Outcome of every run on the 14 OpenFOAM tasks (one run per cell, step limit 100). Column headers give the task number, for example 001 for OF-001. P: successful run. A--I: primary cause of failure as in Table~\ref{tab:cfd-taxonomy}, shaded by stage. A superscript V marks a \texttt{VOID} run, which reaches the step or wall-clock limit without \texttt{DONE}, submits nothing, and scores 0. Pass: number of successful runs. Score: mean partial-credit score in percent.}
    \label{tab:cfd-matrix}
    \begin{tabular}{@{}lcccccccccccccccc@{}}
        \toprule
        Backbone & 001 & 002 & 004 & 005 & 007 & 009 & 010 & 011 & 012 & 013 & 015 & 016 & 017 & 018 & Pass & Score \\
        \midrule
        Fable~5.1 & \cellcolor{cfdP}P & \cellcolor{cfdP}P & \cellcolor{cfdP}P & \cellcolor{cfdP}P & \cellcolor{cfdP}P & \cellcolor{cfdP}P & \cellcolor{cfdP}P & \cellcolor{cfdP}P & \cellcolor{cfdP}P & \cellcolor{cfdP}P & \cellcolor{cfdP}P & \cellcolor{cfdP}P & \cellcolor{cfdP}P & \cellcolor{cfdP}P & 14/14 & 100.0 \\
        GPT-6-Astra & \cellcolor{cfdP}P & \cellcolor{cfdP}P & \cellcolor{cfdP}P & \cellcolor{cfdP}P & \cellcolor{cfdP}P & \cellcolor{cfdP}P & \cellcolor{cfdP}P & \cellcolor{cfdP}P & \cellcolor{cfdP}P & \cellcolor{cfdP}P & \cellcolor{cfdP}P & \cellcolor{cfdE}F & \cellcolor{cfdP}P & \cellcolor{cfdP}P & 13/14 & 93.8 \\
        GPT-5.6-sol & \cellcolor{cfdP}P & \cellcolor{cfdP}P & \cellcolor{cfdH}I & \cellcolor{cfdP}P & \cellcolor{cfdP}P & \cellcolor{cfdP}P & \cellcolor{cfdH}I & \cellcolor{cfdP}P & \cellcolor{cfdP}P & \cellcolor{cfdP}P & \cellcolor{cfdP}P & \cellcolor{cfdP}P & \cellcolor{cfdP}P & \cellcolor{cfdE}F & 11/14 & 82.1 \\
        Opus~5 & \cellcolor{cfdP}P & \cellcolor{cfdP}P & \cellcolor{cfdP}P & \cellcolor{cfdP}P & \cellcolor{cfdP}P & \cellcolor{cfdP}P & \cellcolor{cfdP}P & \cellcolor{cfdP}P & \cellcolor{cfdG}G & \cellcolor{cfdP}P & \cellcolor{cfdG}G$^{\mathrm{V}}$ & \cellcolor{cfdE}F & \cellcolor{cfdG}G & \cellcolor{cfdP}P & 10/14 & 72.3 \\
        Kimi-K3 & \cellcolor{cfdP}P & \cellcolor{cfdB}C$^{\mathrm{V}}$ & \cellcolor{cfdH}H$^{\mathrm{V}}$ & \cellcolor{cfdB}D$^{\mathrm{V}}$ & \cellcolor{cfdP}P & \cellcolor{cfdP}P & \cellcolor{cfdP}P & \cellcolor{cfdB}D$^{\mathrm{V}}$ & \cellcolor{cfdP}P & \cellcolor{cfdP}P & \cellcolor{cfdE}F & \cellcolor{cfdB}C$^{\mathrm{V}}$ & \cellcolor{cfdB}C$^{\mathrm{V}}$ & \cellcolor{cfdP}P & 7/14 & 50.9 \\
        Gemini~3.1~Pro & \cellcolor{cfdP}P & \cellcolor{cfdB}D$^{\mathrm{V}}$ & \cellcolor{cfdB}B & \cellcolor{cfdE}F & \cellcolor{cfdP}P & \cellcolor{cfdP}P & \cellcolor{cfdP}P & \cellcolor{cfdH}H & \cellcolor{cfdB}B & \cellcolor{cfdP}P & \cellcolor{cfdG}G & \cellcolor{cfdE}F & \cellcolor{cfdP}P & \cellcolor{cfdH}H$^{\mathrm{V}}$ & 6/14 & 43.8 \\
        GPT-5.6-terra & \cellcolor{cfdE}E & \cellcolor{cfdH}I & \cellcolor{cfdH}I & \cellcolor{cfdE}F & \cellcolor{cfdP}P & \cellcolor{cfdH}I & \cellcolor{cfdP}P & \cellcolor{cfdG}G & \cellcolor{cfdE}F & \cellcolor{cfdP}P & \cellcolor{cfdP}P & \cellcolor{cfdA}A$^{\mathrm{V}}$ & \cellcolor{cfdP}P & \cellcolor{cfdE}F & 5/14 & 42.0 \\
        Sonnet~5 & \cellcolor{cfdP}P & \cellcolor{cfdB}C$^{\mathrm{V}}$ & \cellcolor{cfdE}F & \cellcolor{cfdB}D$^{\mathrm{V}}$ & \cellcolor{cfdP}P & \cellcolor{cfdP}P & \cellcolor{cfdB}C$^{\mathrm{V}}$ & \cellcolor{cfdB}C$^{\mathrm{V}}$ & \cellcolor{cfdB}C$^{\mathrm{V}}$ & \cellcolor{cfdP}P & \cellcolor{cfdB}C$^{\mathrm{V}}$ & \cellcolor{cfdB}C$^{\mathrm{V}}$ & \cellcolor{cfdP}P & \cellcolor{cfdB}C$^{\mathrm{V}}$ & 5/14 & 35.7 \\
        Qwen3.7-plus & \cellcolor{cfdP}P & \cellcolor{cfdB}D$^{\mathrm{V}}$ & \cellcolor{cfdE}F$^{\mathrm{V}}$ & \cellcolor{cfdE}F & \cellcolor{cfdB}C$^{\mathrm{V}}$ & \cellcolor{cfdB}D$^{\mathrm{V}}$ & \cellcolor{cfdP}P & \cellcolor{cfdE}F$^{\mathrm{V}}$ & \cellcolor{cfdB}D$^{\mathrm{V}}$ & \cellcolor{cfdP}P & \cellcolor{cfdB}D$^{\mathrm{V}}$ & \cellcolor{cfdB}D$^{\mathrm{V}}$ & \cellcolor{cfdH}H$^{\mathrm{V}}$ & \cellcolor{cfdE}F$^{\mathrm{V}}$ & 3/14 & 21.9 \\
        Qwen-CUA & \cellcolor{cfdB}D$^{\mathrm{V}}$ & \cellcolor{cfdE}F$^{\mathrm{V}}$ & \cellcolor{cfdB}D$^{\mathrm{V}}$ & \cellcolor{cfdB}D$^{\mathrm{V}}$ & \cellcolor{cfdG}G & \cellcolor{cfdP}P & \cellcolor{cfdH}I & \cellcolor{cfdP}P & \cellcolor{cfdB}D$^{\mathrm{V}}$ & \cellcolor{cfdA}A$^{\mathrm{V}}$ & \cellcolor{cfdE}F$^{\mathrm{V}}$ & \cellcolor{cfdB}D$^{\mathrm{V}}$ & \cellcolor{cfdH}H$^{\mathrm{V}}$ & \cellcolor{cfdB}D$^{\mathrm{V}}$ & 2/14 & 14.7 \\
        GPT-5.6-luna & \cellcolor{cfdE}E & \cellcolor{cfdA}A & \cellcolor{cfdE}F & \cellcolor{cfdE}F & \cellcolor{cfdB}B & \cellcolor{cfdP}P & \cellcolor{cfdE}E & \cellcolor{cfdA}A$^{\mathrm{V}}$ & \cellcolor{cfdB}C$^{\mathrm{V}}$ & \cellcolor{cfdE}F & \cellcolor{cfdB}D$^{\mathrm{V}}$ & \cellcolor{cfdB}D & \cellcolor{cfdG}G & \cellcolor{cfdE}E & 1/14 & 10.7 \\
        MiniMax-M3 & \cellcolor{cfdB}D$^{\mathrm{V}}$ & \cellcolor{cfdB}D$^{\mathrm{V}}$ & \cellcolor{cfdB}D$^{\mathrm{V}}$ & \cellcolor{cfdB}D$^{\mathrm{V}}$ & \cellcolor{cfdB}D & \cellcolor{cfdB}D$^{\mathrm{V}}$ & \cellcolor{cfdB}D$^{\mathrm{V}}$ & \cellcolor{cfdB}D$^{\mathrm{V}}$ & \cellcolor{cfdE}F$^{\mathrm{V}}$ & \cellcolor{cfdA}A$^{\mathrm{V}}$ & \cellcolor{cfdB}C$^{\mathrm{V}}$ & \cellcolor{cfdB}C$^{\mathrm{V}}$ & \cellcolor{cfdB}B & \cellcolor{cfdB}D$^{\mathrm{V}}$ & 0/14 & 0.0 \\
        \midrule
        Pass & 8/12 & 4/12 & 3/12 & 4/12 & 8/12 & 9/12 & 7/12 & 5/12 & 4/12 & 9/12 & 4/12 & 2/12 & 6/12 & 4/12 & 77/168 & 47.3 \\
        \bottomrule
    \end{tabular}
\end{table}

\paragraph{Scope of the analysis.} The analysis has one run per backbone and task, so the counts describe these 168 trajectories and carry no variance estimate. Each failed run receives a single primary cause. Replays run the agent's own case files in the benchmark image, unchanged or with a single change, and are scored by the same grader as the runs. Two backbones run under conditions that differ from the others. Kimi-K3 runs under its developer's own agent loop, which also returns the exit status and error output of each action as text and re-asks the model after a malformed reply without spending a step. Qwen-CUA receives the screenshots downscaled to $1280\times720$, and its click coordinates are scaled back to the full screen. Only two runs, both by Qwen-CUA (OF-005 and OF-013), reach the 3~h limit, and their records are cut there. The launch steps and the practice counts in the paragraphs on successful runs and on correct cases that never reach the grader come from a pattern search over the terminal history and the logged action code of each run, with OF-013 excluded. A match records that a command was issued, not how its output was used, and the search misses commands that a run hid in a script or encoded, for example in base64. The parallel-run counts use the processor directories that each run left behind, and the \texttt{WAIT} counts come from the logged replies.
\clearpage